\documentclass[runningheads]{llncs}
\usepackage[T1]{fontenc}
\usepackage{graphicx}
\usepackage{amsmath}
\usepackage{amssymb}
\usepackage{multirow}

\usepackage{makecell}
\usepackage[pagebackref,breaklinks,colorlinks]{hyperref}
\usepackage[caption=false,font=footnotesize]{subfig}

\begin{document}
\title{Adjust Your Focus: Defocus Deblurring From Dual-Pixel Images Using Explicit Multi-Scale Cross-Correlation}
\titlerunning{Adjust Your Focus: Defocus Deblurring From Dual-Pixel Images}
%
\author{Kunal Swami}
%
\authorrunning{K. Swami}
%
\institute{Samsung Research India Bangalore\\
Indian Institute of Science\\
\email{kunal.swami@samsung.com, kunalswami@iisc.ac.in}}
%
\maketitle              
\begin{abstract}
Defocus blur is a common problem in photography. It arises when an image is captured with a wide aperture, resulting in a shallow depth of field. Sometimes it is desired, e.g., in portrait effect. Otherwise, it is a problem from both an aesthetic point of view and downstream computer vision tasks, such as segmentation and depth estimation. Defocusing an out-of-focus image to obtain an all-in-focus image is a highly challenging and often ill-posed problem. A recent work exploited dual-pixel (DP) image information, widely available in consumer DSLRs and high-end smartphones, to solve the problem of defocus deblurring. DP sensors result in two sub-aperture views containing defocus disparity cues. A given pixel's disparity is directly proportional to the distance from the focal plane. However, the existing methods adopt a na\"{\i}ve approach of a channel-wise concatenation of the two DP views without explicitly utilizing the disparity cues within the network. In this work, we propose to perform an explicit cross-correlation between the two DP views to guide the network for appropriate deblurring in different image regions. We adopt multi-scale cross-correlation to handle blur and disparities at different scales. Quantitative and qualitative evaluation of our multi-scale cross-correlation network (MCCNet) reveals that it achieves better defocus deblurring than existing state-of-the-art methods despite having lesser computational complexity.

\keywords{Dual-Pixel Sensors \and Defocus Deblurring \and Cross-Correlation \and All-in-focus Image}
\end{abstract}
\section{Introduction}
In photography, defocus blur is a common problem. When the aperture size is large, the depth of field becomes shallow, resulting in defocus blur. Sometimes defocus blur is desired, e.g., portrait effect. In other cases, it results from a trade-off between aperture size and shutter speed. To obtain a well-lit sharp image, the photographer can decrease the shutter speed while keeping the aperture fixed or increase the aperture while keeping the shutter speed fixed. The first option can lead to motion blur if the scene is dynamic, whereas the second option decreases the depth of field, leading to defocus blur. Apart from being an aesthetic problem, defocus blur also affects downstream computer vision tasks, such as depth estimation and semantic segmentation. As a result, it is essential to address this problem.

The task of defocus deblurring is ill-posed. If the blur is high, the original scene information is lost. The blur is also spatially varying because it depends on scene depth, the distance of the point from the focal plane, aperture size, and the optical properties of the camera. 

Recently, Abuolaim \textit{et al.} \cite{dpdnet_eccv2020} exploited Dual-Pixel (DP) image information to solve the problem of defocus deblurring. DP sensors provide two sub-aperture views, which contain the defocus disparity cues. Specifically, in DP images, a given pixel's disparity is proportional to its distance from the focal plane. This disparity information can serve as a valuable cue to guide the network about the amount of deblurring required for a pixel or image region. 

However, the authors in \cite{dpdnet_eccv2020} use a na\"{\i}ve approach of a channel-wise concatenation of the two DP views without explicitly utilizing the disparity cues within the network. This approach leads to partial deblurring and artifacts in the output. Therefore, we are motivated to explore correspondence matching or cross-correlation between the dual pixel images within the network architecture to assist the network in accurate deblurring. 

Therefore, in this work, we propose a new architecture to perform explicit cross-correlation between the DP images. More specifically, we adopt multi-scale cross-correlation within the network architecture to guide the network to perform appropriate deblurring in a given image region. The proposed multi-scale cross-correlation mechanism significantly improves network performance at the defocus deblurring task. We also show that the proposed model requires fewer parameters and FLOPS than DPDNet \cite{dpdnet_eccv2020} and other state-of-the-art methods in literature to achieve better defocus deblurring with higher PSNR and SSIM.

To summarize, following are the important contributions of this work:
\vspace{-5pt}
\begin{enumerate}
	\item We propose a new network architecture for DP defocus image deblurring called MCCNet that utilizes an explicit multi-scale cross-correlation between the DP left and right images. 
	\item The proposed model achieves state-of-the-art quantitative and qualitative results on the standard DPDD \cite{dpdnet_eccv2020} dataset and demands lesser computational complexity. We also report the results of several ablation studies to demonstrate the effectiveness of different modules in MCCNet.
\end{enumerate}

\section{Related Work}
\label{sec:relatedwork}

\subsection{Dual-Pixel Sensors}
\label{subsec:dualpixelsensor}

\begin{figure}[t]
	\centering
	\includegraphics[scale=0.42]{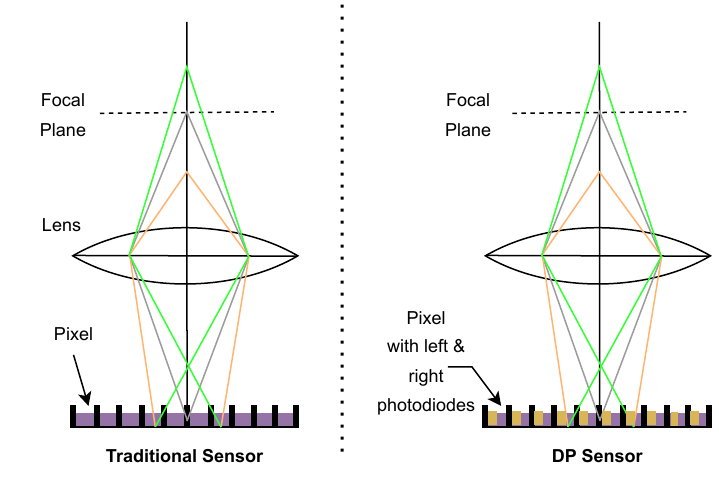}
	\caption{DP sensor image formation.}
	\label{fig:dualpixelsensor}
	\vspace{-14pt}
\end{figure}

Fig.~\ref{fig:dualpixelsensor} shows the image formation in a traditional camera sensor and a DP sensor. A DP sensor contains two photodiodes at each pixel, allowing the pixel reading to split into two halves. As a result, any points which do not lie on the focal plane get distributed across multiple pixels, and exhibit disparity in DP left and right images. Considering the DP left image as a reference, the disparity direction of a point depends on whether the point lies in front (\textcolor{yellow}{yellow} in Fig.~\ref{fig:dualpixelsensor}) or behind (\textcolor{green}{green} in Fig.~\ref{fig:dualpixelsensor}) the focal plane. The amount of disparity of a point depends on the number of sensor pixels it gets spread. This point spread function (PSF) depends on the aperture size and the distance of the point from the focal plane. The defocus disparity information in DP images is a valuable cue to determine the deblurring required for a given pixel or image region. The reader is encouraged to refer to \cite{dpdnet_eccv2020,rdpd_iccv2021} for a detailed technical discussion on DP image formation.

\subsection{Defocus Deblurring}
\label{subsec:defocusdeblurmethods}

The defocus deblurring methods can be classified into two categories in the literature. The methods in the first category adopt a two-stage approach. The first stage estimates a defocus map, and the second uses the defocus map to perform non-blind deconvolution to restore the all-in-focus image. The methods in the second category adopt an end-to-end learning-based approach to generate deblurred output images directly. Representative methods from the first category include \cite{jnb_cvpr2015,ebdb_tip2017,dme_cvpr2019}. \cite{ebdb_tip2017} used image gradients to compute the blur difference between the original and re-blurred images. \cite{dme_cvpr2019} proposed a deep learning based approach for defocus map estimation.

Recently, Abuolaim \textit{et al.} \cite{dpdnet_eccv2020} proposed and performed DP based defocused deblurring for the first time. They adopt a UNet\cite{unet_miccai2015} style encoder-decoder network, which takes the concatenated DP left and right images as input. Abuolaim \textit{et al.} \cite{rdpd_iccv2021} recently also proposed a method to generate a realistic synthetic DP dataset to solve misalignment issues in real-world DP dataset in \cite{dpdnet_eccv2020}. Apart from these methods, a recent method \cite{single_defocus_deblur_wacv2022} adopts a multi-task learning approach using three decoders. The three decoders learn to estimate the left, right, and all-in-focus images, respectively. Lee \textit{et al.} \cite{iter_filter_adapt_cvpr2021} incorporate an auxiliary DP view supervision based disparity estimation task for improving the performance of the main defocus deblurring task.

In contrast to the existing methods, we focus on incorporating an explicit cross-correlation within the network architecture to guide the network about the amount of deblurring required for a given image region.

\section{Proposed Method}
\label{sec:proposedmethod}

The proposed method exploits the disparity information from the two DP images using cross-correlation within the network architecture. In this section, we first describe the proposed network architecture MCCNet, followed by a detailed explanation of different modules in the MCCNet.

\begin{figure*}[t]
	\centering
	\includegraphics[scale=0.31]{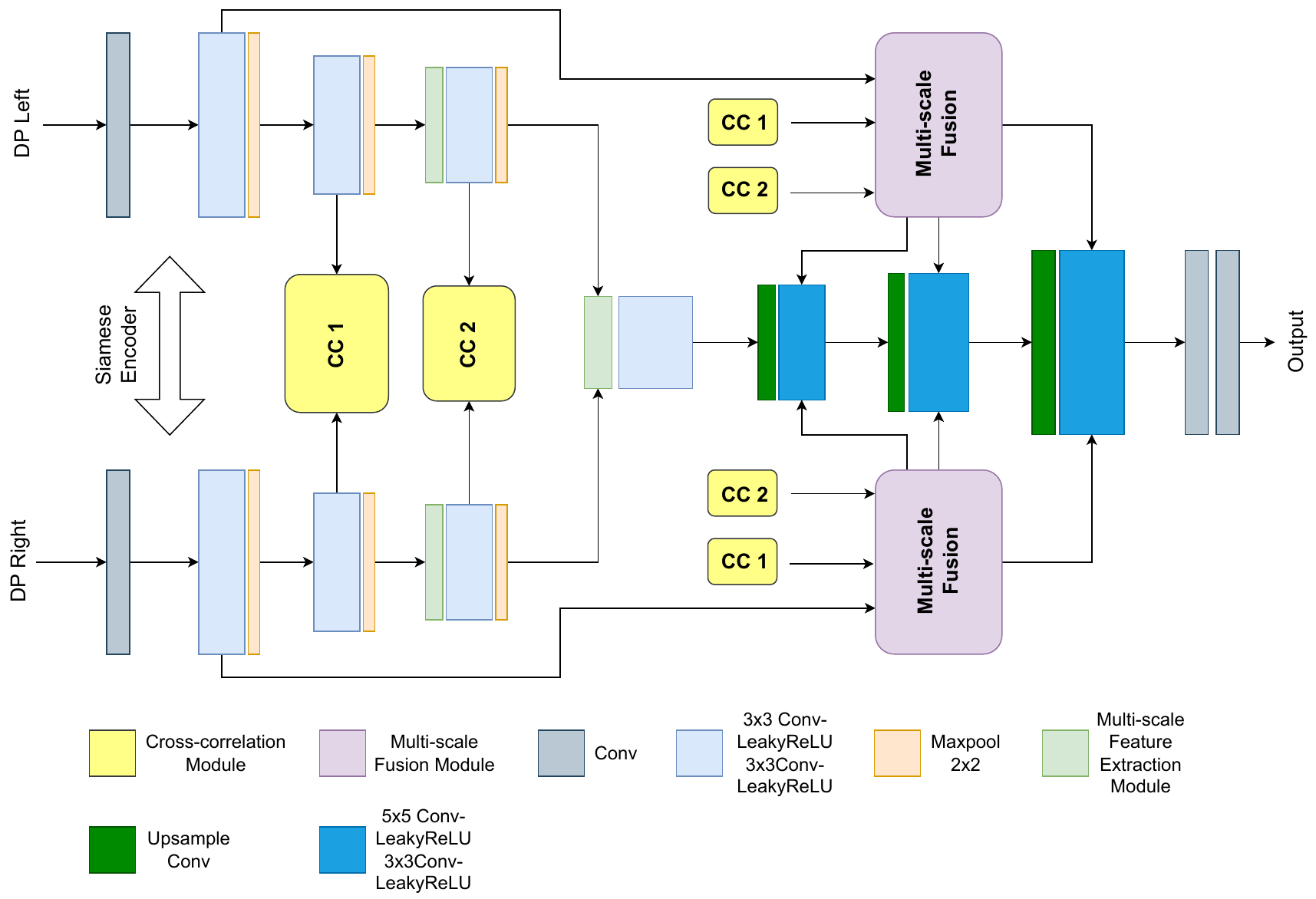}
	\caption{The architecture of the proposed MCCNet architecture. Note that the Cross-correlation module outputs both left and right features. The two Multi-scale Fusion modules (one each for left and right features) share parameters.}
	\label{fig:proposed_arch}
	\vspace{-12pt}
\end{figure*}

\subsection{MCCNet Architecture}
\label{subsec:mccnet}

Fig.~\ref{fig:proposed_arch} shows the detailed architecture of MCCNet. MCCNet adopts a siamese encoder to extract the DP left and right image features. More specifically, the parameters of the left and right encoder branches are shared. There are four encoder blocks, each composed of Conv-LeakyReLU-Conv-LeakyReLU layers. A MaxPool layer for downsampling follows the first, second and third encoder blocks. In the third and fourth encoder blocks, we adopt a Multi-scale Feature Extraction module (see Section~\ref{subsec:msfemodule}) to achieve a higher receptive field crucial to restoring severely blurred image regions. The fourth encoder block fuses the DP left and right features, which the decoder uses to output a single deblurred image. The first convolution layer in the encoder outputs $16$ channel feature map, whereas each encoder block outputs $32$, $64$, $64$, and $128$, channel feature maps, respectively.

The decoder blocks gradually upsample the feature maps while using the skip connections from the Multi-scale Fusion module (see Section~\ref{subsec:msfmodule}). Each decoder block comprises an upsample layer and a $1$x$1$ convolution layer to merge the skip connections. Lastly, the decoder block consists of Conv-LeakyReLU-Conv-LeakyReLU layers, where, unlike encoder blocks, the first convolution layer kernel size is $5$x$5$. Three decoder blocks in MCCNet finally output full resolution deblurred output image.

The skip connections to the decoder blocks consist of multi-scale cross-correlation feature maps. Two Cross-correlation blocks (CC1 and CC2 in Fig.~\ref{fig:proposed_arch}) are employed to perform cross-correlation at the second and third encoder blocks. Due to the high resolution of the input images ($1680$x$1120$), cross-correlation at the first encoder block becomes prohibitively memory expensive. A Multi-scale Fusion module fuses the cross-correlation information from multiple scales, which are then used by the decoder blocks as skip connections. Finally, the last two convolution layers output the full-resolution deblurred output image.

\subsection{Cross-correlation Module}
\label{subsec:ccmodule}
\begin{figure}[t]
	\centering
	\includegraphics[scale=0.34]{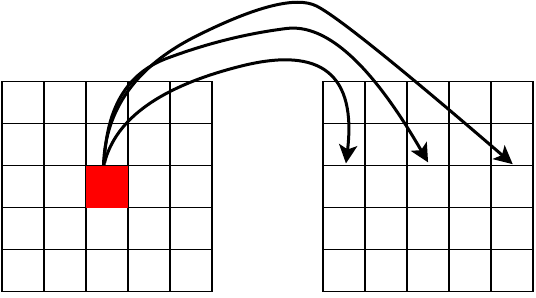}
	\caption{This figure shows how a given pixel in the left DP image is cross-correlated with pixels in the right DP image.}
	\label{fig:sapa}
	\vspace{-4pt}
\end{figure}

\begin{figure}[t]
	\centering
	\includegraphics[scale=0.31]{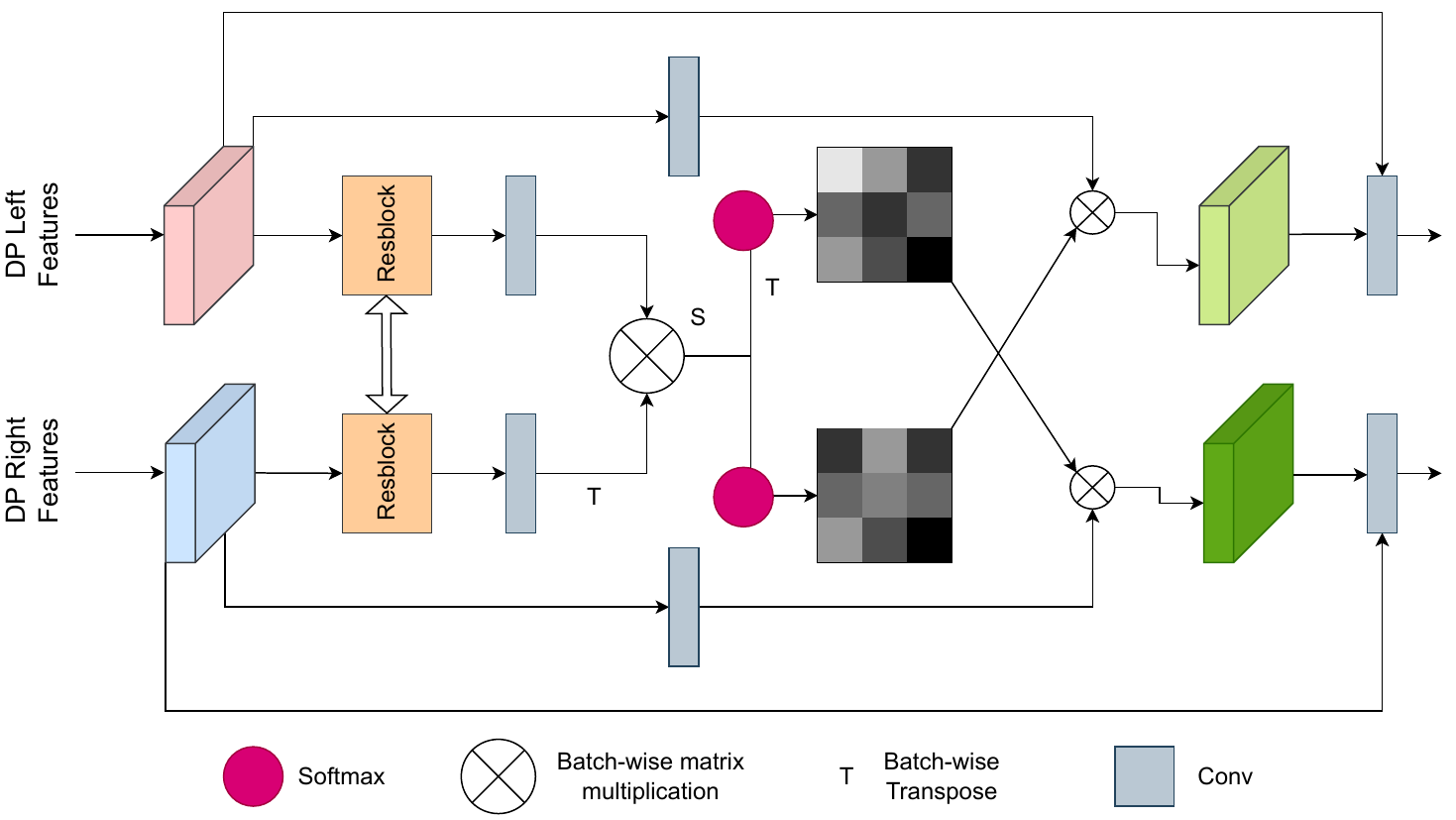}
	\caption{This figure describes the Cross-correlation module used in this work for explicit cross-correlation between the left and right DP images.}
	\label{fig:crosscorr}
	\vspace{-12pt}
\end{figure}

Similar to the case of stereo images, where the correspondence matching needs to be performed only along the epipolar line, in the case of DP images, we need to perform the cross-correlation along the disparity direction. Fig.~\ref{fig:sapa} shows that because of the DP disparity constraint, a given pixel in DP left only needs to be matched with pixels along the same row in the DP right image. Recently, \cite{pam_tpami2022} proposed a modified self-attention mechanism by applying the stereo constraint. Motivated from \cite{pam_tpami2022}, we also use the modified self-attention to compute the cross-correlation between a given pixel in the DP left image with all the pixels along the disparity direction in the DP right image. The detailed structure of the cross-correlation module is shown in Fig.~\ref{fig:crosscorr}. 

As shown in Fig.~\ref{fig:crosscorr}, the Cross-correlation module used in this work takes the DP left and right feature maps and performs cross-correlation across the disparity direction. The computed attention maps are used to scale the input left and right feature maps. Mathematically, given two the DP left and right feature maps $\textbf{A}, \textbf{B} \in \mathbb{R}^{H \text{x} W \text{x} C}$, they are first passed through a residual block with shared parameters, followed by a $1$x$1$ convolution to obtain the query $\textbf{Q} \in \mathbb{R}^{H \text{x} W \text{x} C}$ and key $\textbf{K} \in \mathbb{R}^{H \text{x} W \text{x} C}$ feature maps respectively. Then batch matrix multiplication (considering $H$ as batch size) is performed between $\textbf{Q}$ and $\textbf{K}$ to obtain the matrix $\textbf{S} \in \mathbb{R}^{H \text{x} W \text{x} W}$. Now, to compute the attention map of right feature maps w.r.t. the left feature maps, we take batch transpose (considering $H$ as batch size) of $\textbf{S}$, which is equivalent to obtaining $\textbf{Q}$ from right feature maps and $\textbf{K}$ form left feature maps and computing $\textbf{S}$. $\textbf{S}$ and $\textbf{S}^T$ are passed through the softmax function to obtain the attention scores. The attention scores are multiplied with the input left and right feature maps.

Compared to the mechanism in \cite{pam_tpami2022}, we simplify the Cross-correlation module by excluding the occlusion mask computation since it is not required in the case of DP images. DP images only have a few pixel disparities and thus do not contain significant occlusions.

\subsection{Multi-scale Feature Extraction Module}
\label{subsec:msfemodule}
\begin{figure}[h]
	\centering
	\includegraphics[scale=0.38]{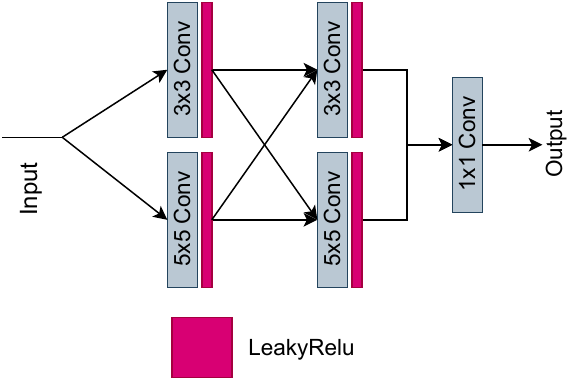}
	\caption{This figure describes the Multi-scale Feature Extraction module.}
	\label{fig:msfe}
	\vspace{-24pt}
\end{figure}

Fig.~\ref{fig:msfe} shows the structure of the Multi-scale Feature Extraction module. This module is used to achieve a larger receptive field crucial to restoring severely blurred image regions. It also performs multi-scale feature fusion, which helps to detect features of different scales and makes the network robust to different blur sizes \cite{wrf_cvprw2021,msrb_eccv2018}. The incoming features are processed by $3$x$3$ and $5$x$5$ convolution layers, respectively. The outputs of previous convolution layers are concatenated and fed to the next $3$x$3$ and $5$x$5$ convolution layers, respectively. Finally, the outputs of previous convolutions layers are merged using a $1$x$1$ convolution layer, which forms the output of this module. This module is inspired by the multi-scale residual block used for image super-resolution task \cite{msrb_eccv2018}.

\subsection{Multi-scale Fusion Module}
\label{subsec:msfmodule}
\begin{figure}[h]
	\centering
	\includegraphics[scale=0.40]{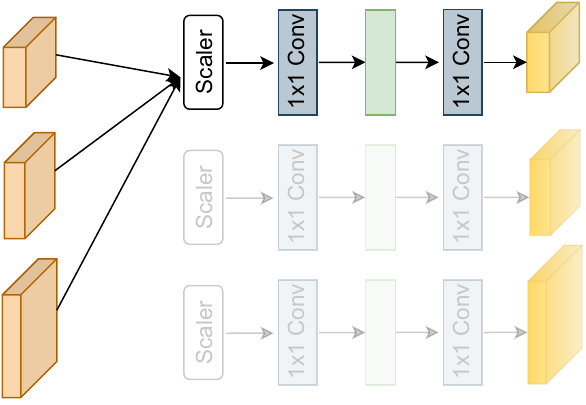}
	\caption{This figure describes the Multi-scale Fusion module, which is used to fuse the encoder and cross-correlation features at different scales.}
	\label{fig:multiscalefusion}
	\vspace{-20pt}
\end{figure}

The Multi-scale Fusion module aims to perform interaction across cross-correlation features of different scales. Fig.~\ref{fig:multiscalefusion} shows the detailed structure of this module. Given three feature maps at different scales, the Multi-scale Fusion module fuses these feature maps at different scales. It outputs three modified feature maps, each corresponding to one of the input feature maps. More specifically, a scaler module is used to scale all the feature maps to the feature map's scale under consideration (e.g., the smallest one in Fig.~\ref{fig:multiscalefusion}). The scaled feature maps are then concatenated and processed by a $1$x$1$ convolution layer, followed by a Multi-scale Feature Extraction module. A $1$x$1$ convolution layer finally outputs the modified feature map corresponding to the one under consideration. The other input feature maps at different scales undergo the same process.

\subsection{Loss Functions}
\label{subsec:loss_functions}
We consider the Charbonnier \cite{charbonnier_cvpr2019}  and MS-SSIM \cite{ssim_tip2004,loss_img_restoration_tci2017} loss functions for training our model MCCNet. Charbonnier loss function is a fully differentiable formulation of the $L_{1}$-norm loss function, which has a discontinuity at the origin. The Charbonnier loss function is formulated as follows:

\begin{equation}
	\sqrt{x^2 + \epsilon^2}
	\label{eq:charbonnier}
\end{equation}

The $\epsilon$ is set to a lower value, such as $1e-3$. This formulation of the $L_{1}$-norm smoothes the curve at the origin, thus, making it fully differentiable. 

As advocated in \cite{loss_img_restoration_tci2017}, we adopt a mix \cite{loss_img_restoration_tci2017} loss function combining the Charbonnier loss and the MS-SSIM loss functions with equal weightage, leading to superior results.

\section{Experimental Setup}
\label{sec:experimentalsetup}

This section describes the dataset details, implementation details, and evaluation criteria used in this work.

\subsection{Dataset}
\label{subsec:dataset}

We use the DPDD dataset from the seminal work by Abuolaim \textit{et al. }\cite{dpdnet_eccv2020}. The DPDD dataset was captured using a Canon DSLR camera. It contains $500$ samples, each containing three images, viz., left and right DP views, and all in-focus images captured using a narrow aperture. Following \cite{dpdnet_eccv2020}, $350$ samples are used for training, $74$ for validation and $76$ samples are used for testing. For training and validation, the images are cropped into patches with size $512$x$512$. In contrast, the evaluation is performed on original $1680$x$1120$ size images.

\subsection{Implementation Details}
\label{subsec:implementationdetails}

The implementation of our work is done using the PyTorch deep learning framework. During training, the parameters of all models were initialized using the strategy proposed by He \textit{et al.} \cite{he_init_iccv2015}. Adam optimization is used with an initial learning rate of $1e-4$, $\beta_{1} = 0.9$, and $\beta_{2} = 0.999$. The learning rate is halved after every $60$ epoch. The total number of training epochs is set to $200$.

\subsection{Evaluation Criteria}
\label{subsec:evaluationcriteria}

For quantitative evaluation purposes, we adopt Peak Signal-to-Noise Ratio (PSNR) \cite{loss_img_restoration_tci2017}, Structural Similarity Index Measurement (SSIM) \cite{ssim_tip2004} and Mean Absolute Error (MAE) \cite{loss_img_restoration_tci2017} metrics. The number of model parameters and FLOPS are also considered for the quantitative evaluation. Additionally, qualitative evaluation and comparison of results are also performed. We used publicly available source codes provided by authors to generate the quantitative and qualitative results for comparison.

\begin{table*}[t]
	\renewcommand{\arraystretch}{1.4}
	\centering
	\setlength{\tabcolsep}{1em}
	\caption{Quantitative comparison against state-of-the-art methods. Compared to other methods, MCCNet removes defocus deblurring more effectively and efficiently. Note: DPDNet+ and RDPD+ were additionally trained with synthetic DP data in \cite{rdpd_iccv2021}.}	
	\label{tab:quantitaiveresults}
	\vspace{-8pt}
	\scalebox{0.8}{
		\begin{tabular}{c|c|c|c|c|c|c}
			\hline			
			\textbf{Method} & \textbf{Year} & \textbf{PSNR} $\uparrow$ & \textbf{SSIM} $\uparrow$ & \textbf{MAE} $\downarrow$ & \textbf{Params} $\downarrow$ & \textbf{FLOPS} $\downarrow$\\
			
			\hline \hline
			JNB~\cite{jnb_cvpr2015}		& CVPR 2020 & 23.84 & 0.715 & 0.048 & - & - \\ 
			
			EBDB~\cite{ebdb_tip2017}	& CVPR 2020 & 23.45 & 0.683 & 0.049 & - & - \\ 		
			
			DMENet~\cite{dme_cvpr2019}	& CVPR 2020 & 23.55 & 0.720 & 0.049 & 26.94M & - \\ 
			
			IFAN~\cite{iter_filter_adapt_cvpr2021}	& CVPR 2021 & 25.37 & 0.789 & 0.039 & 10.48M & 725.8G \\
			
			KPAC~\cite{kpac_iccv2021}	& ICCV 2021 & 25.22 & 0.774 & 0.040 & 1.58M & 730.91G \\
			
			MDP~\cite{single_defocus_deblur_wacv2022}	& WACV 2022 & 25.35 & 0.763 & 0.040 & 46.8M & 7751.27G \\
			
			DPDNet~\cite{dpdnet_eccv2020}	& ECCV 2020 & 25.13 & 0.786 & 0.041 & 34.52M & 1883.74G \\
			
			DPDNet+~\cite{rdpd_iccv2021}	& ECCV 2020 & 25.12 & 0.784 & 0.042 & 34.52M & 1883.74G \\ 
			
			RDPD+~\cite{rdpd_iccv2021}	& ICCV 2021 & 25.39 & 0.772 & 0.040 & 27.51M & 612.05G \\
			
			\hline
			
			MCCNet	& - & 25.85 & 0.802 & 0.037 & 5.52M & 978.79G \\
			
			\hline
			
		\end{tabular}
	}
	\vspace{-14pt}
\end{table*}

\section{Results and Discussion}
\label{sec:results}

Tab.~\ref{tab:quantitaiveresults} show the quantitative results of MCCNet against the state-of-the-art defocus deblurring methods in the literature. Similar to DPDNet \cite{dpdnet_eccv2020} and other methods in the literature, we include JNB \cite{jnb_cvpr2015}, and EBDB \cite{ebdb_tip2017}, which use traditional hand-crafted features to estimate a defocus map which is used to perform non-blind deconvolution. DMENet \cite{dme_cvpr2019} uses a deep neural network to only estimate a defocus map. Also, DPDNet+ and RDPD+ \cite{rdpd_iccv2021} were trained on an additional synthetic DP dataset.

It can be seen that MCCNet obtains the highest PSNR, SSIM values and lowest MAE value compared to the other state-of-the-art methods. MCCNet obtains a $1.81\%$ increase ($0.46$ dB increase) in PSNR value compared to the second best PSNR value obtained by RDPD+\cite{rdpd_iccv2021}, whereas $1.65\%$ increase in SSIM value compared to the second best SSIM value obtained by IFAN \cite{iter_filter_adapt_cvpr2021}. Regarding MAE, MCCNet obtains a $5.13\%$ decrease compared to the second best MAE value obtained by IFAN \cite{iter_filter_adapt_cvpr2021}.

Furthermore, the number of parameters of MCCNet is the second lowest after IFAN \cite{iter_filter_adapt_cvpr2021}. Compared to DPDNet \cite{dpdnet_eccv2020} and RDPD+\cite{rdpd_iccv2021}, MCCNet parameters are lesser by at least $80\%$. The FLOPS of MCCNet are $25-30\%$ higher than \cite{iter_filter_adapt_cvpr2021,kpac_iccv2021,rdpd_iccv2021} methods, whereas it is $2$x lesser than DPDNet \cite{dpdnet_eccv2020} and almost $8$x lesser than \cite{single_defocus_deblur_wacv2022}. 

The quantitative results and comparison reveals that MCCNet generates defocus deblurring more effectively and efficiently than the state-of-the-art.

\begin{figure}[t]
	\centering
	\includegraphics[width=0.16\textwidth]{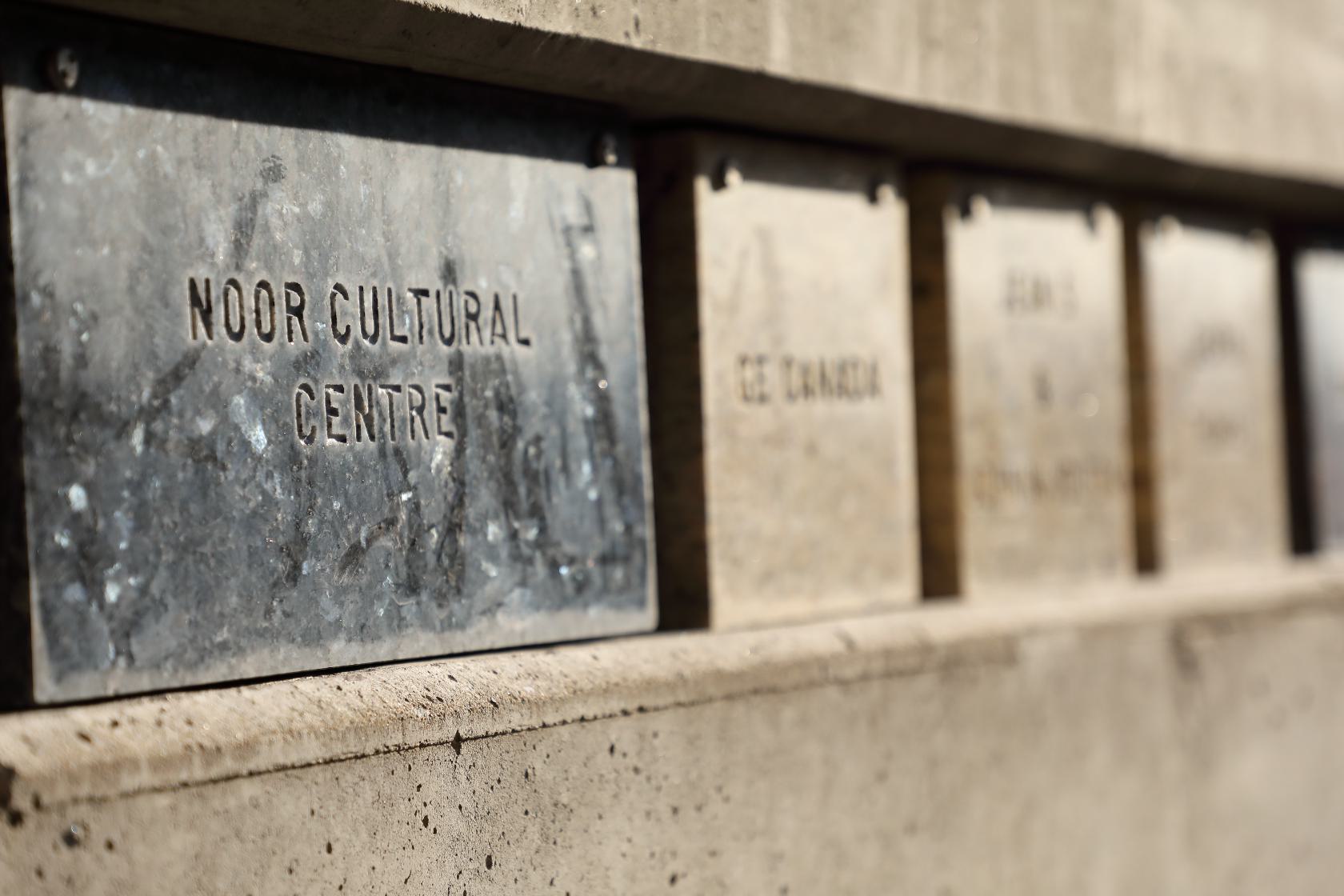}
	\includegraphics[width=0.16\textwidth]{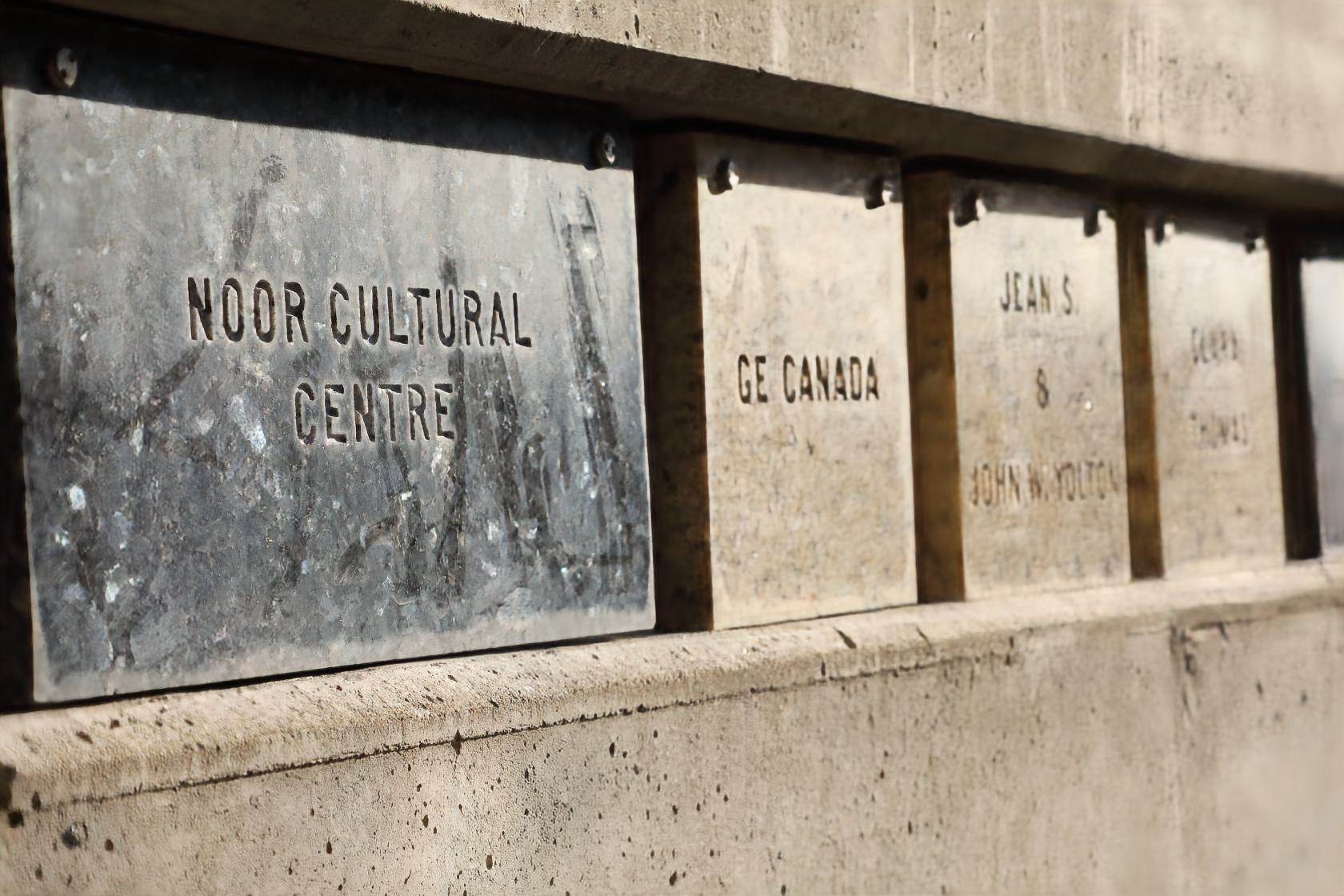}
	\includegraphics[width=0.16\textwidth]{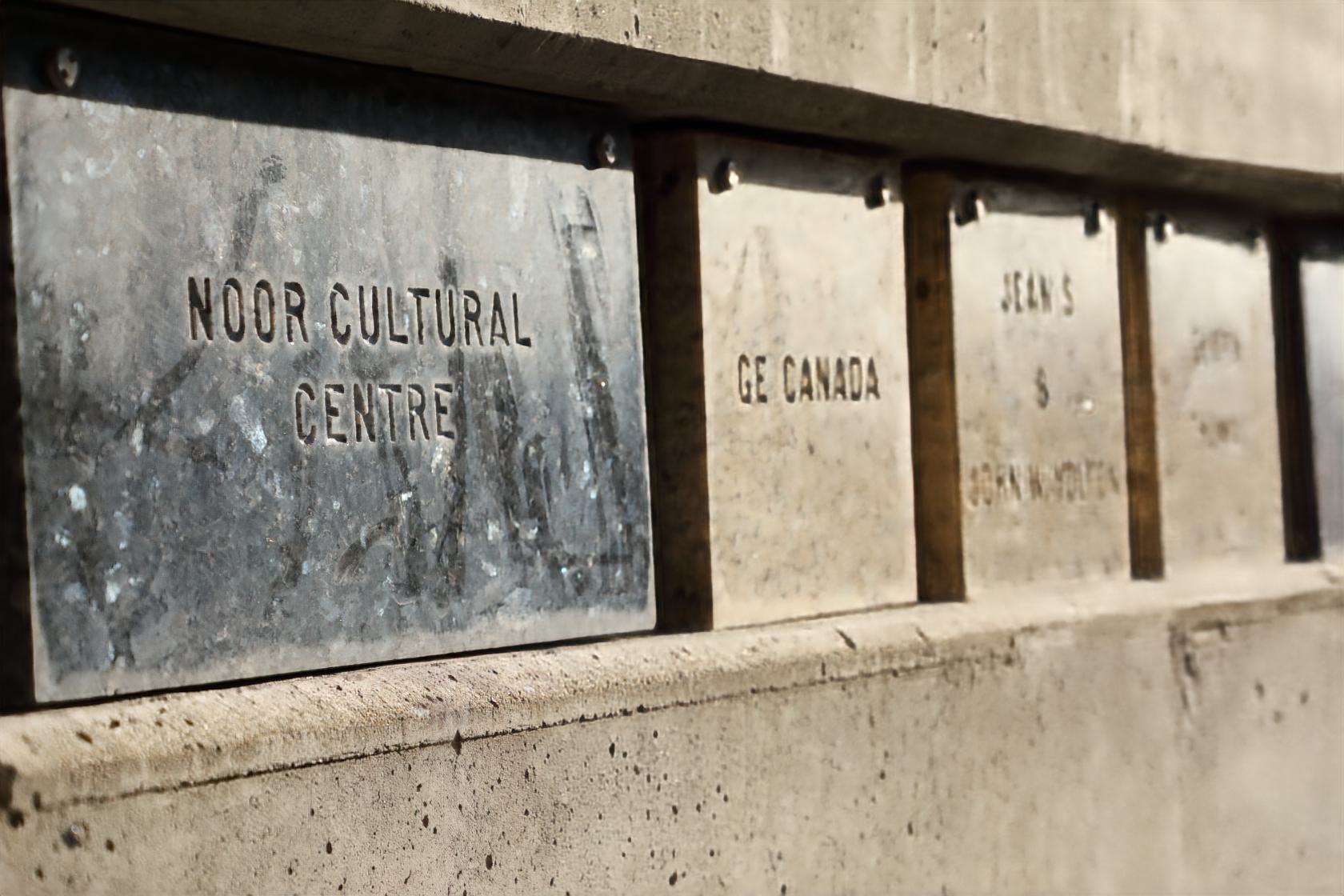}
	\includegraphics[width=0.16\textwidth]{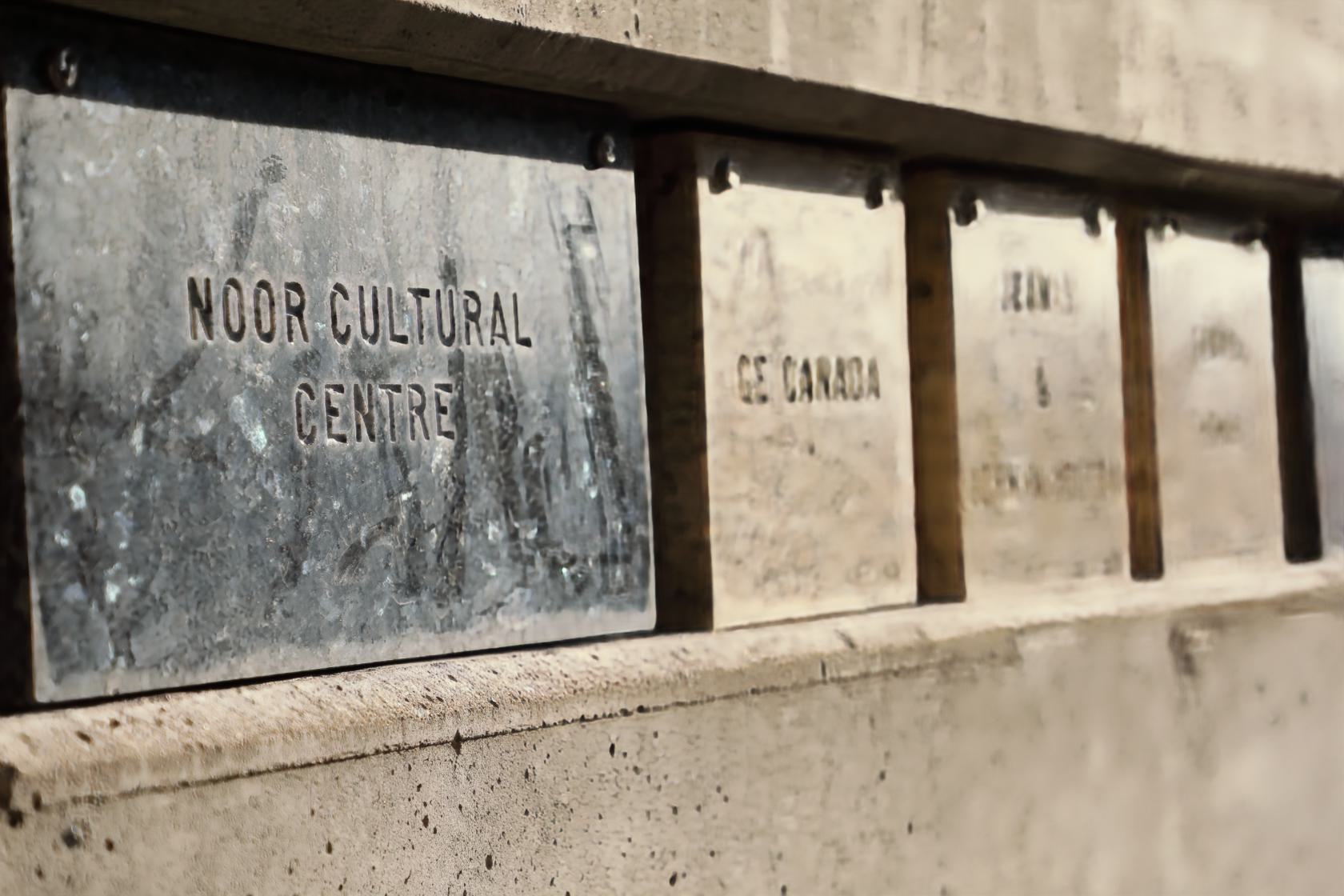}
	\includegraphics[width=0.16\textwidth]{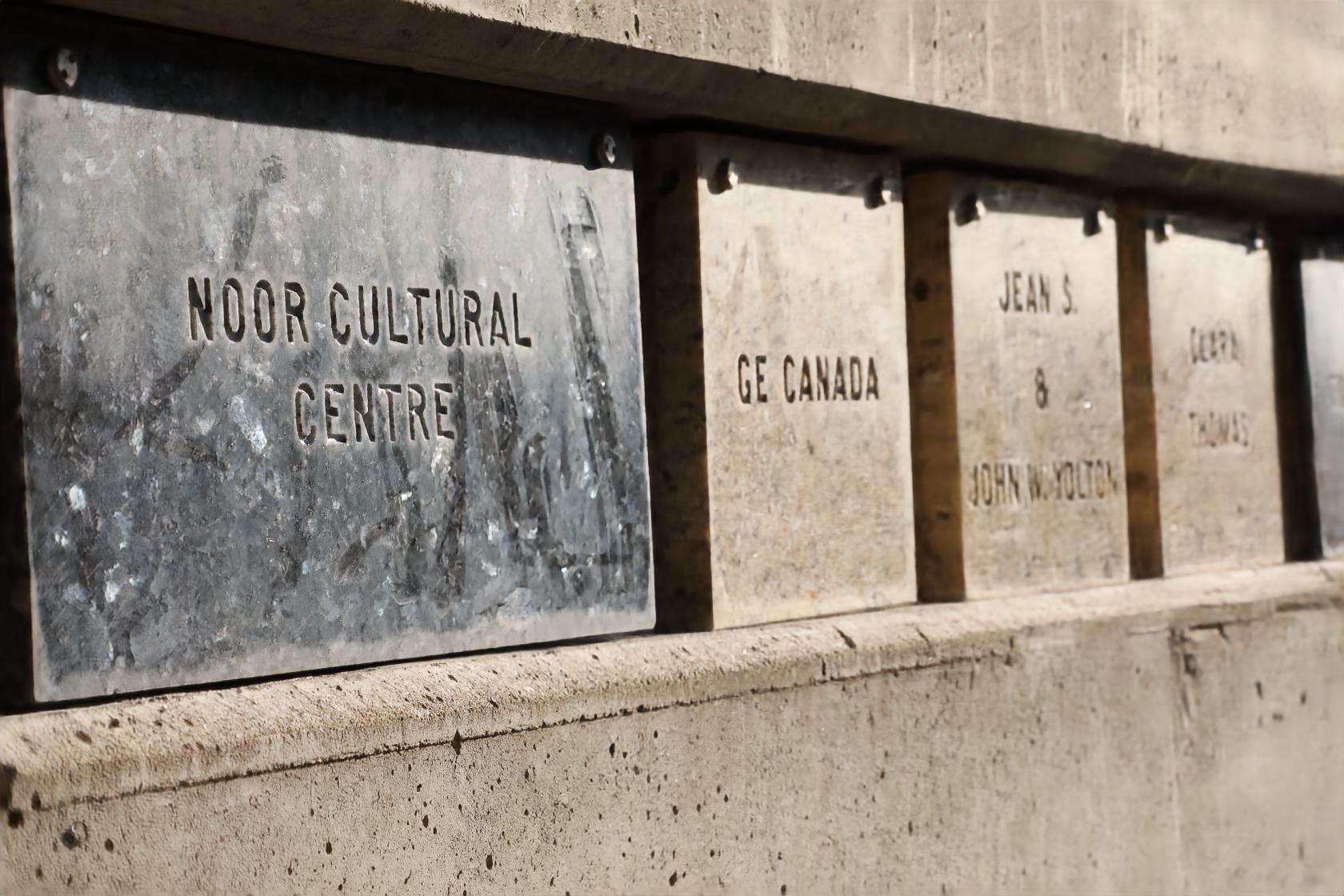}
	\includegraphics[width=0.16\textwidth]{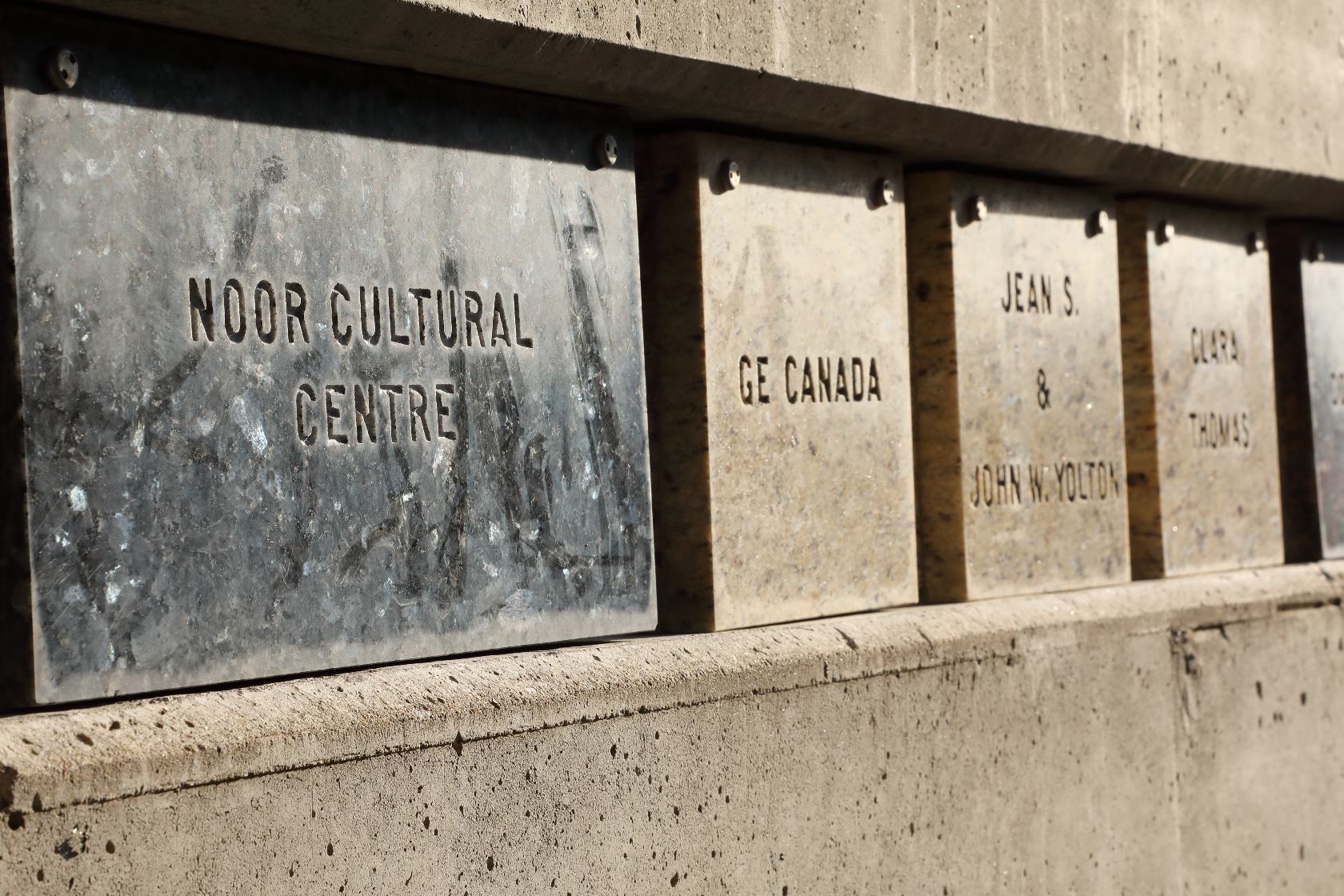}
	
	\frame{\includegraphics[width=0.16\textwidth]{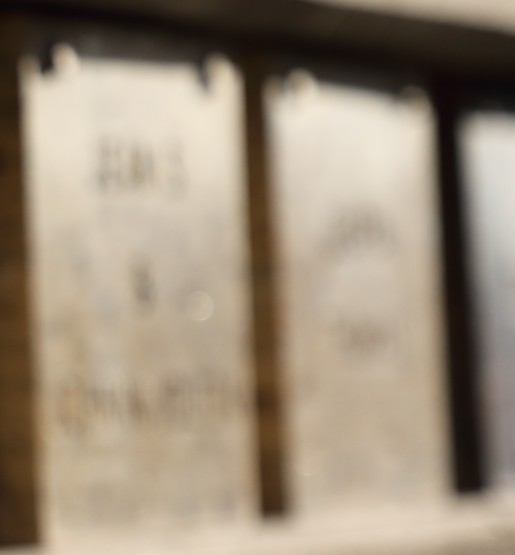}}
	\frame{\includegraphics[width=0.16\textwidth]{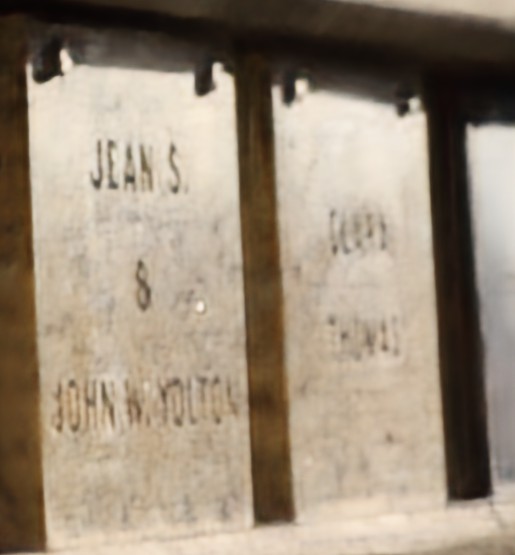}}
	\frame{\includegraphics[width=0.16\textwidth]{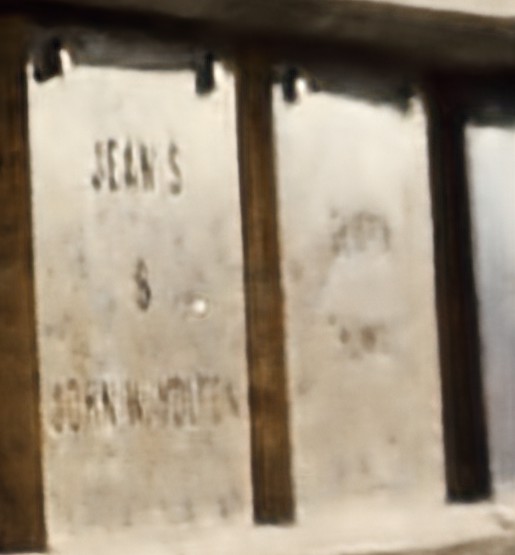}}
	\frame{\includegraphics[width=0.16\textwidth]{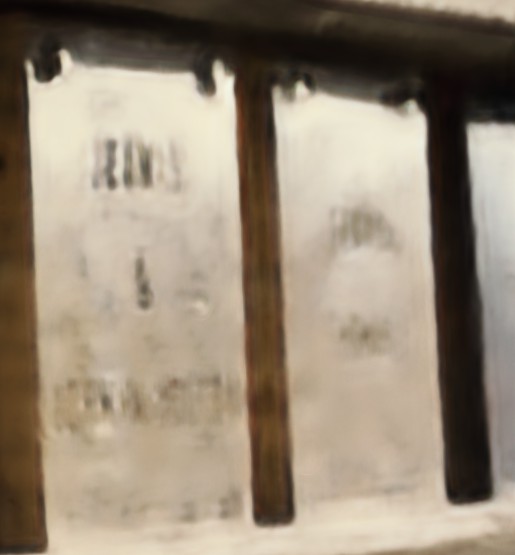}}
	\frame{\includegraphics[width=0.16\textwidth]{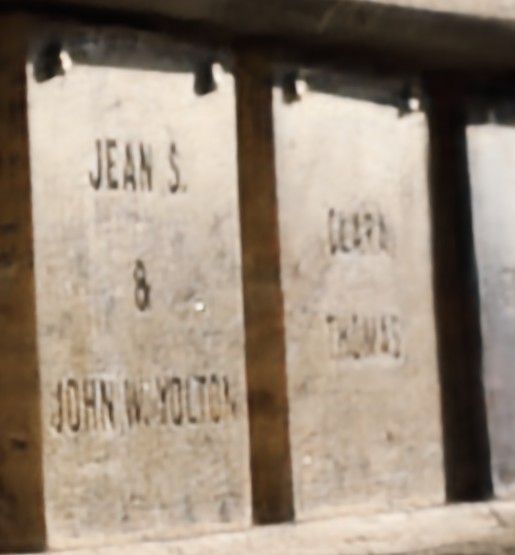}}
	\frame{\includegraphics[width=0.16\textwidth]{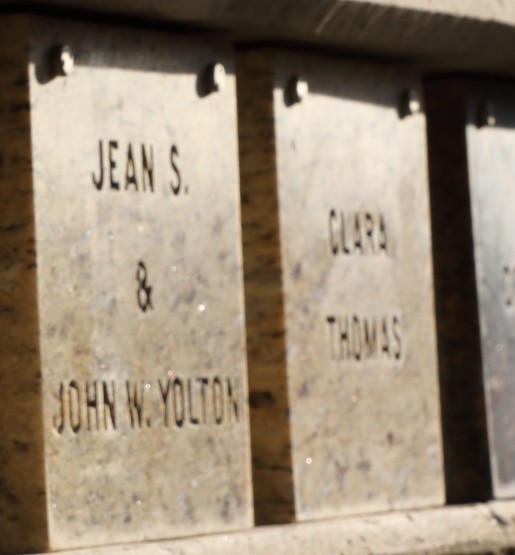}}
	
	\vspace{4pt}
	
	\includegraphics[width=0.16\textwidth]{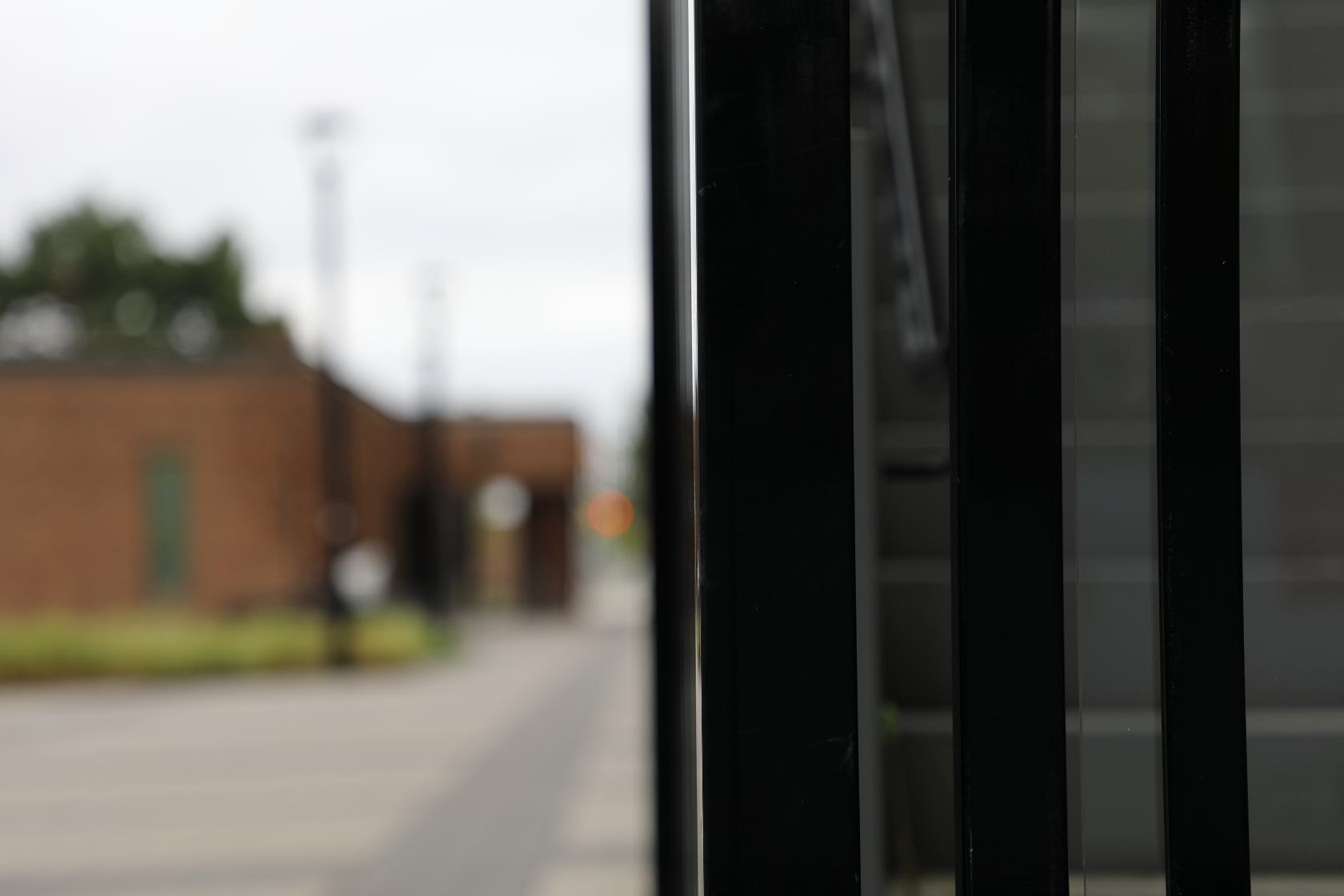}
	\includegraphics[width=0.16\textwidth]{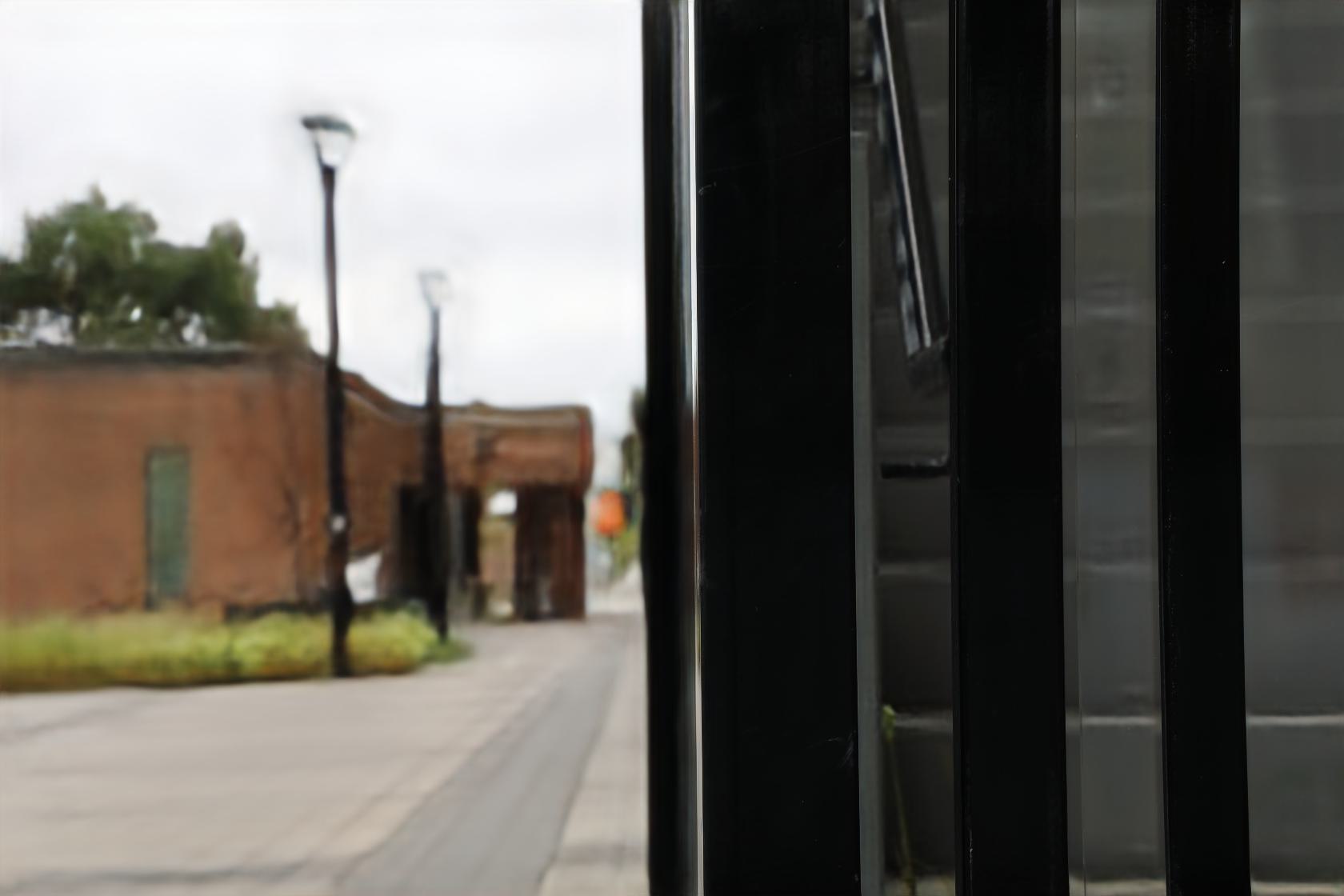}
	\includegraphics[width=0.16\textwidth]{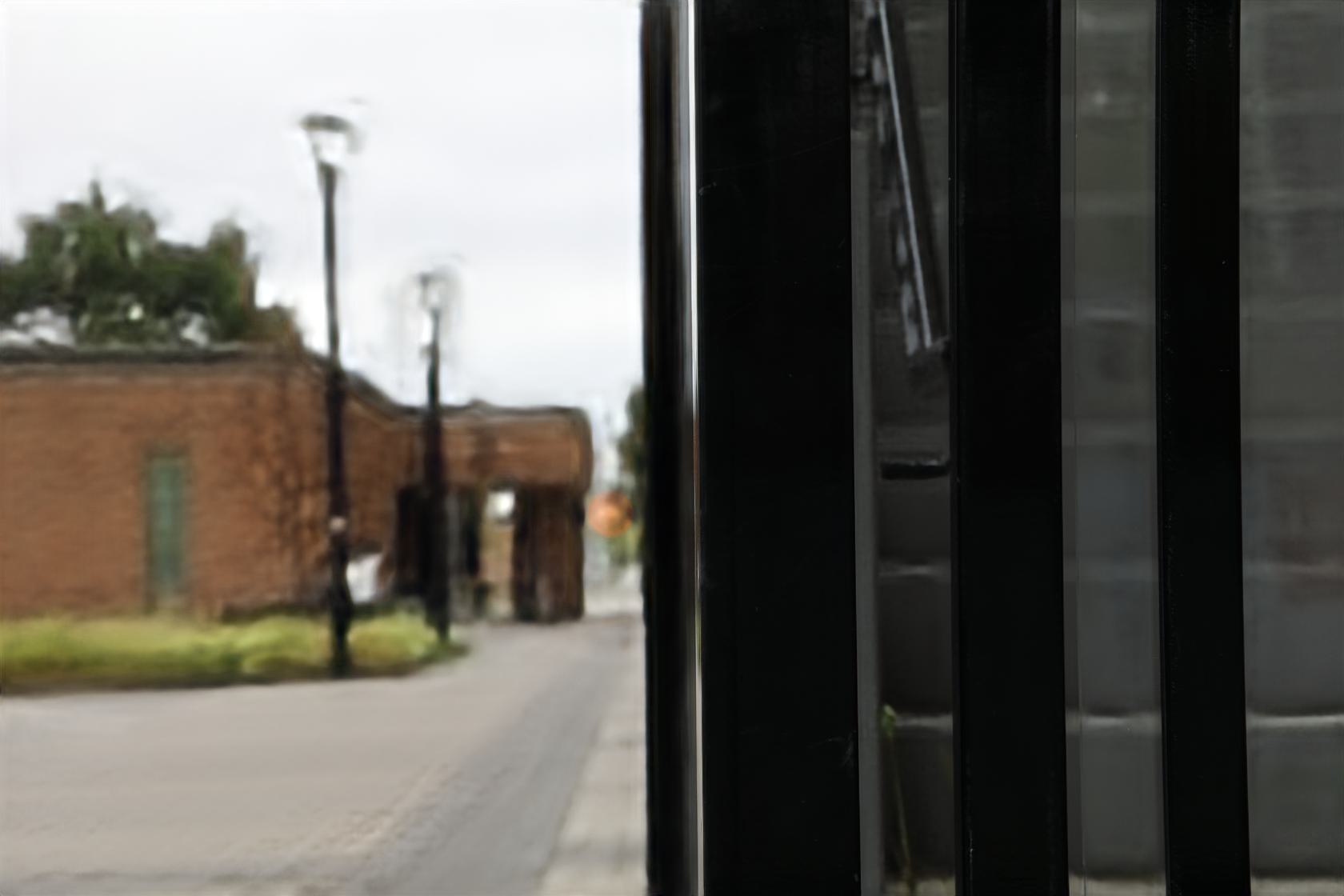}
	\includegraphics[width=0.16\textwidth]{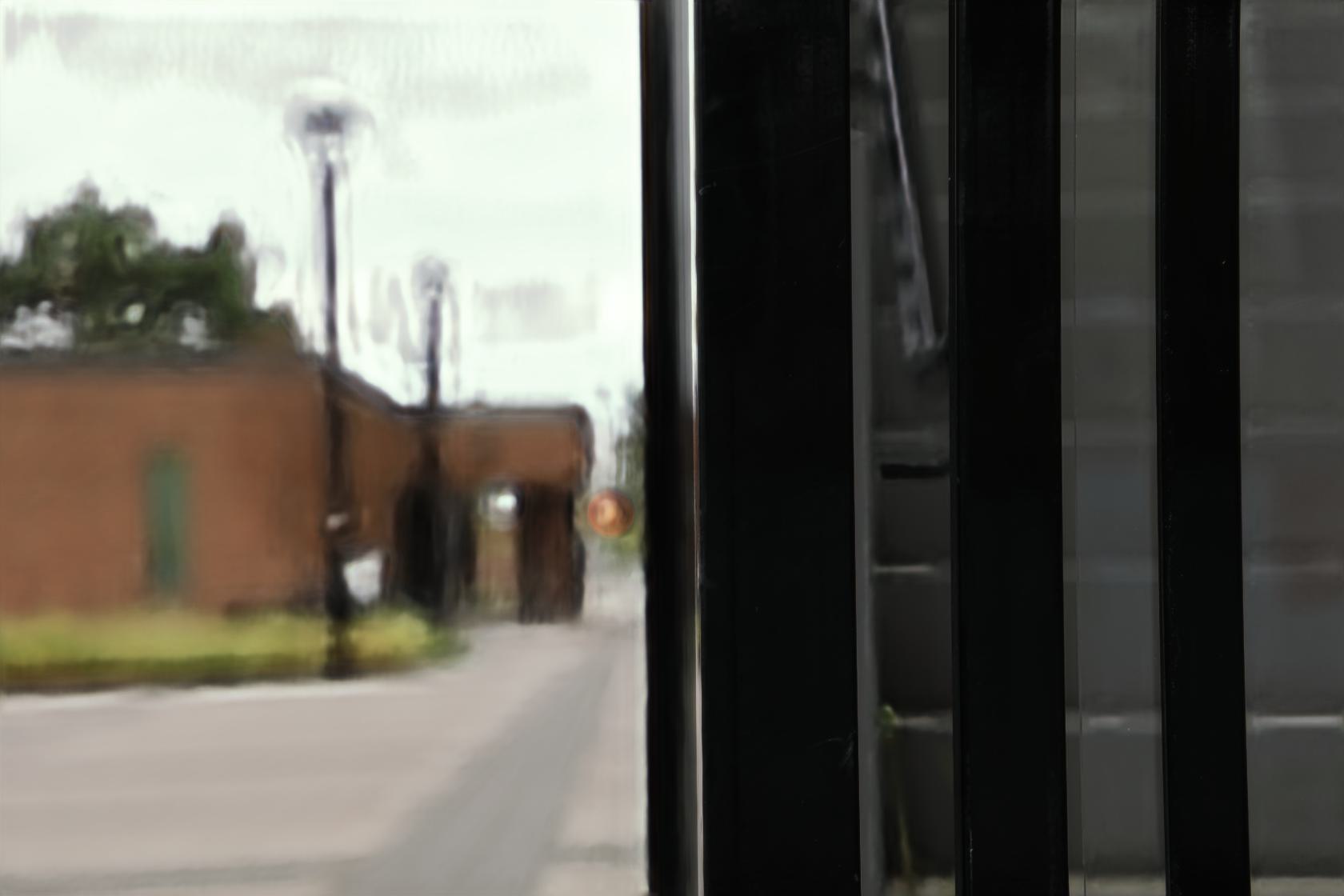}
	\includegraphics[width=0.16\textwidth]{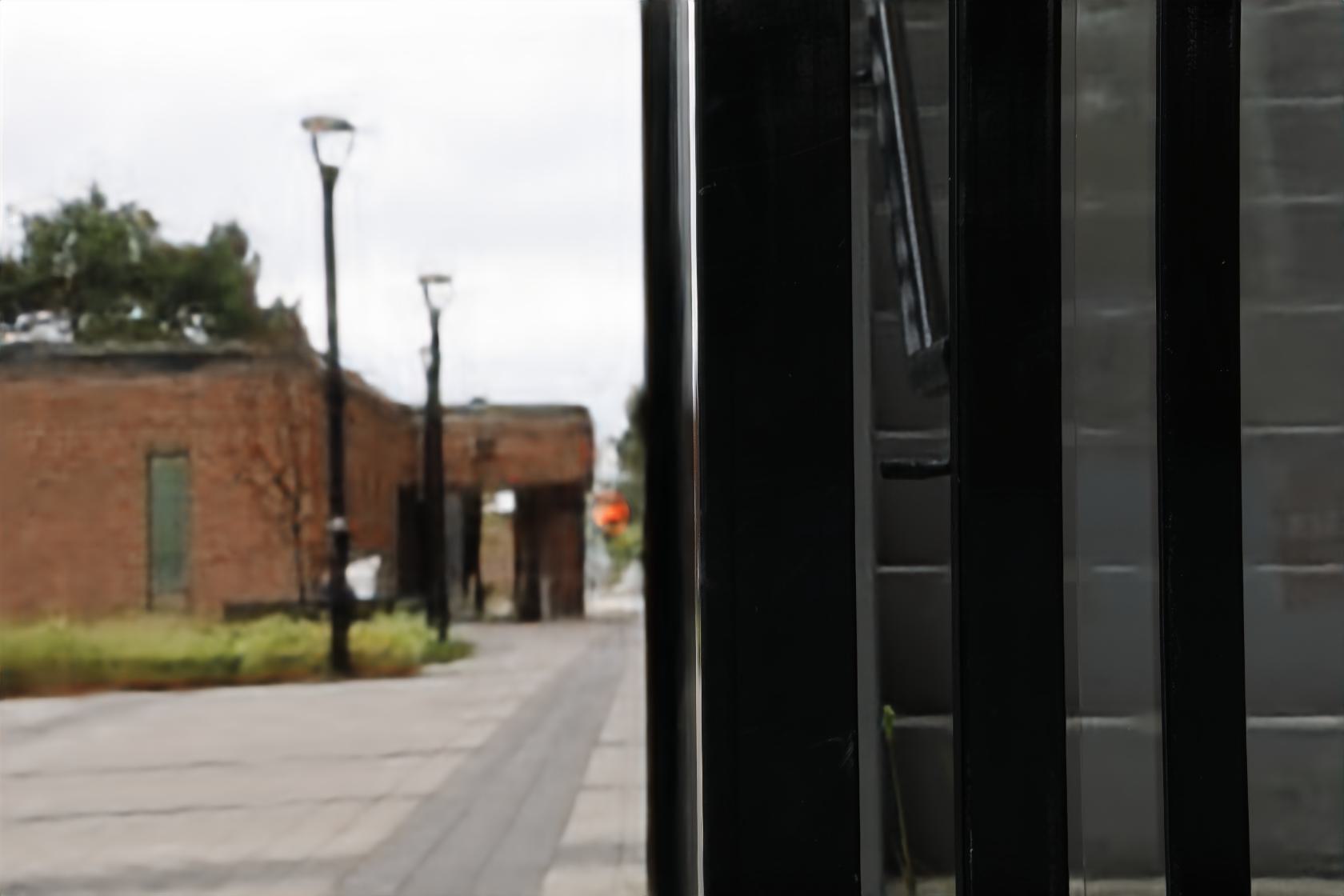}
	\includegraphics[width=0.16\textwidth]{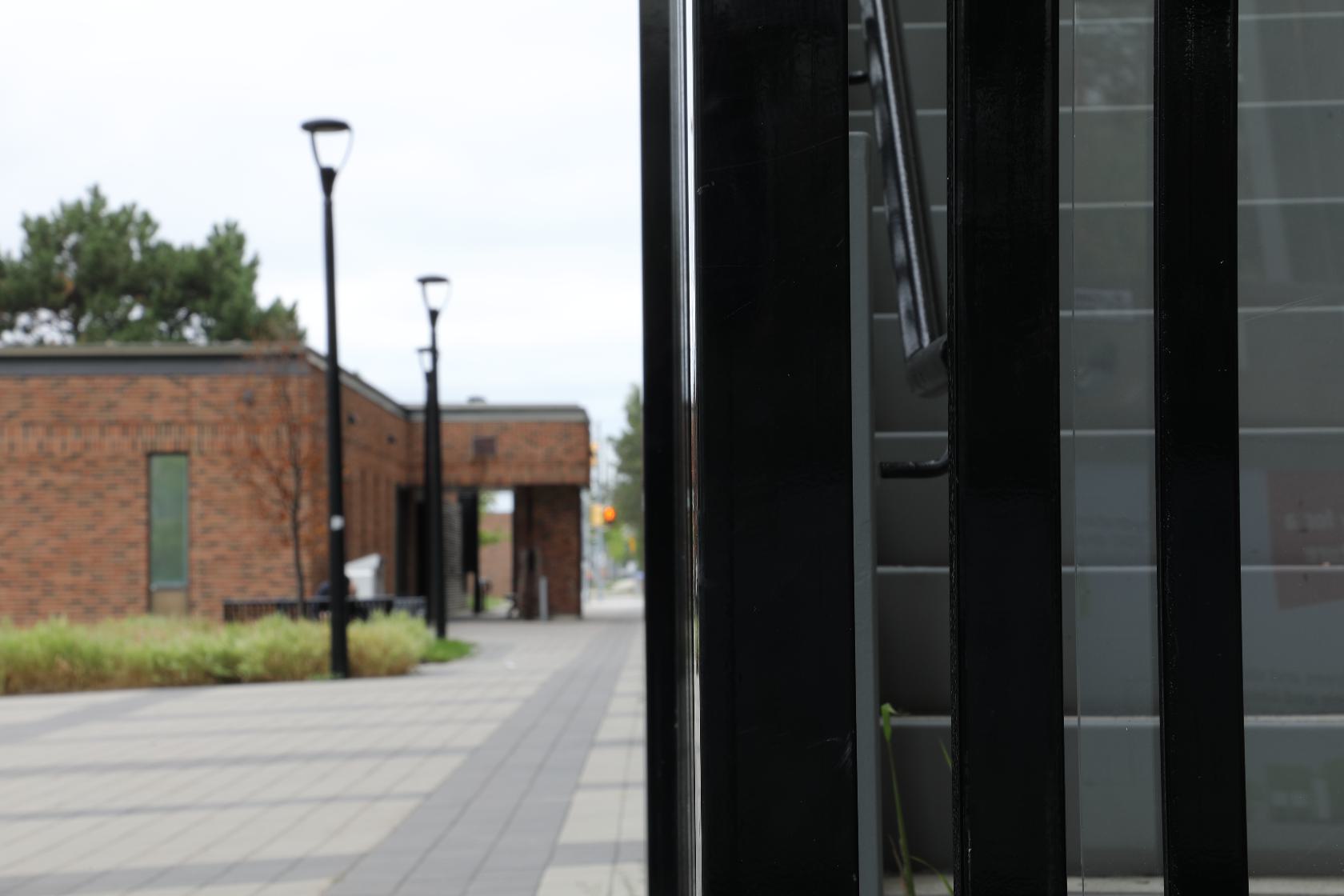}
	
	\frame{\includegraphics[width=0.16\textwidth]{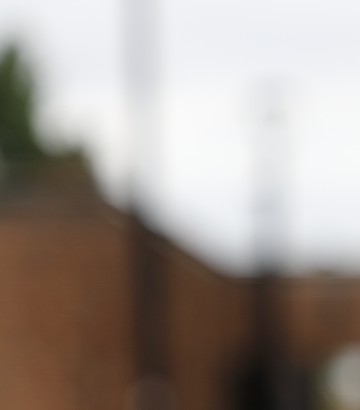}}
	\frame{\includegraphics[width=0.16\textwidth]{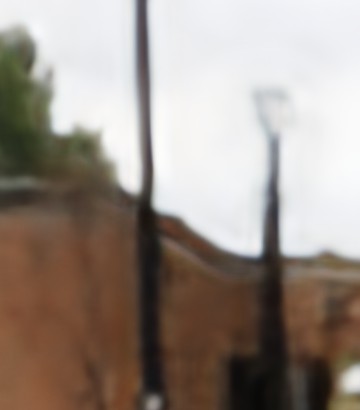}}
	\frame{\includegraphics[width=0.16\textwidth]{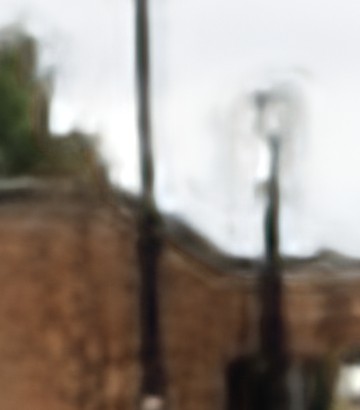}}
	\frame{\includegraphics[width=0.16\textwidth]{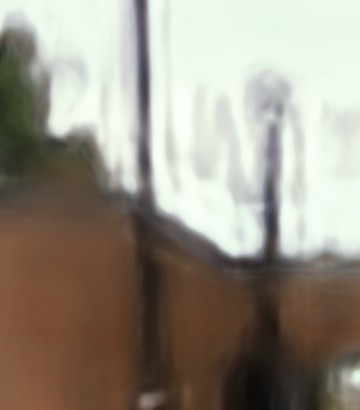}}
	\frame{\includegraphics[width=0.16\textwidth]{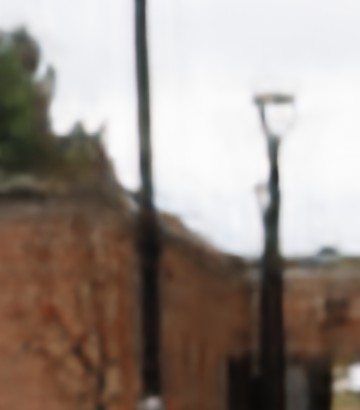}}
	\frame{\includegraphics[width=0.16\textwidth]{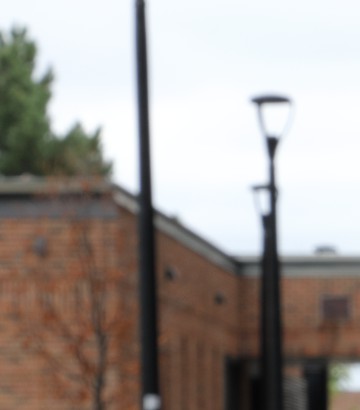}}
	
	\vspace{4pt}
	
	\includegraphics[width=0.16\textwidth]{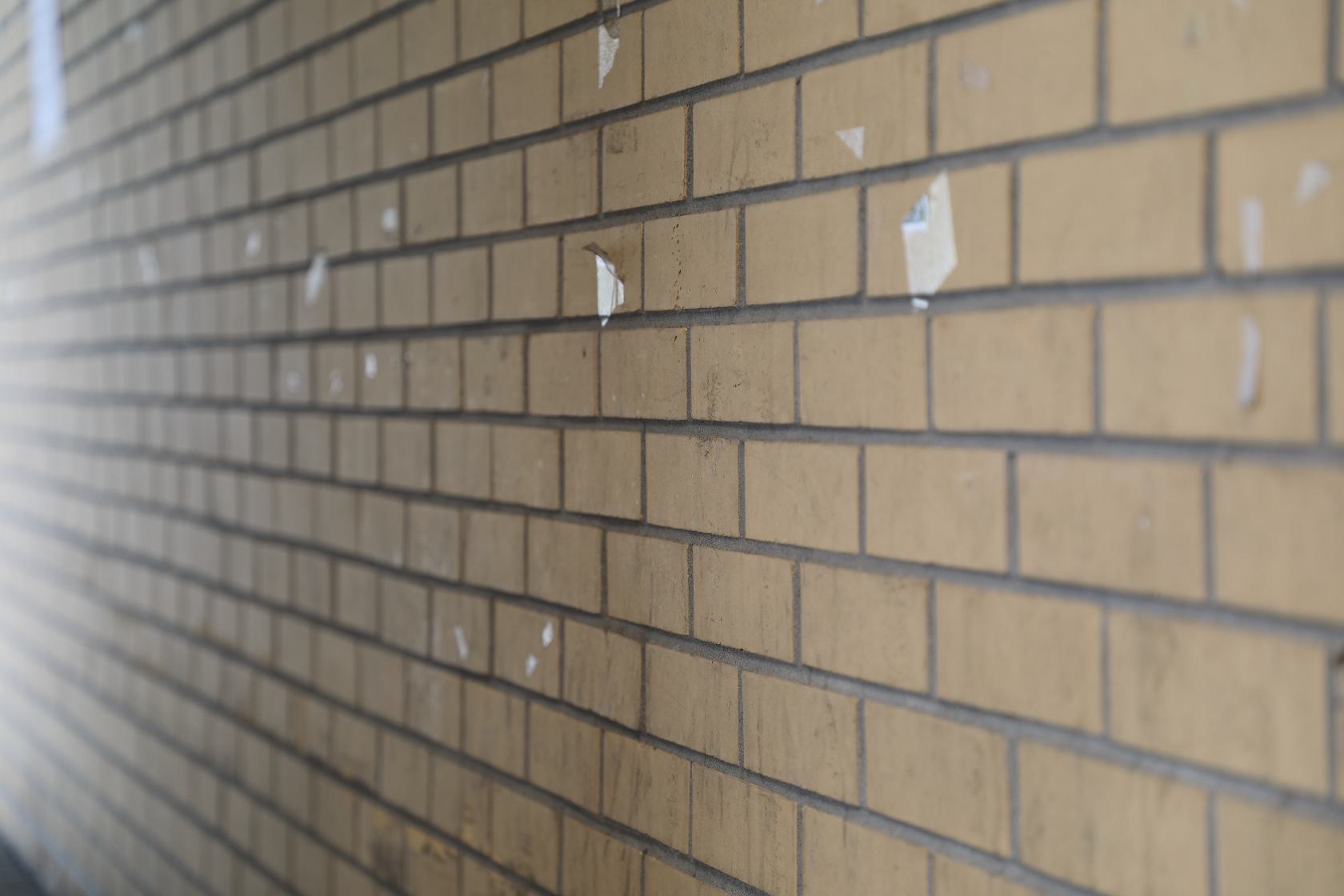}
	\includegraphics[width=0.16\textwidth]{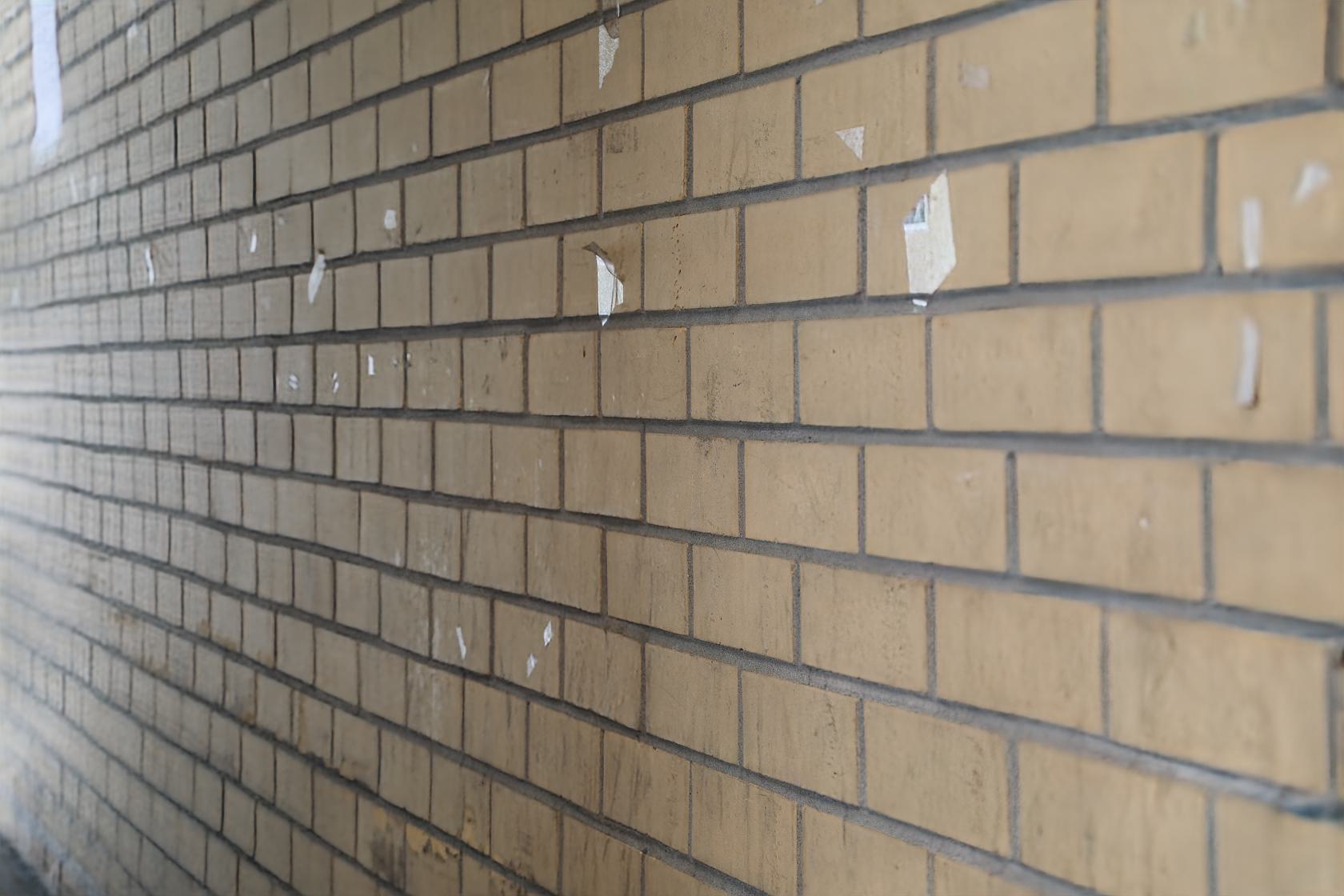}
	\includegraphics[width=0.16\textwidth]{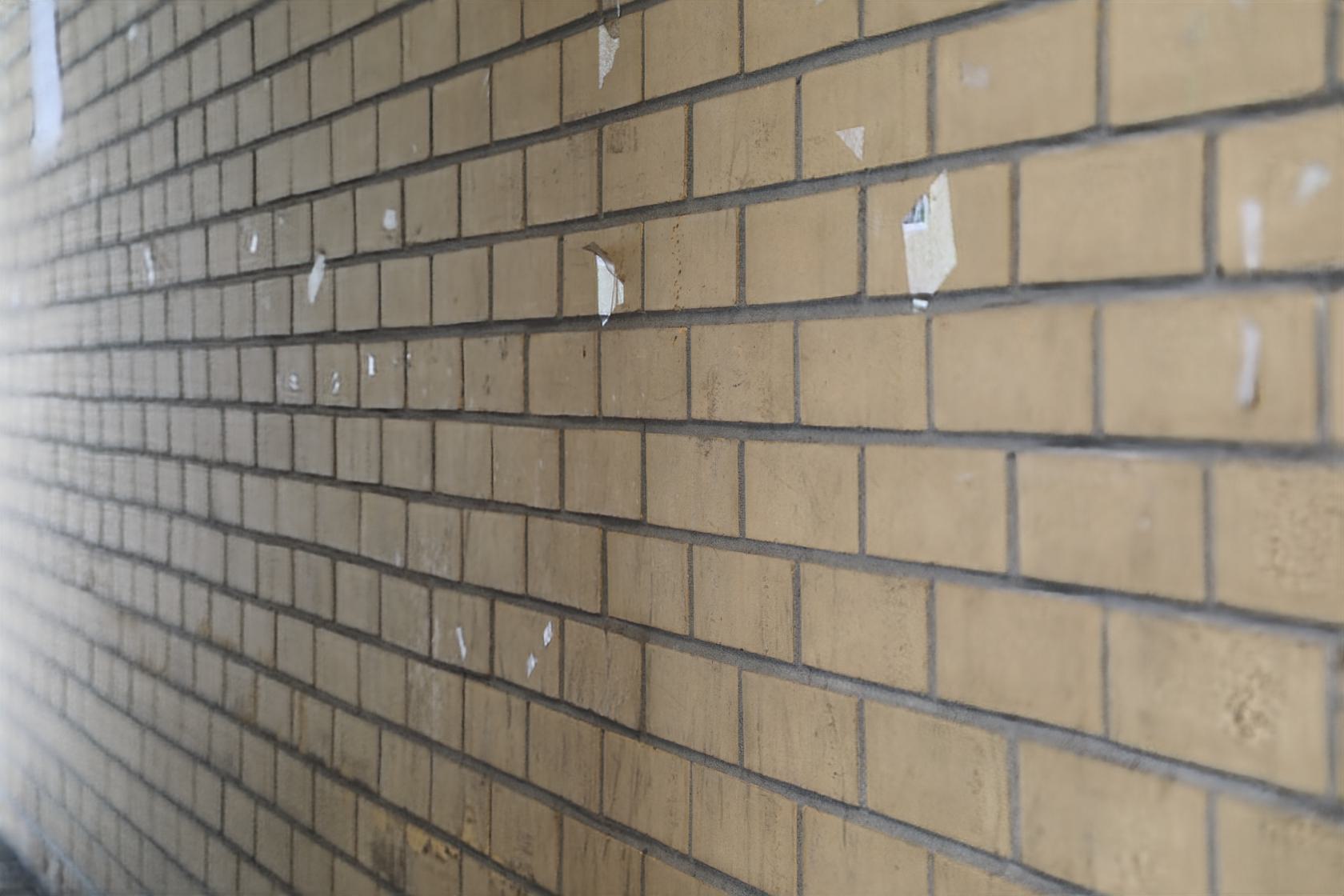}
	\includegraphics[width=0.16\textwidth]{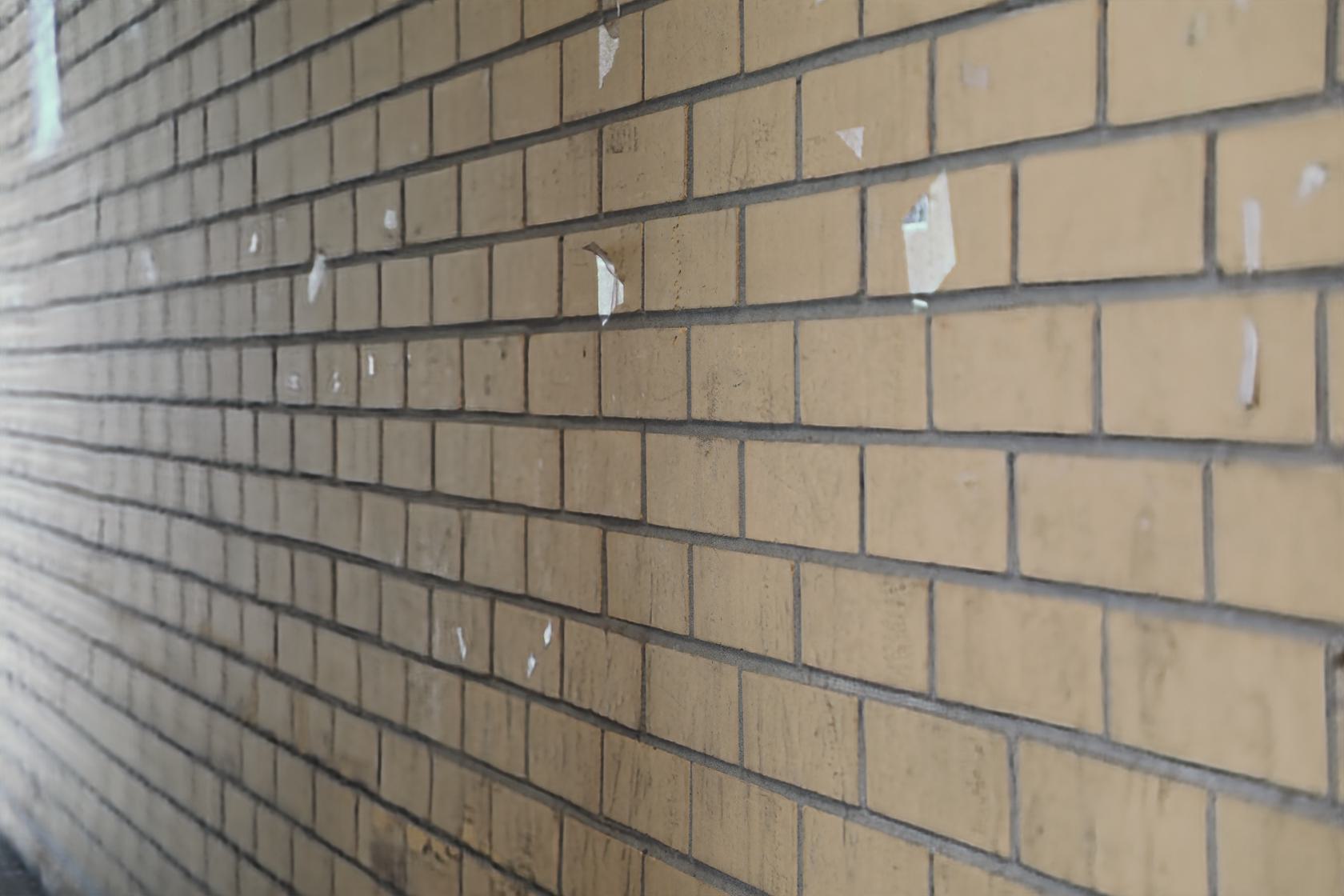}
	\includegraphics[width=0.16\textwidth]{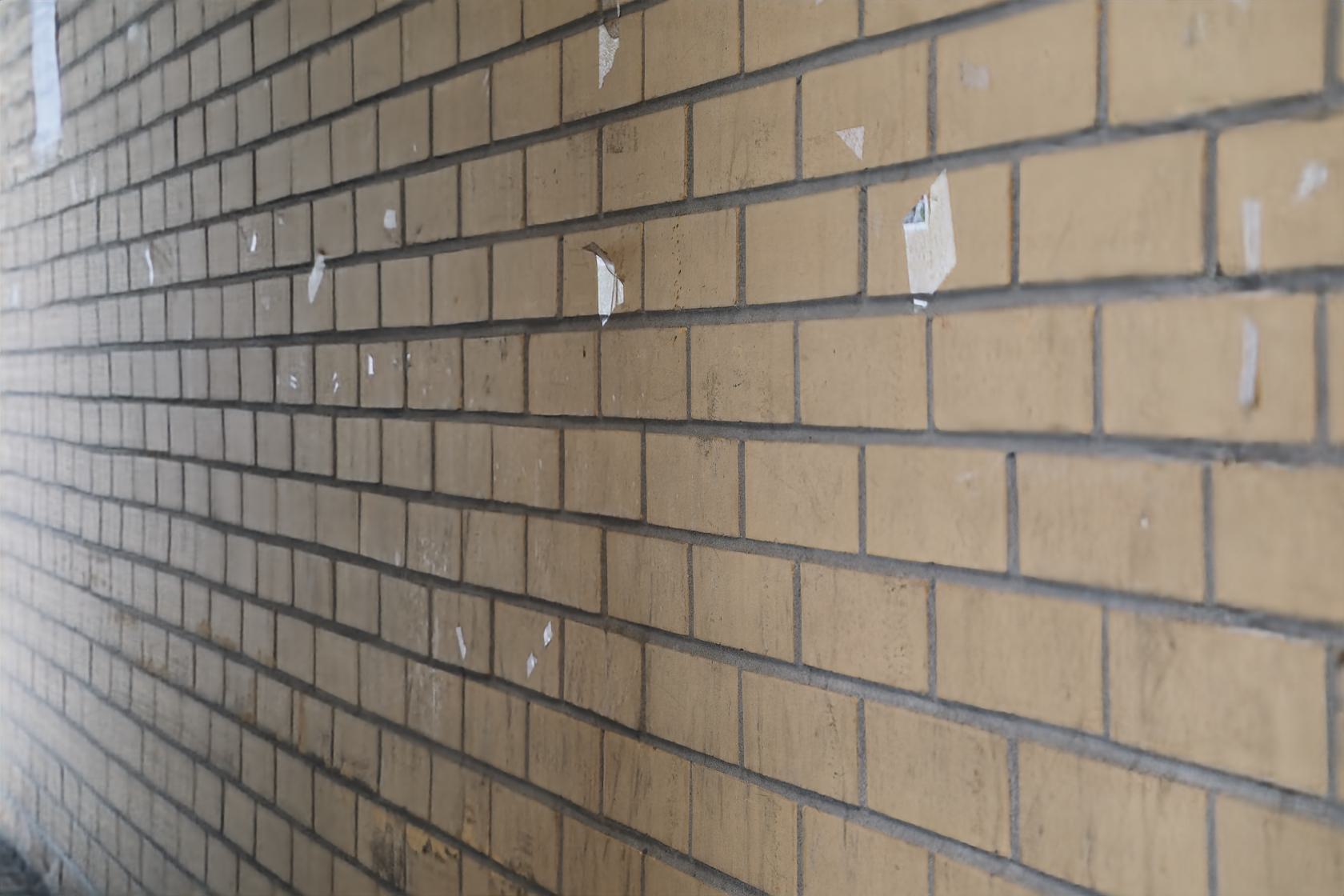}
	\includegraphics[width=0.16\textwidth]{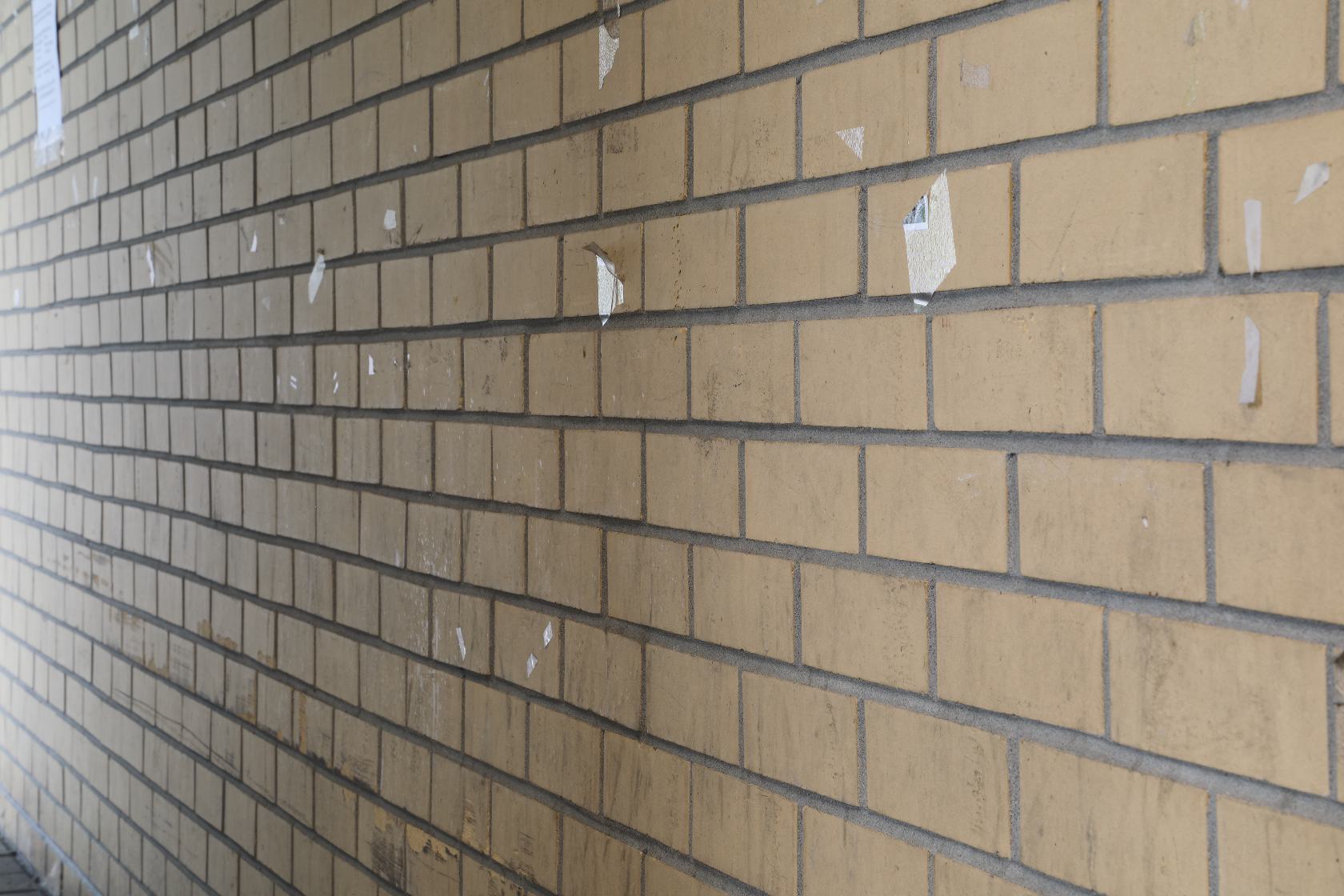}
	
	\frame{\includegraphics[width=0.16\textwidth]{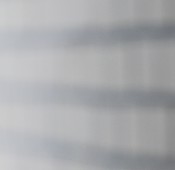}}
	\frame{\includegraphics[width=0.16\textwidth]{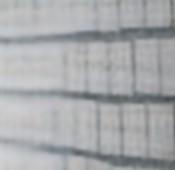}}
	\frame{\includegraphics[width=0.16\textwidth]{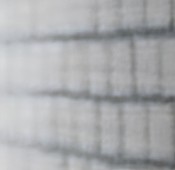}}
	\frame{\includegraphics[width=0.16\textwidth]{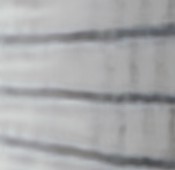}}
	\frame{\includegraphics[width=0.16\textwidth]{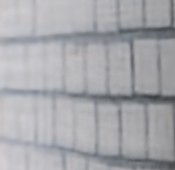}}
	\frame{\includegraphics[width=0.16\textwidth]{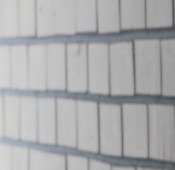}}
	
	\caption{Qualitative comparisons (part 1) of MCCNet against state-of-the-art methods. We include some zoomed and cropped regions for each test image to demonstrate the clear advantages of MCCNet. From left to right: Input, DPDNet \cite{dpdnet_eccv2020}, RDPD+\cite{rdpd_iccv2021}, MDP \cite{single_defocus_deblur_wacv2022}, MCCNet, Ground-truth.}
	\label{fig:qual_comp1}
	\vspace{-8pt}
\end{figure}

\begin{figure}[th]
	\centering
	\includegraphics[width=0.16\textwidth]{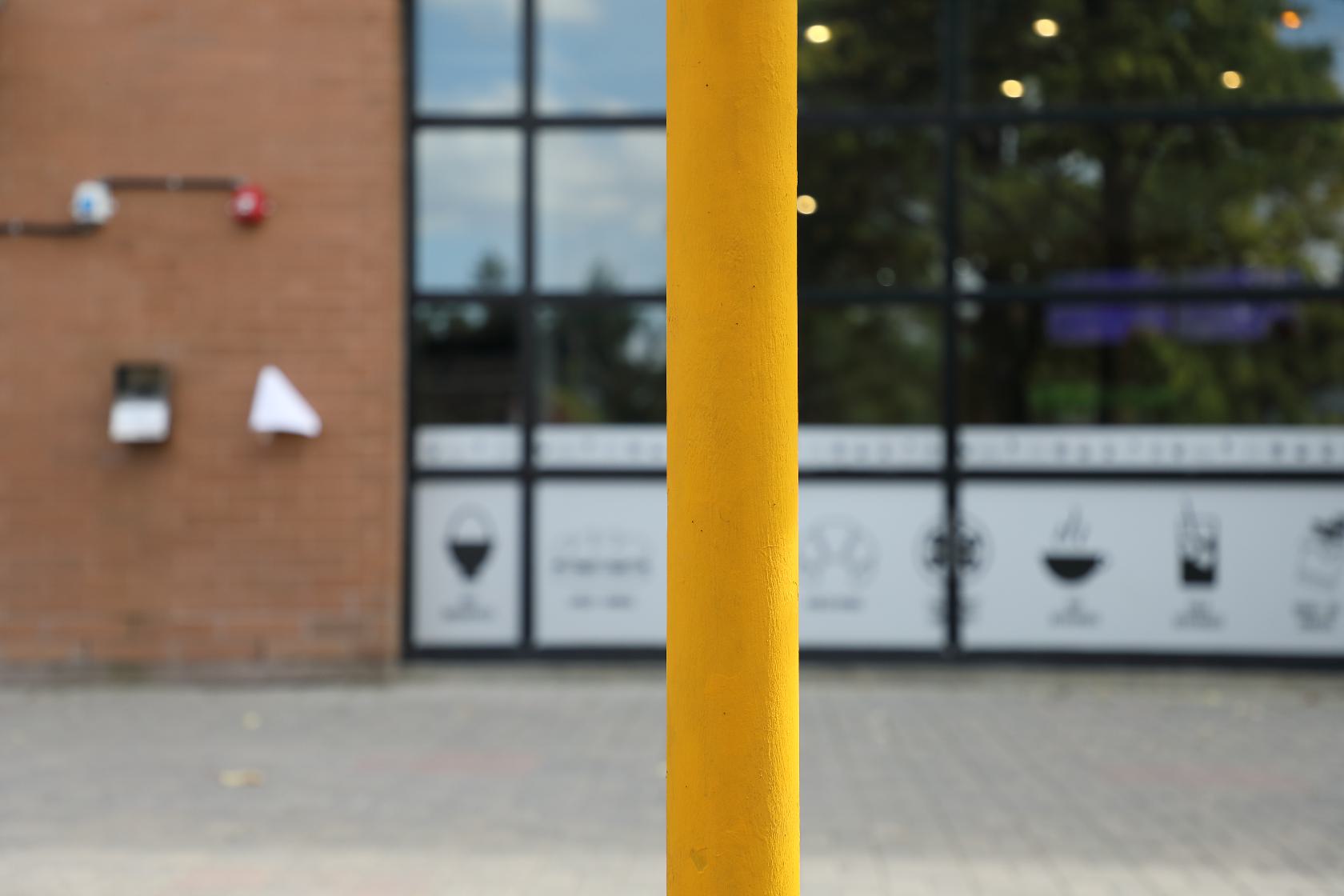}
	\includegraphics[width=0.16\textwidth]{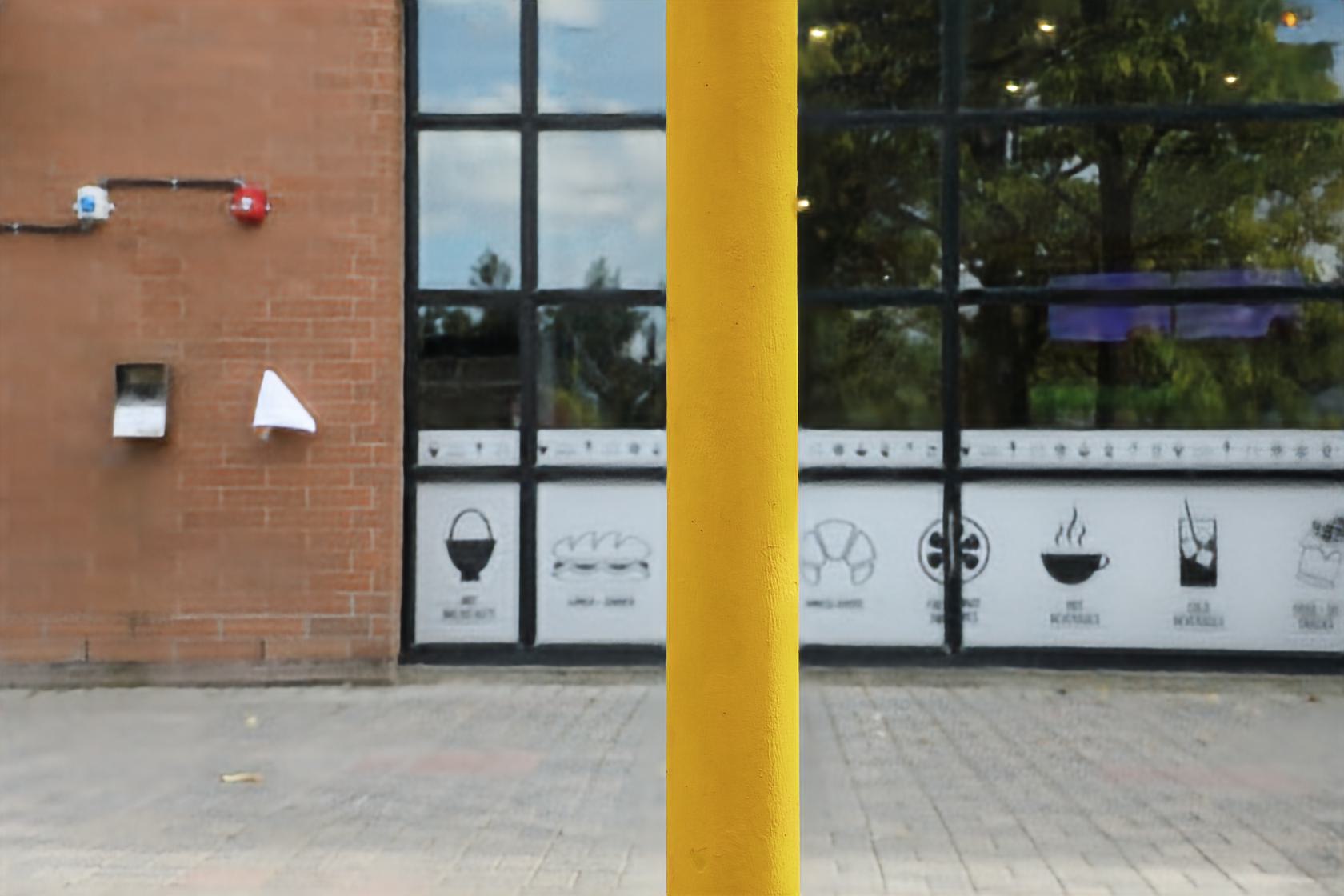}
	\includegraphics[width=0.16\textwidth]{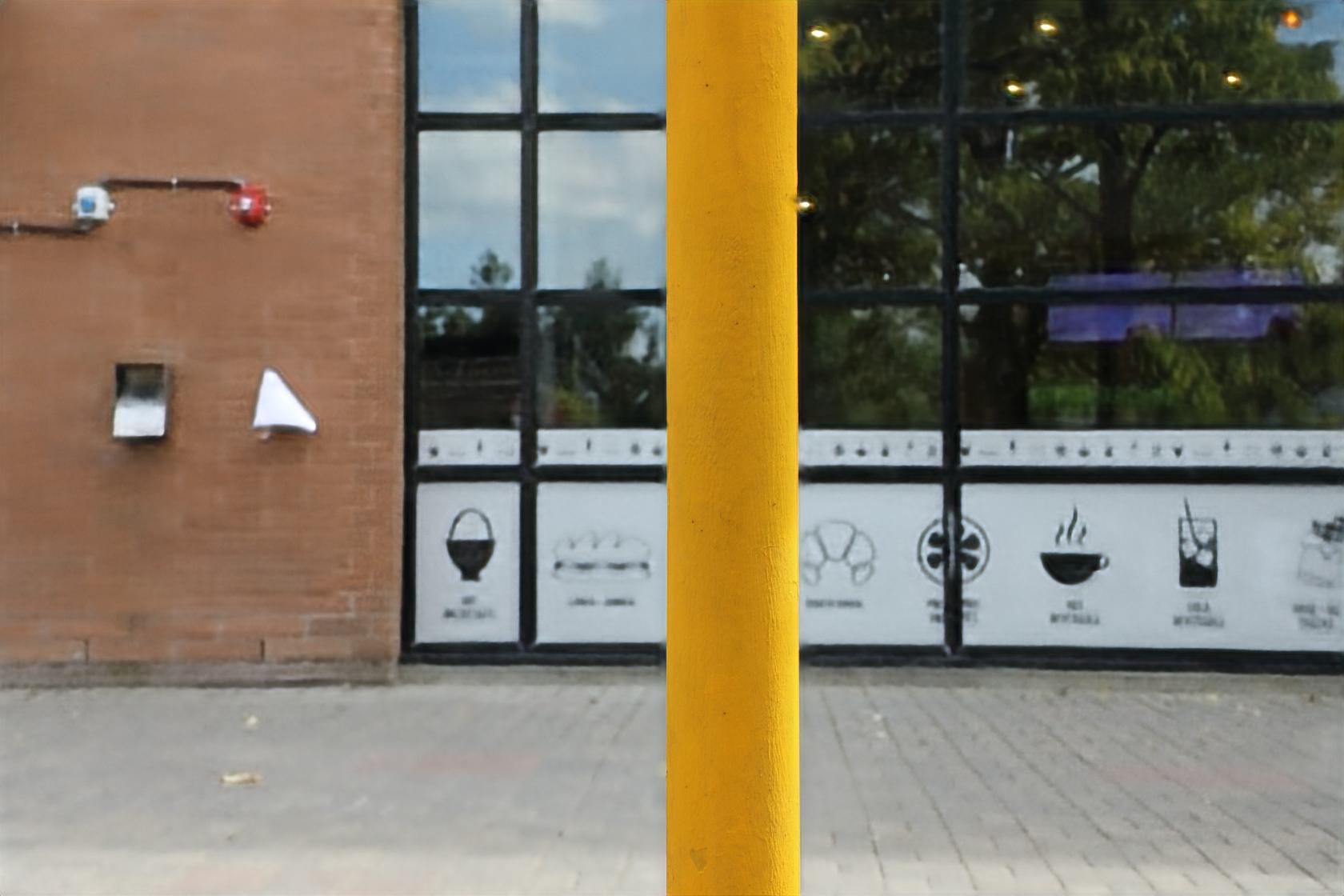}
	\includegraphics[width=0.16\textwidth]{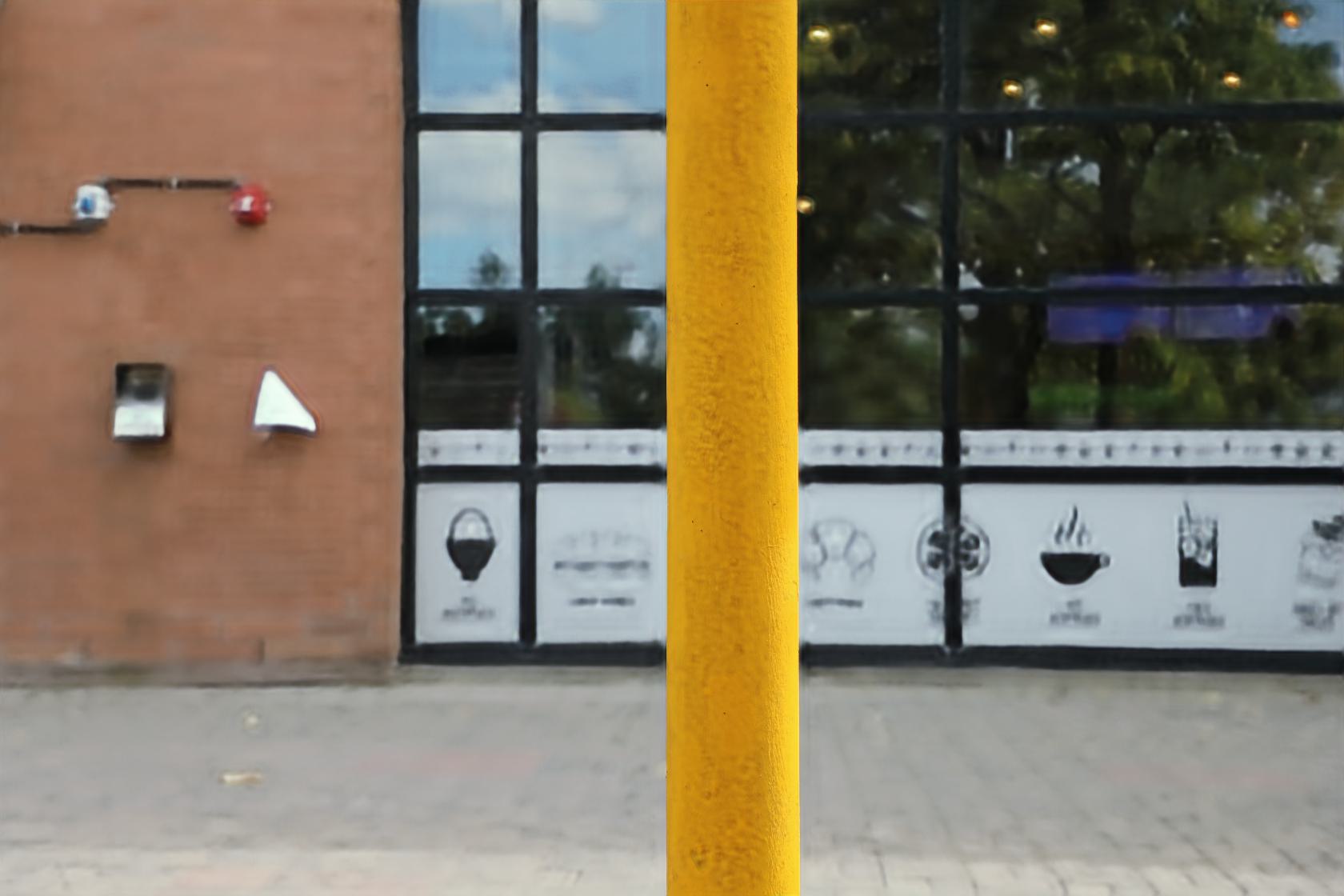}
	\includegraphics[width=0.16\textwidth]{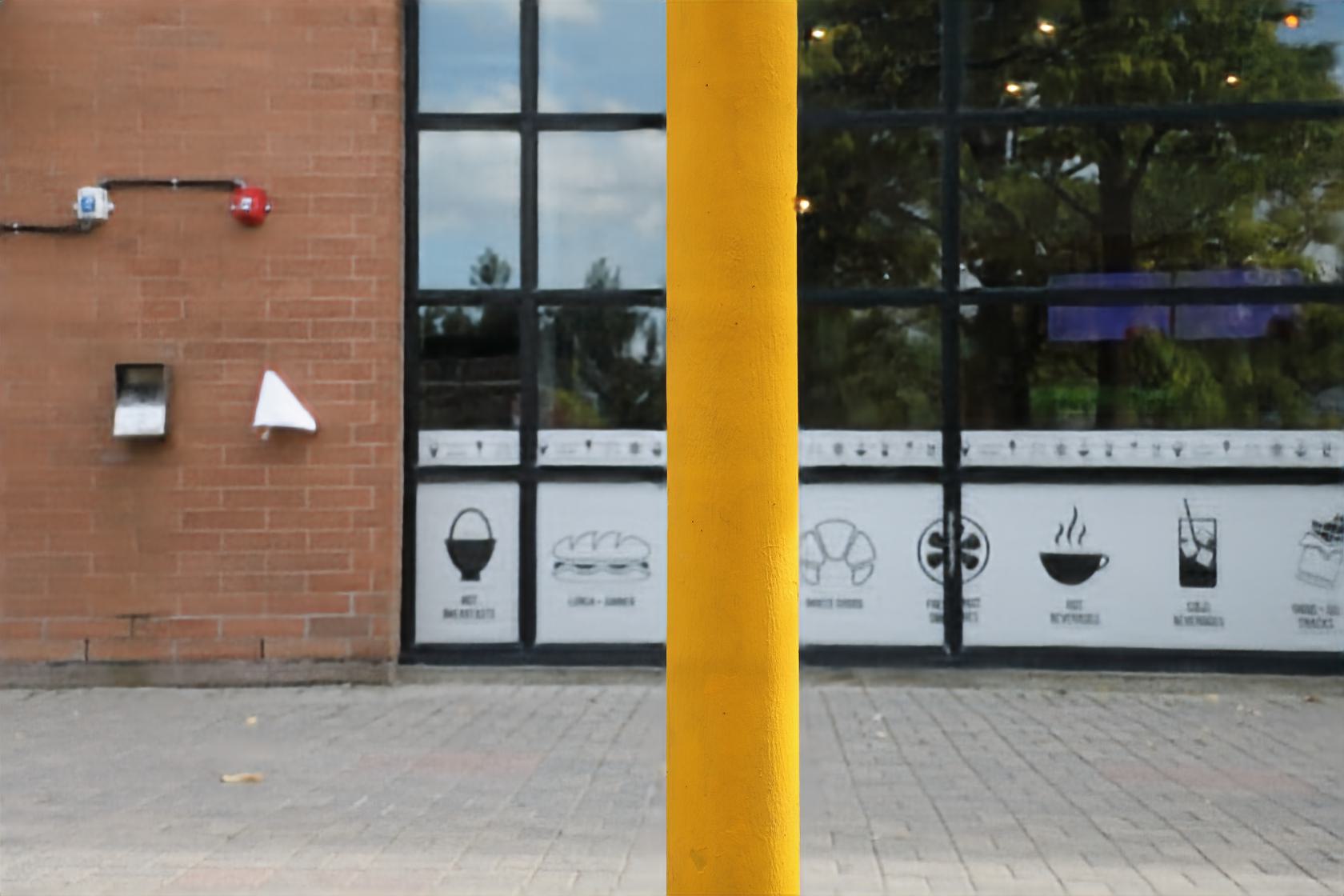}
	\includegraphics[width=0.16\textwidth]{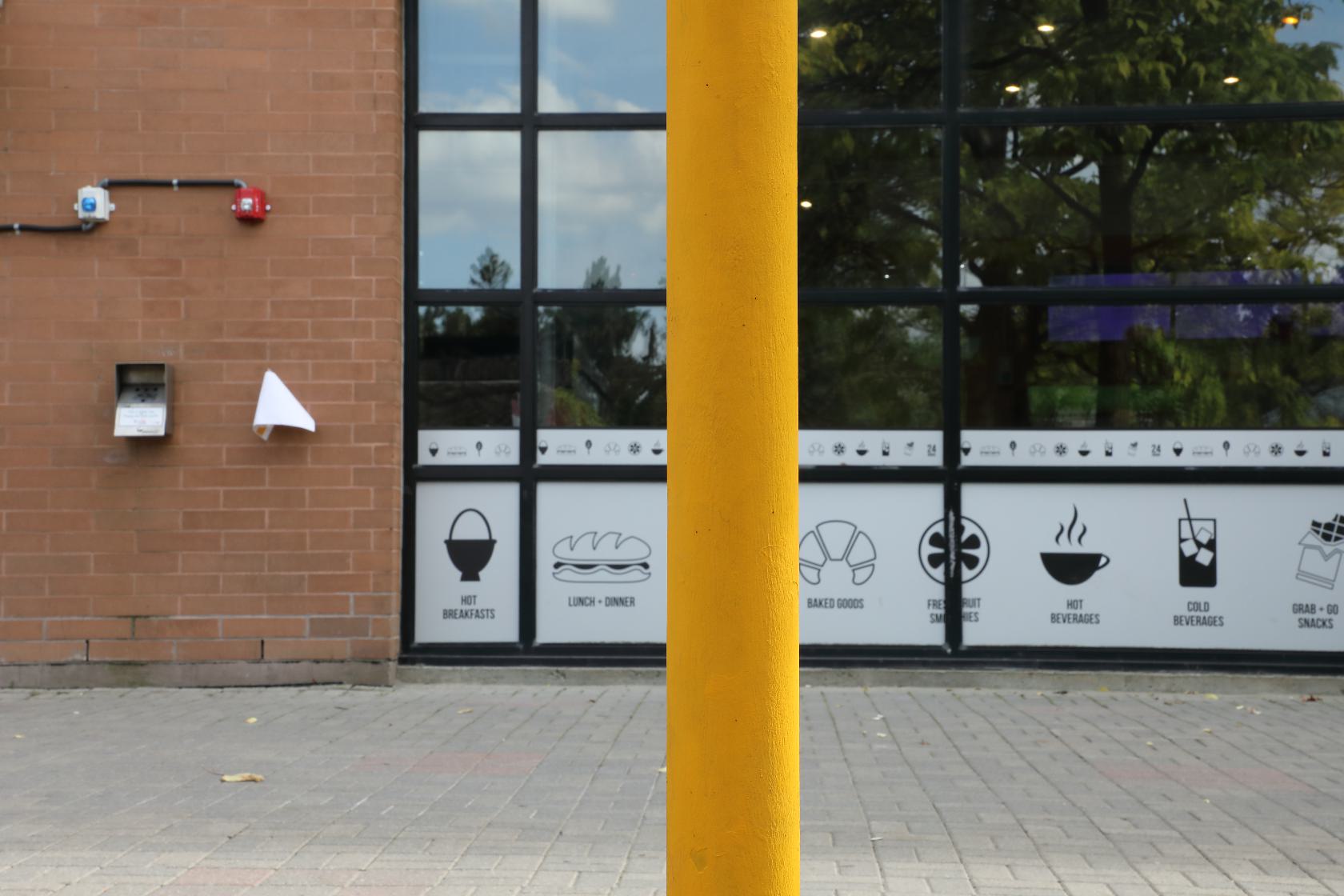}
	
	\frame{\includegraphics[width=0.16\textwidth]{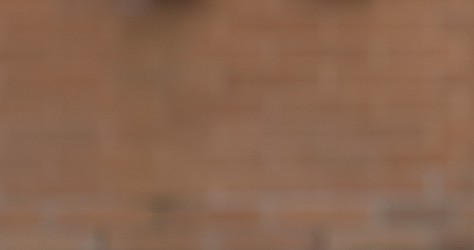}}
	\frame{\includegraphics[width=0.16\textwidth]{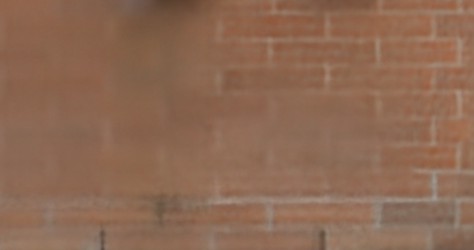}}
	\frame{\includegraphics[width=0.16\textwidth]{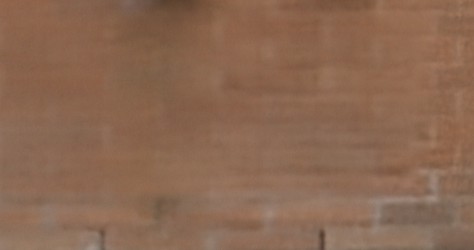}}
	\frame{\includegraphics[width=0.16\textwidth]{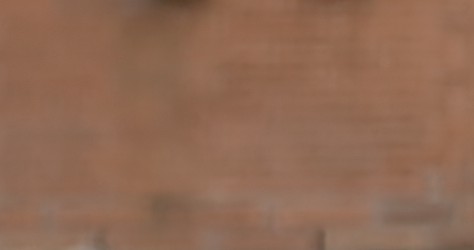}}
	\frame{\includegraphics[width=0.16\textwidth]{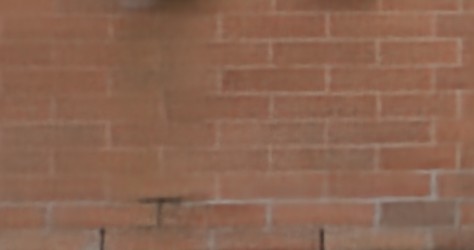}}
	\frame{\includegraphics[width=0.16\textwidth]{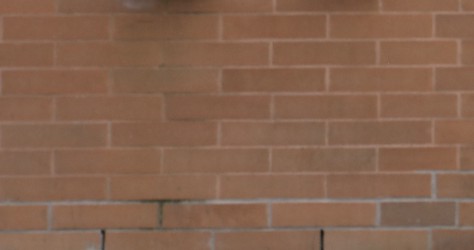}}
	
	\frame{\includegraphics[width=0.16\textwidth]{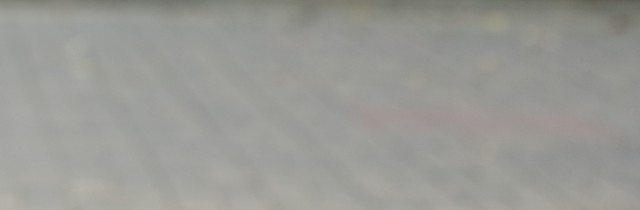}}
	\frame{\includegraphics[width=0.16\textwidth]{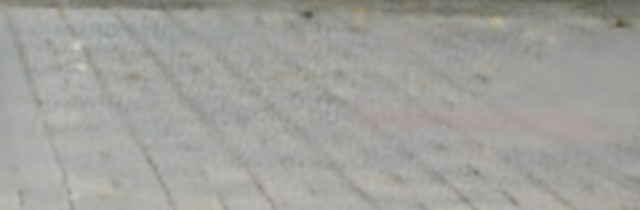}}
	\frame{\includegraphics[width=0.16\textwidth]{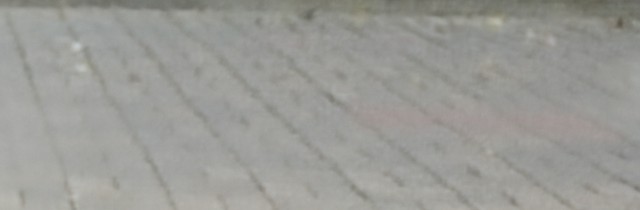}}
	\frame{\includegraphics[width=0.16\textwidth]{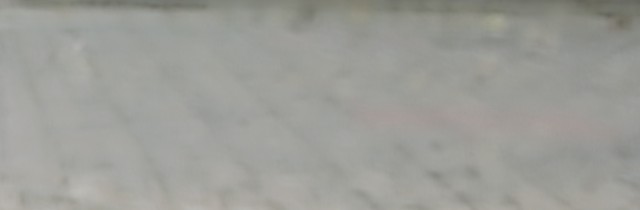}}
	\frame{\includegraphics[width=0.16\textwidth]{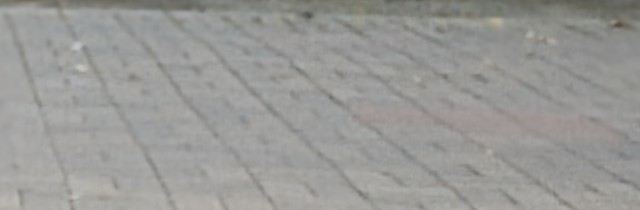}}
	\frame{\includegraphics[width=0.16\textwidth]{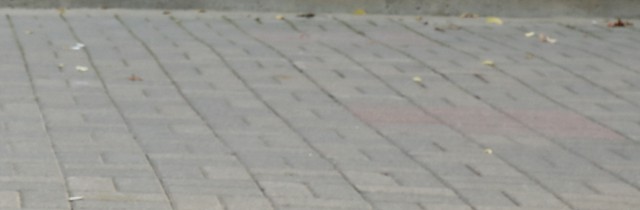}}
	
	\vspace{4pt}
	
	\includegraphics[width=0.16\textwidth]{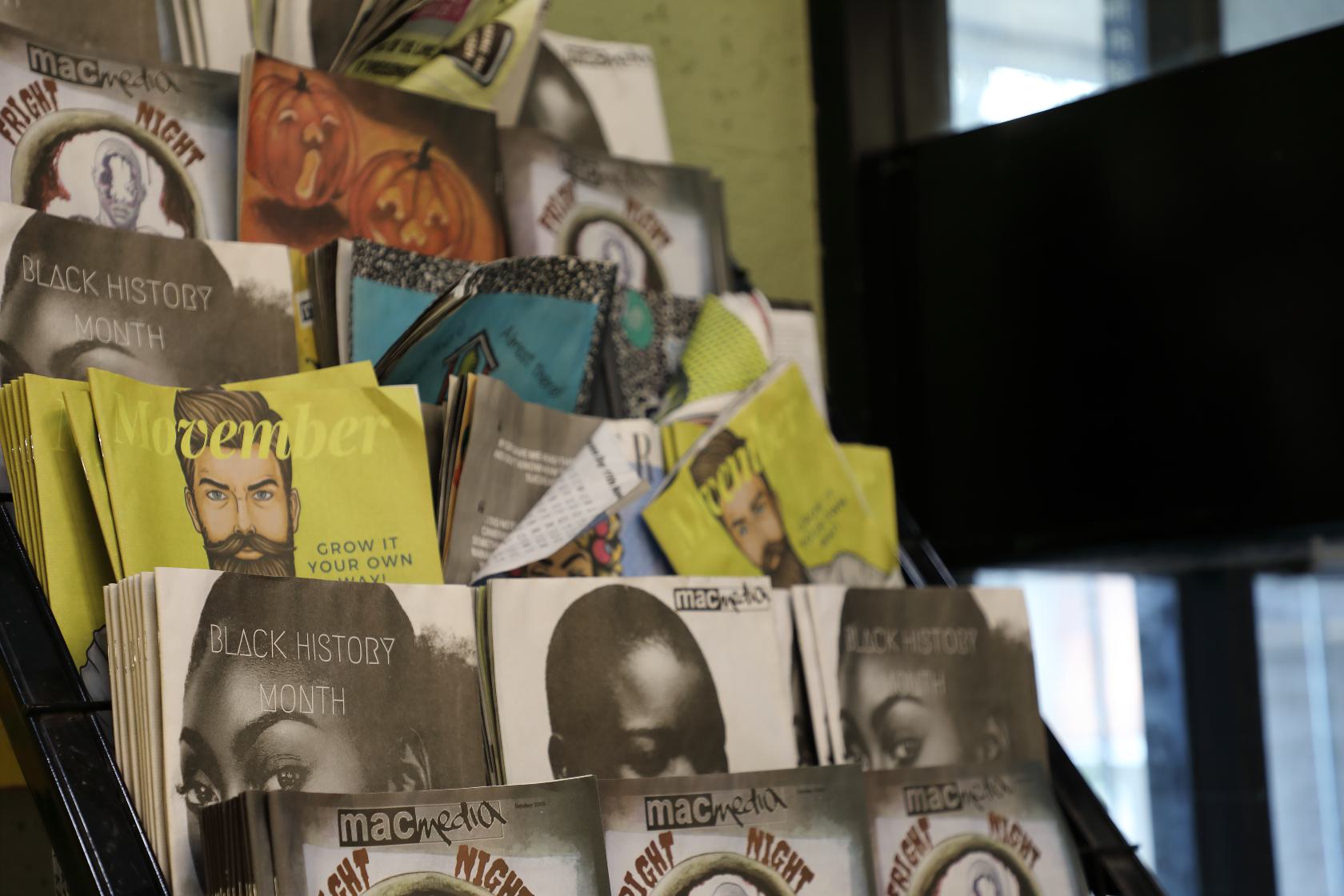}
	\includegraphics[width=0.16\textwidth]{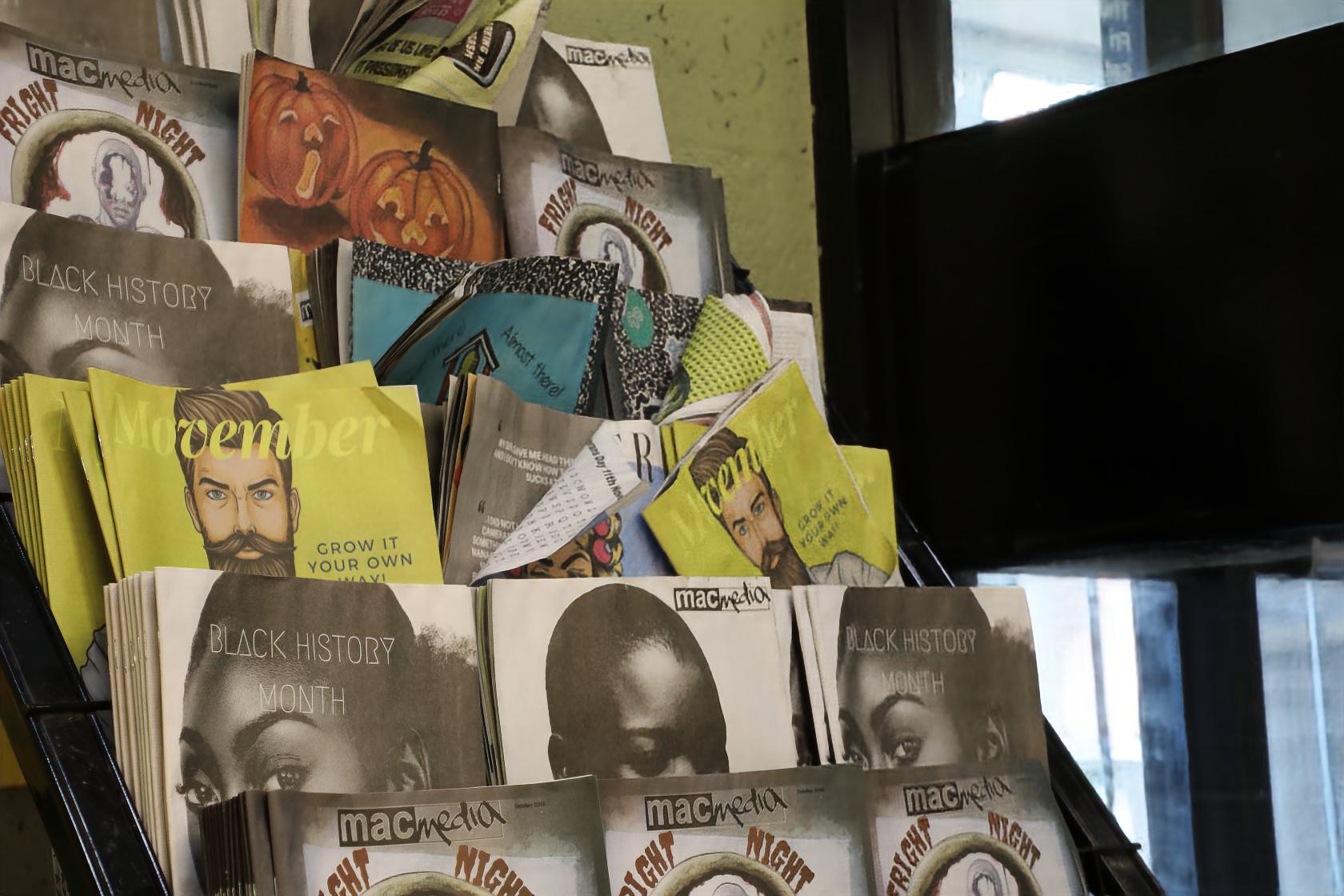}
	\includegraphics[width=0.16\textwidth]{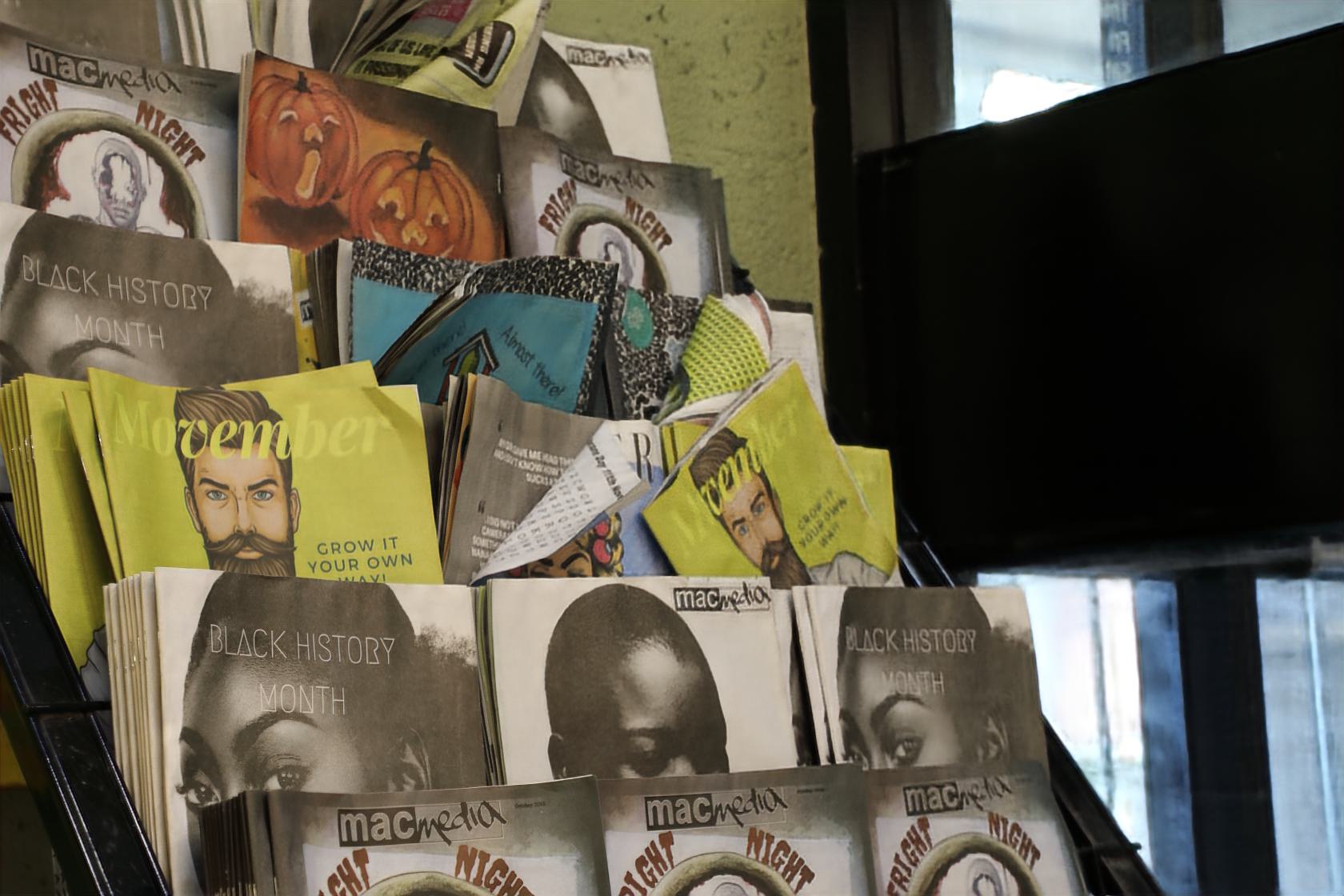}
	\includegraphics[width=0.16\textwidth]{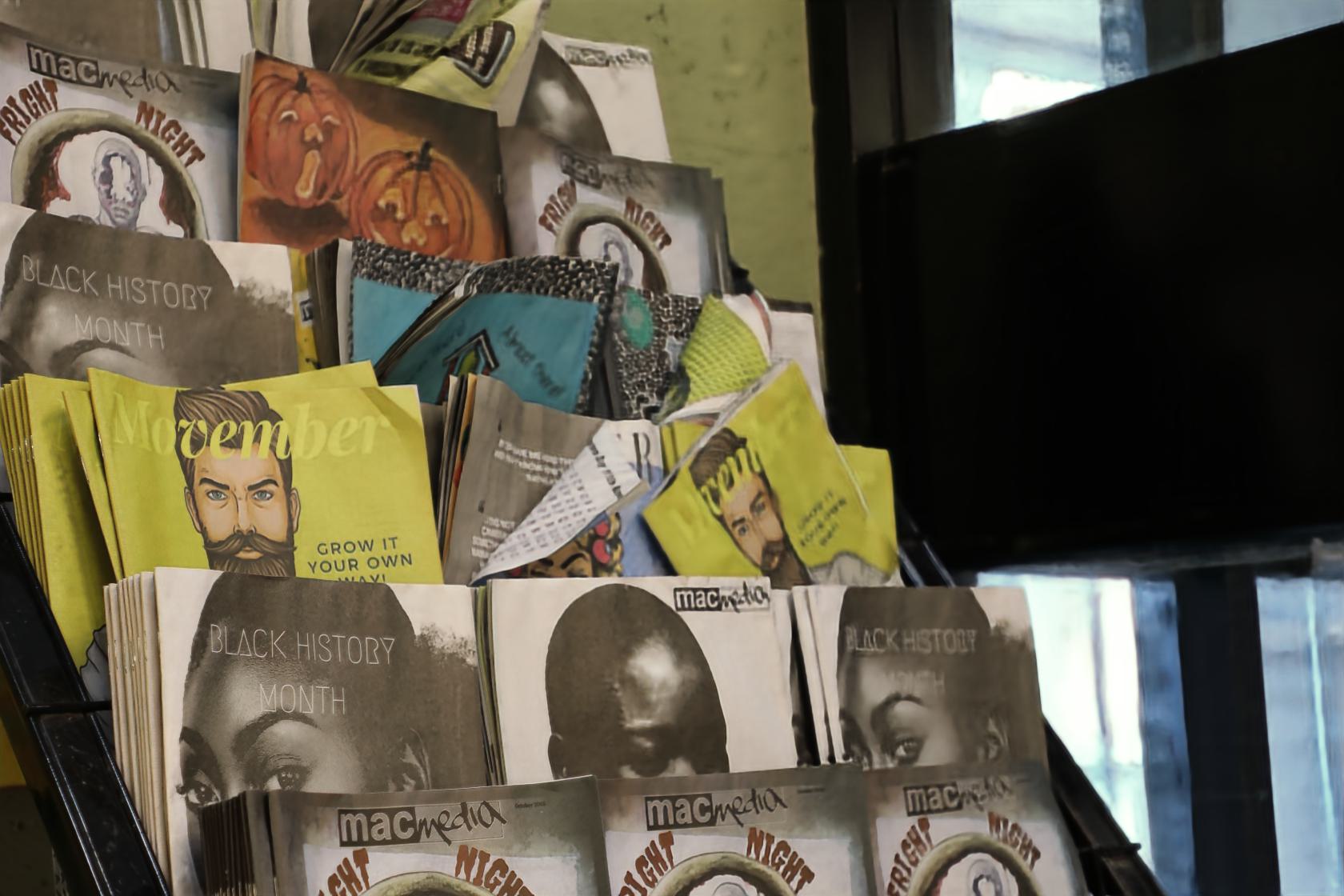}
	\includegraphics[width=0.16\textwidth]{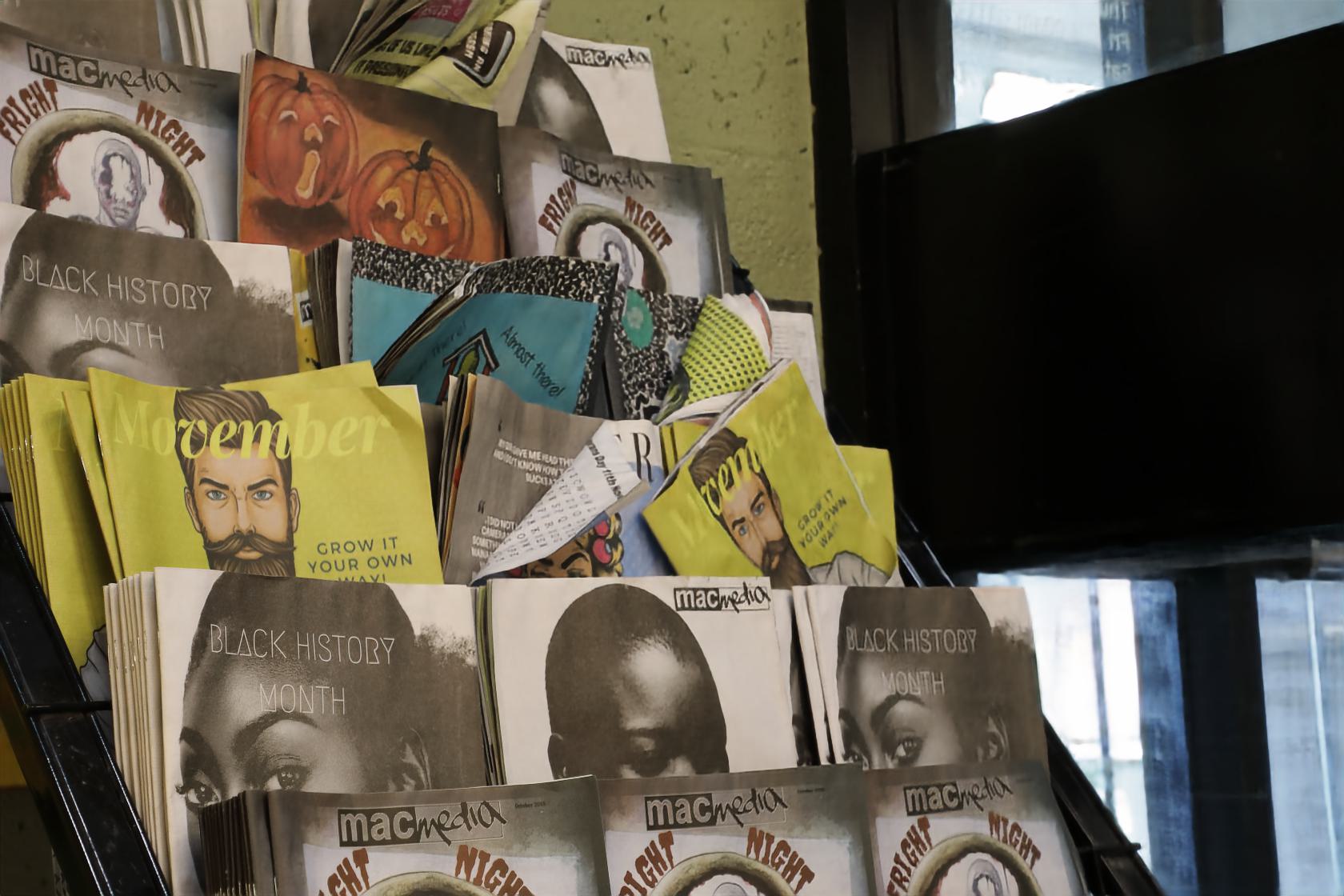}
	\includegraphics[width=0.16\textwidth]{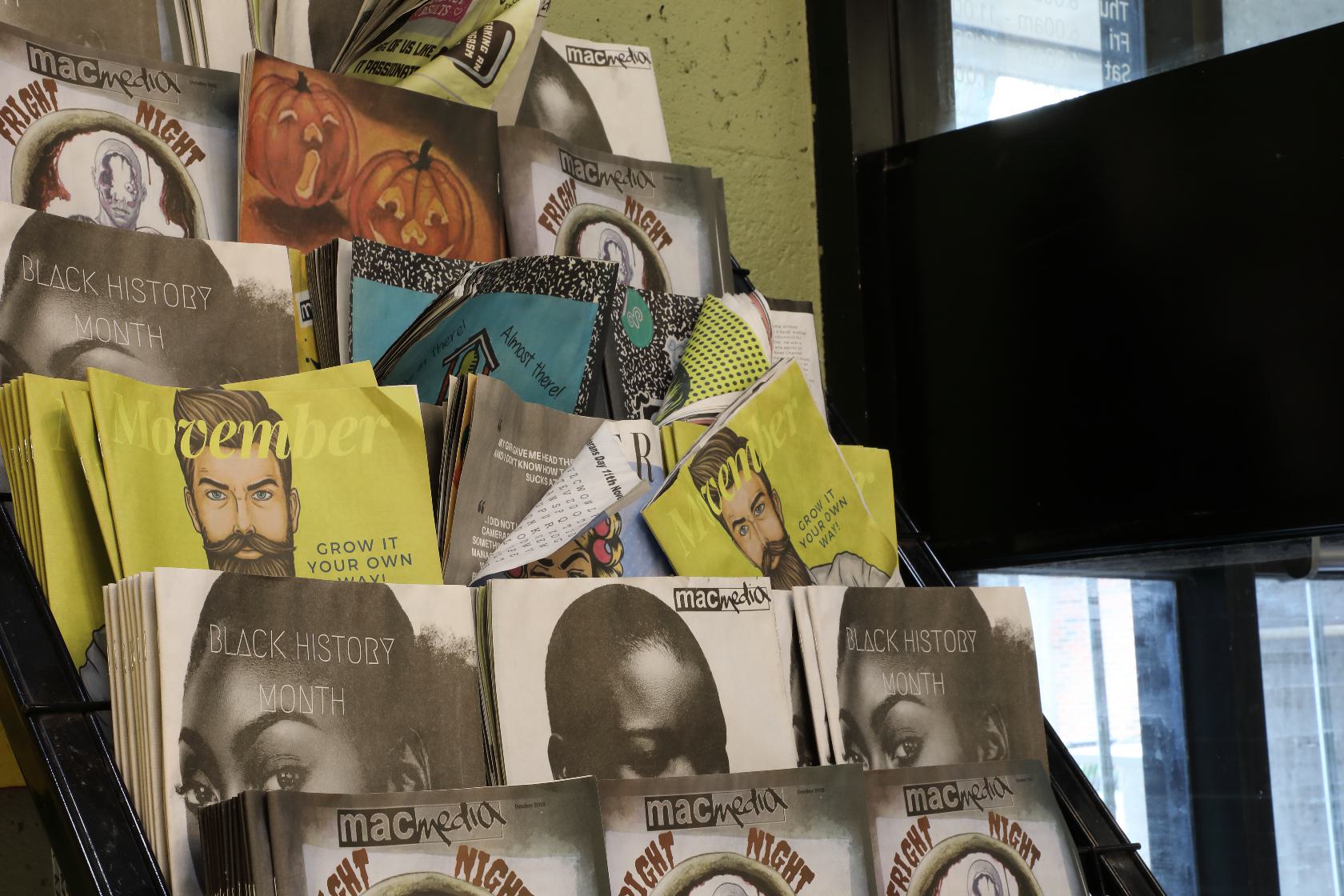}
	
	\frame{\includegraphics[width=0.16\textwidth]{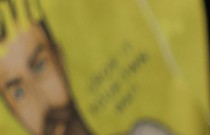}}
	\frame{\includegraphics[width=0.16\textwidth]{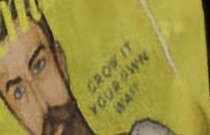}}
	\frame{\includegraphics[width=0.16\textwidth]{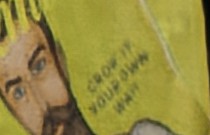}}
	\frame{\includegraphics[width=0.16\textwidth]{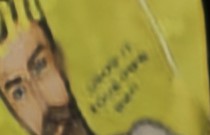}}
	\frame{\includegraphics[width=0.16\textwidth]{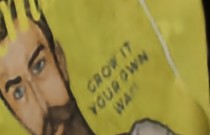}}
	\frame{\includegraphics[width=0.16\textwidth]{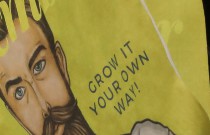}}
	
	\frame{\includegraphics[width=0.16\textwidth]{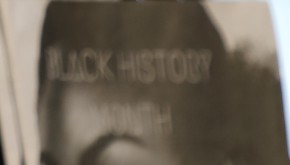}}
	\frame{\includegraphics[width=0.16\textwidth]{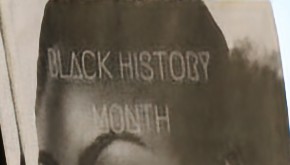}}
	\frame{\includegraphics[width=0.16\textwidth]{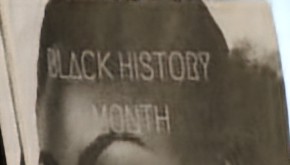}}
	\frame{\includegraphics[width=0.16\textwidth]{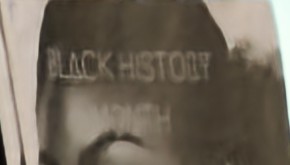}}
	\frame{\includegraphics[width=0.16\textwidth]{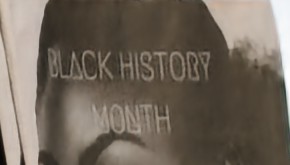}}
	\frame{\includegraphics[width=0.16\textwidth]{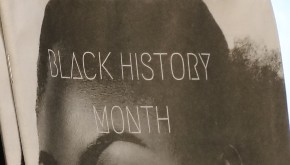}}
	
	\vspace{4pt}
	
	\includegraphics[width=0.16\textwidth]{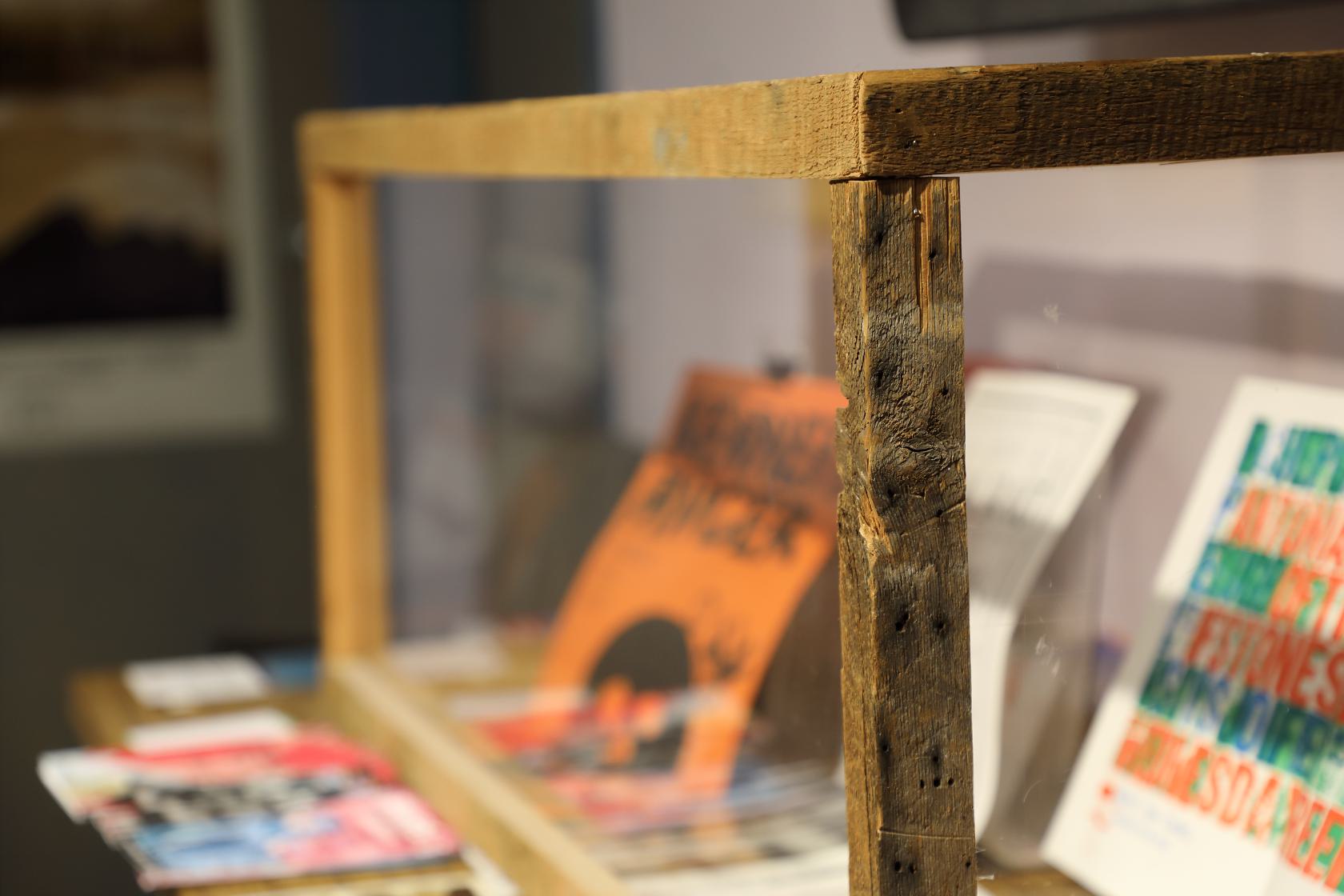}
	\includegraphics[width=0.16\textwidth]{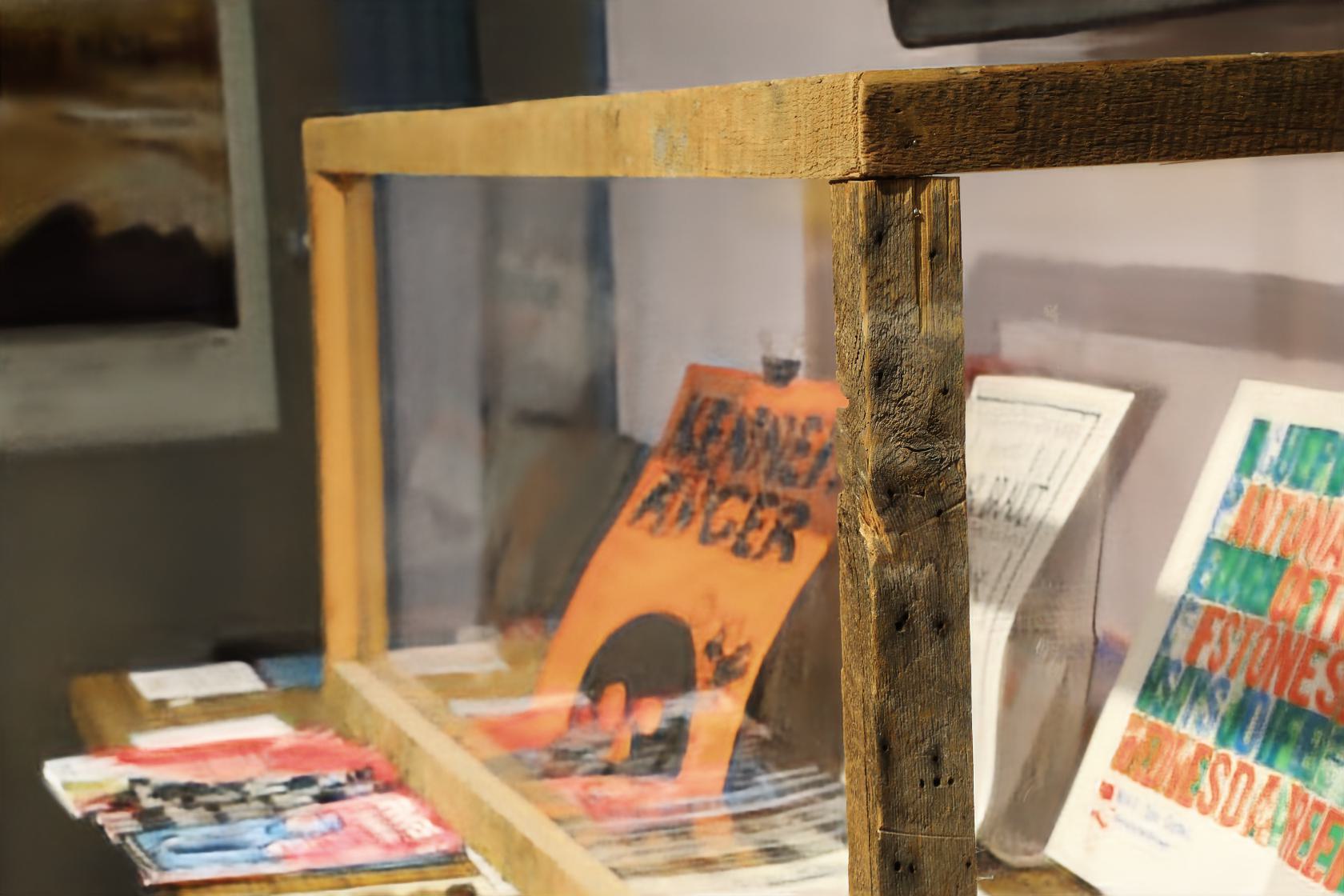}
	\includegraphics[width=0.16\textwidth]{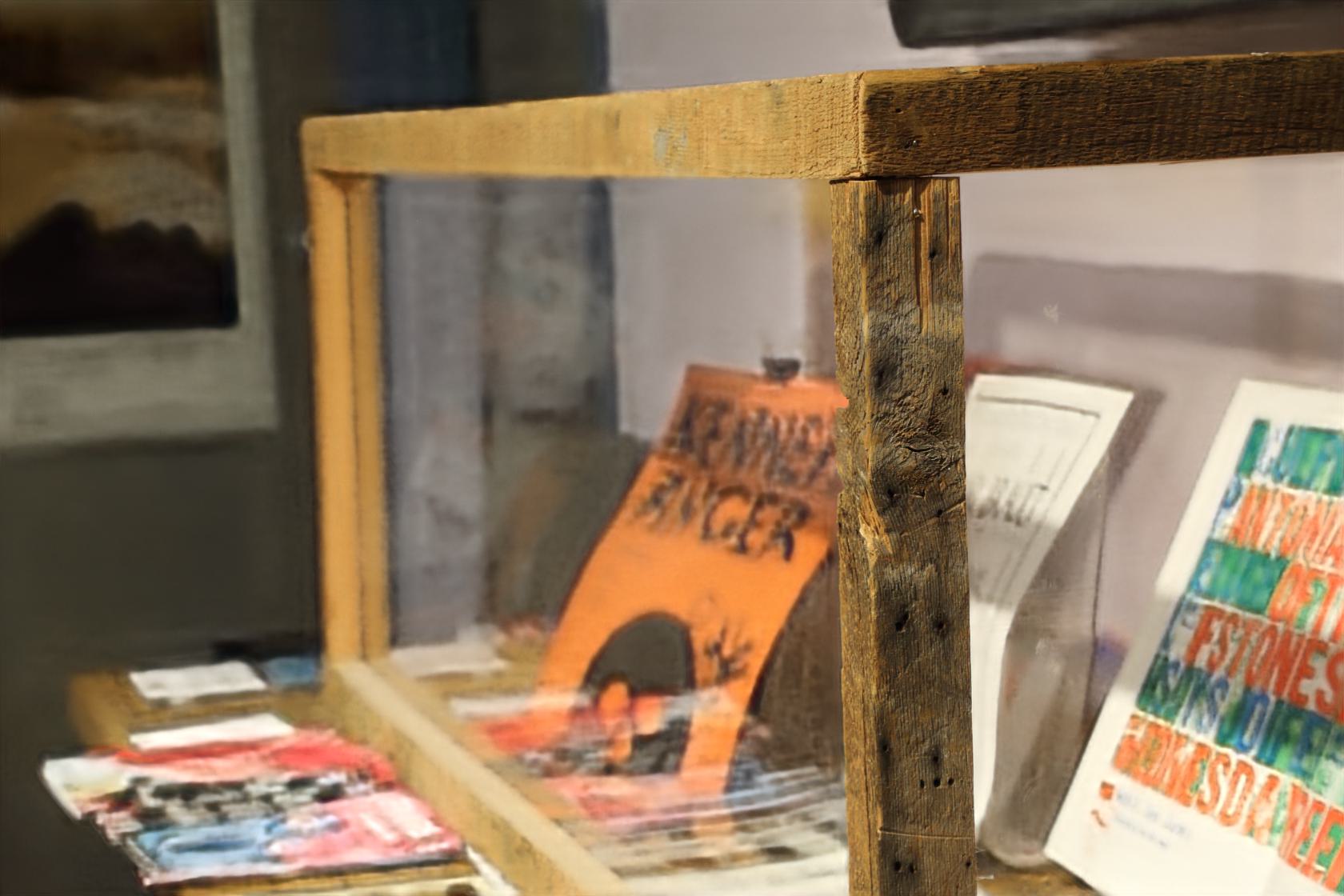}
	\includegraphics[width=0.16\textwidth]{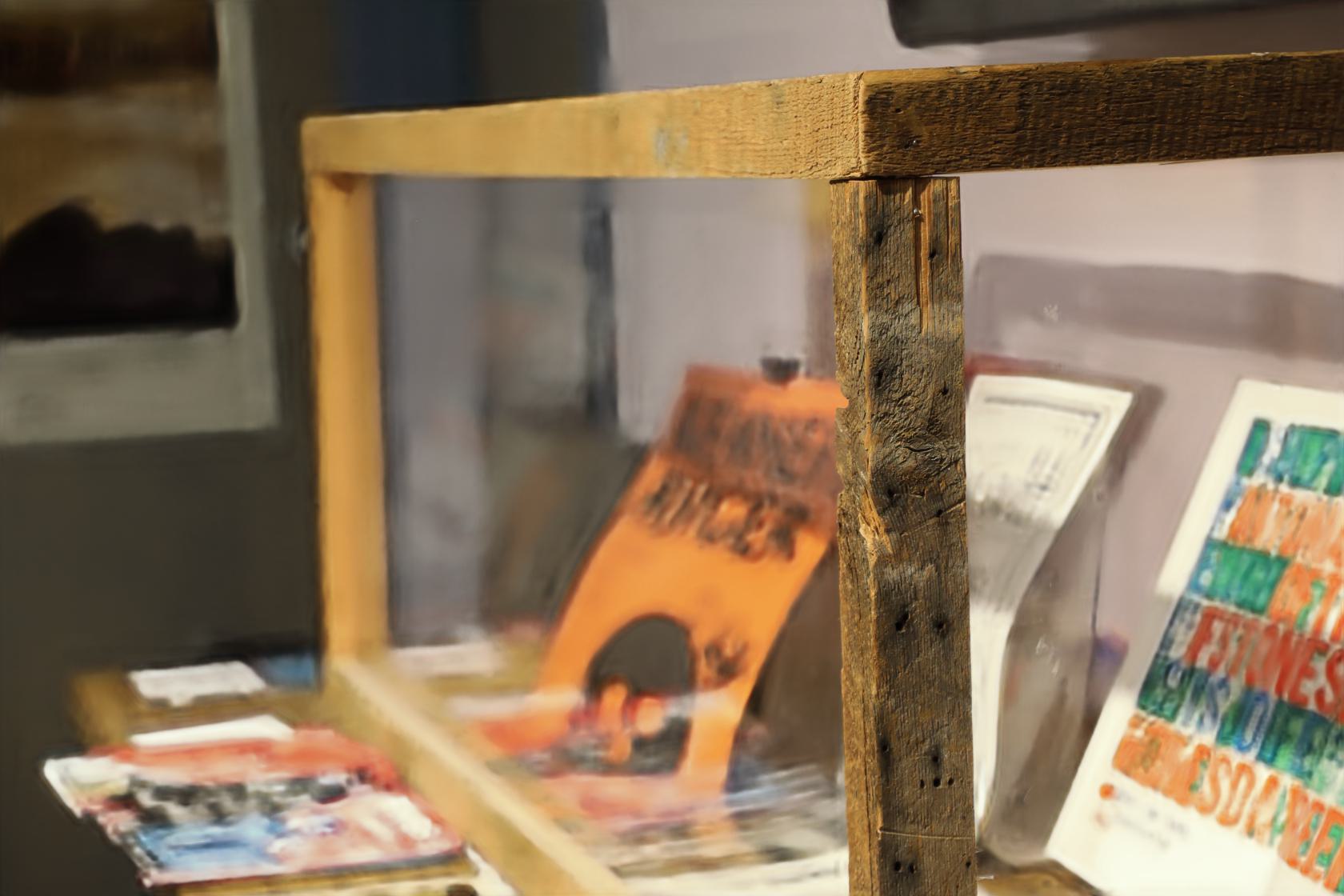}
	\includegraphics[width=0.16\textwidth]{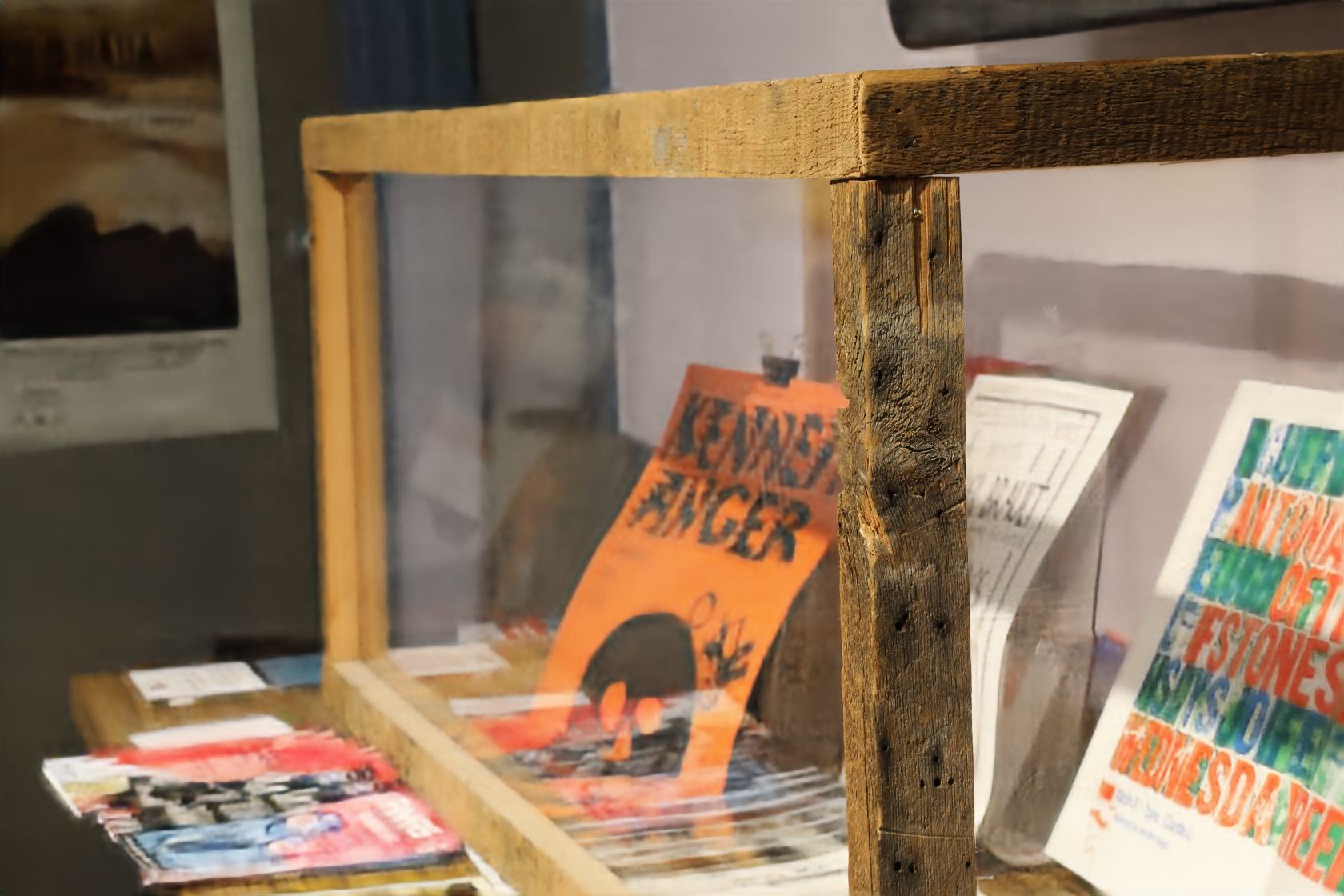}
	\includegraphics[width=0.16\textwidth]{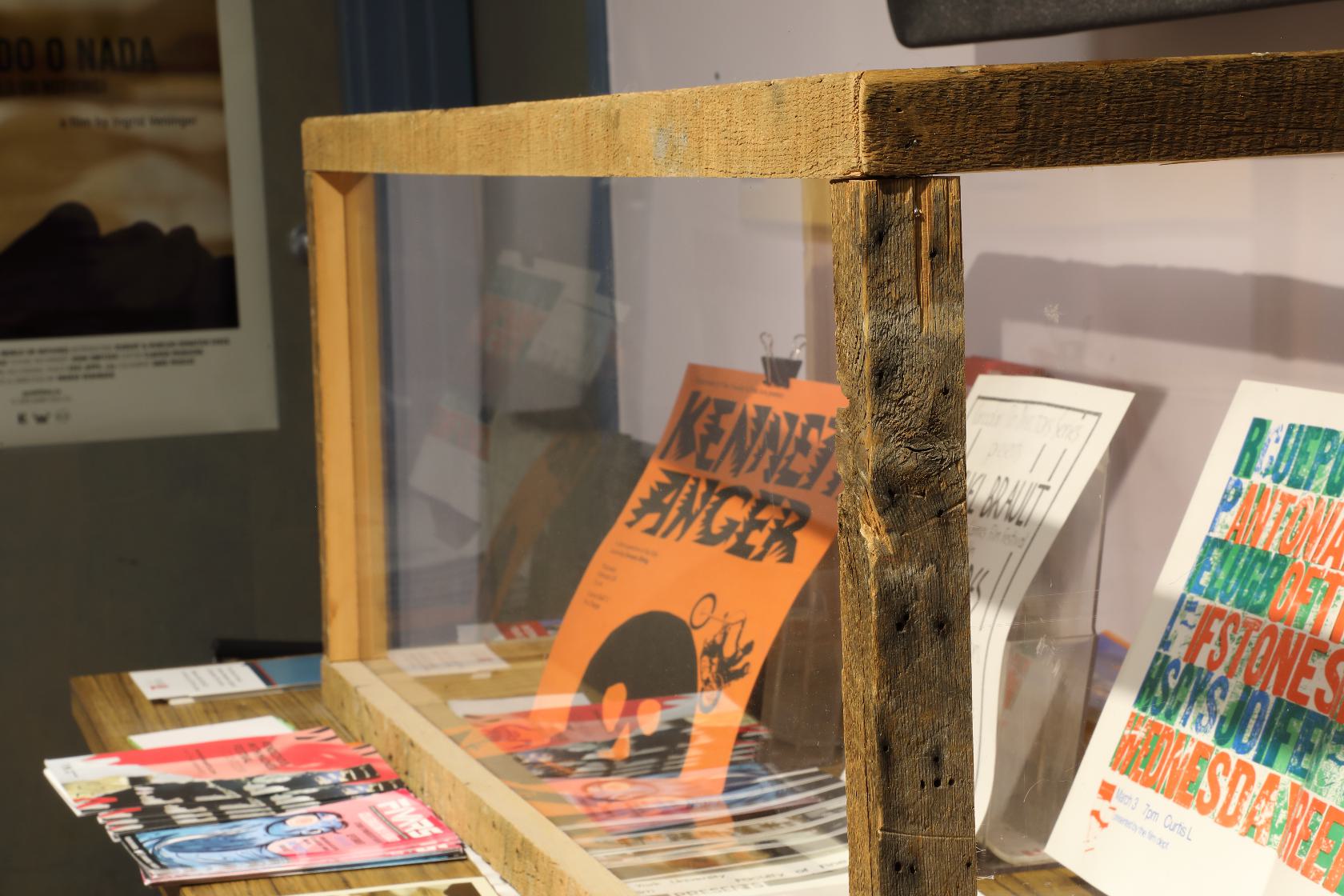}
	
	\frame{\includegraphics[width=0.16\textwidth]{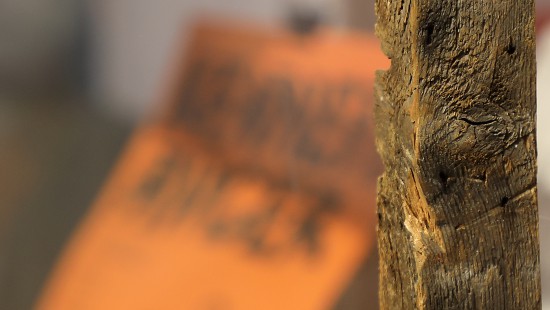}}
	\frame{\includegraphics[width=0.16\textwidth]{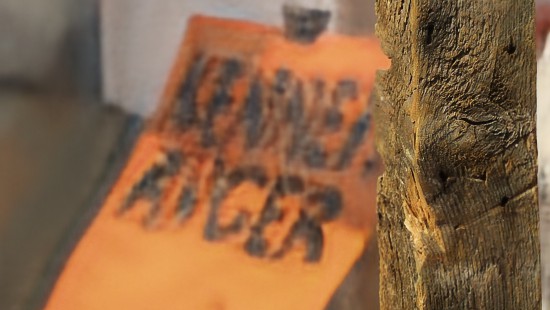}}
	\frame{\includegraphics[width=0.16\textwidth]{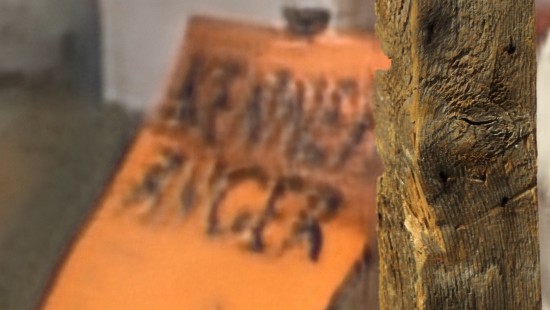}}
	\frame{\includegraphics[width=0.16\textwidth]{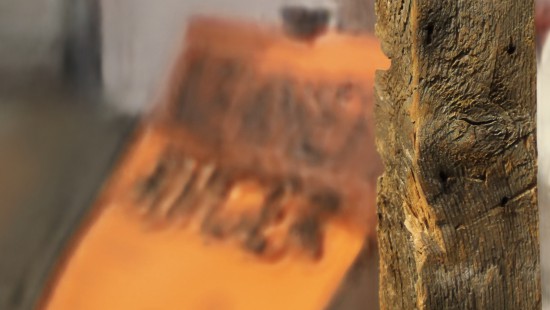}}
	\frame{\includegraphics[width=0.16\textwidth]{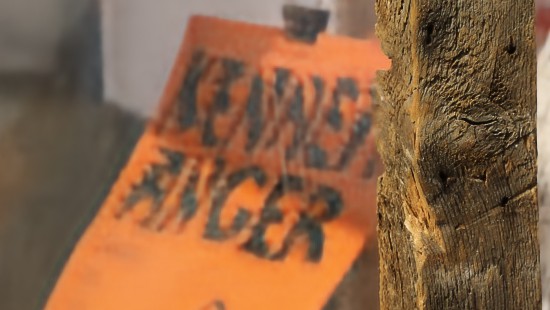}}
	\frame{\includegraphics[width=0.16\textwidth]{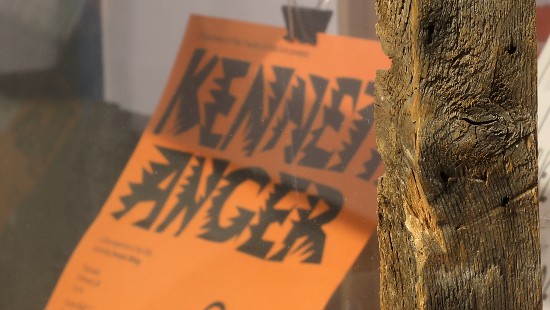}}
	
	\vspace{4pt}
	
	\includegraphics[width=0.16\textwidth]{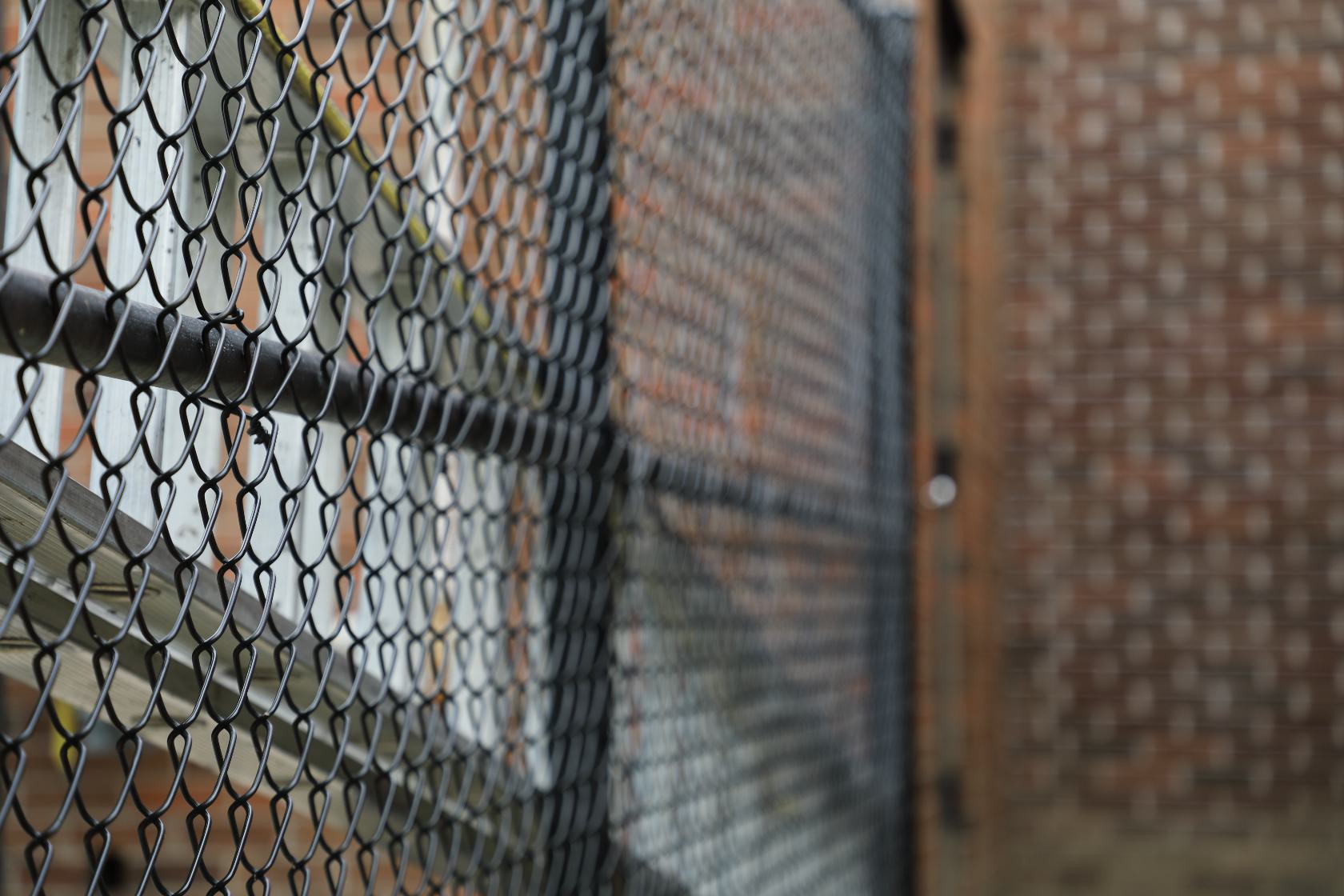}
	\includegraphics[width=0.16\textwidth]{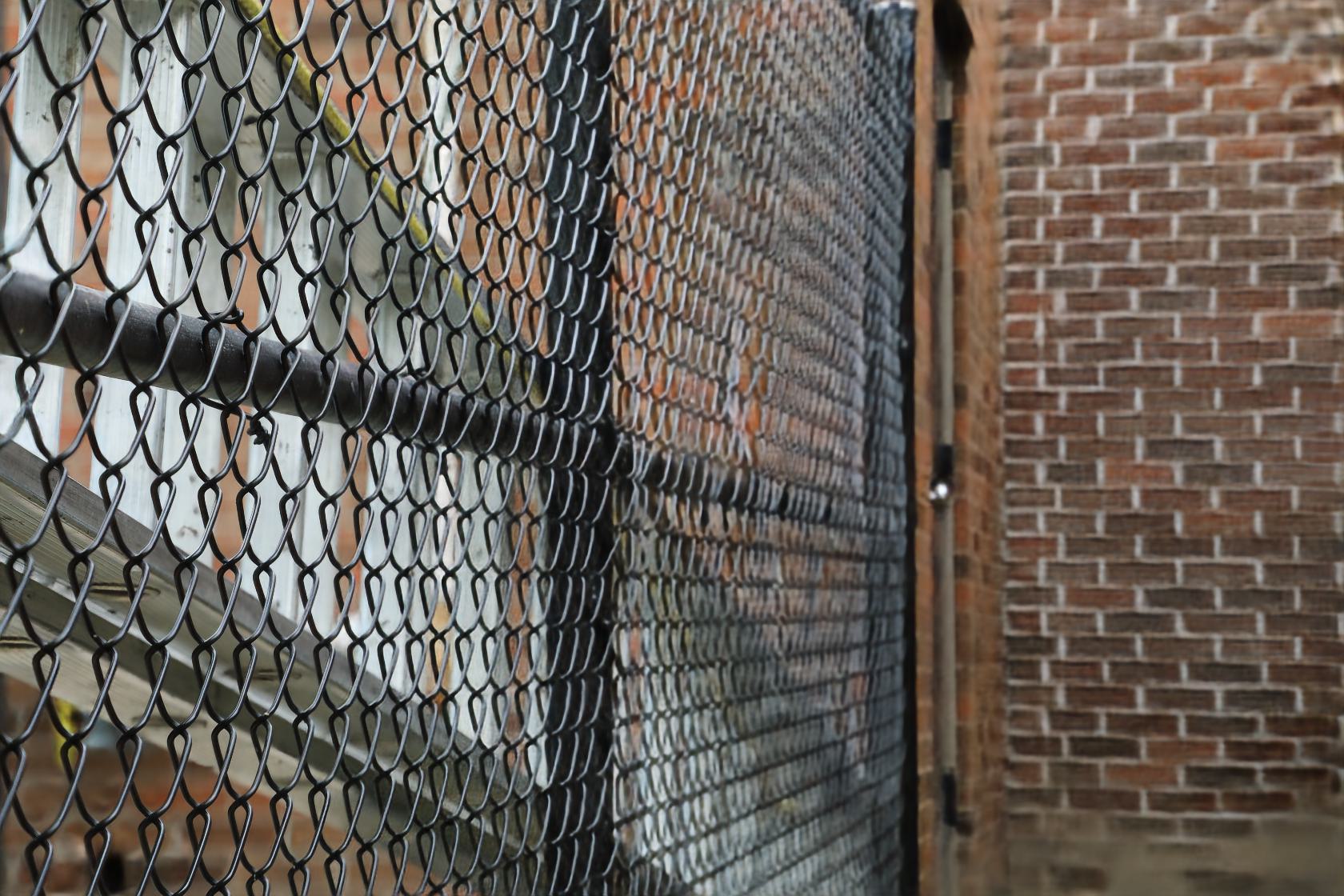}
	\includegraphics[width=0.16\textwidth]{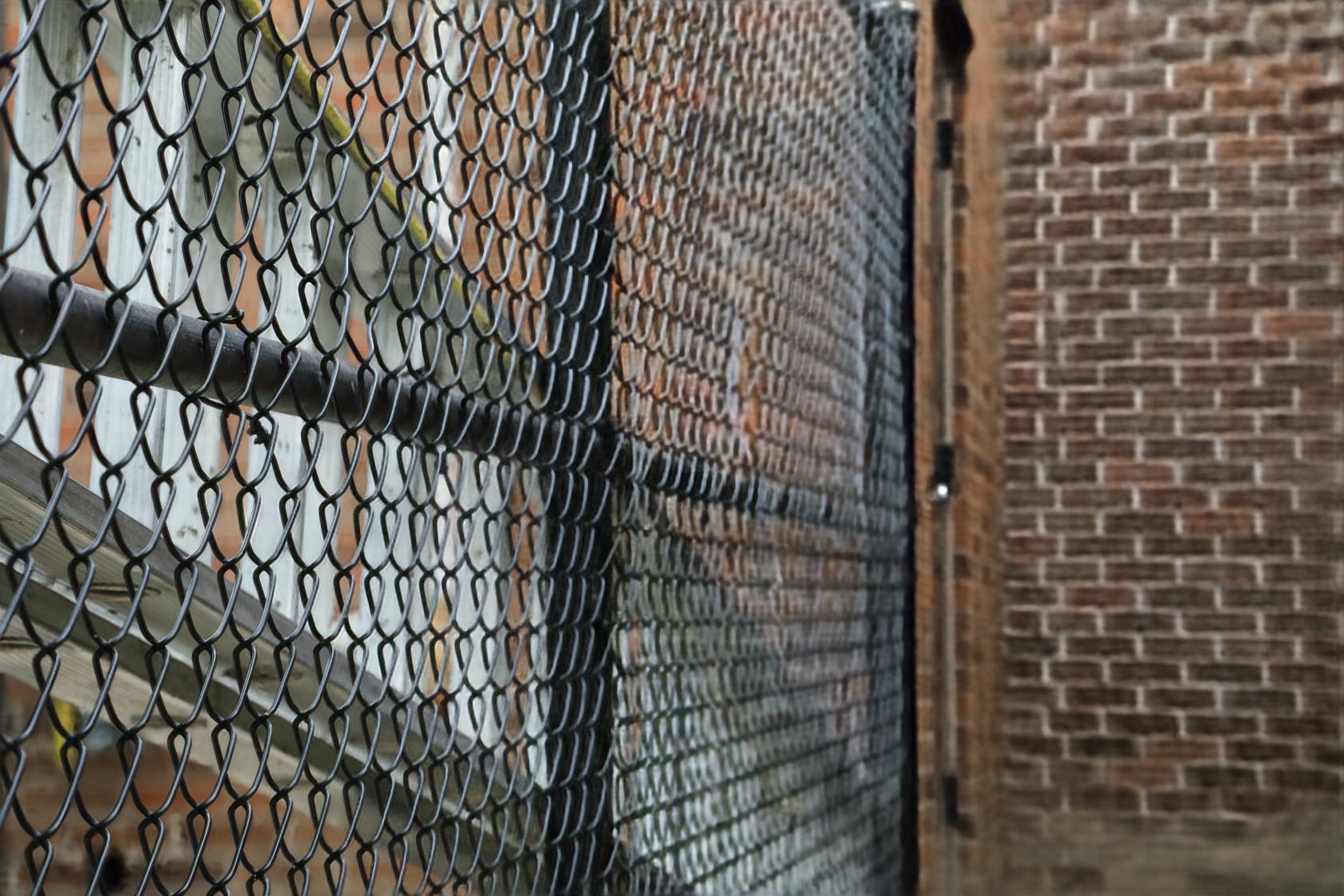}
	\includegraphics[width=0.16\textwidth]{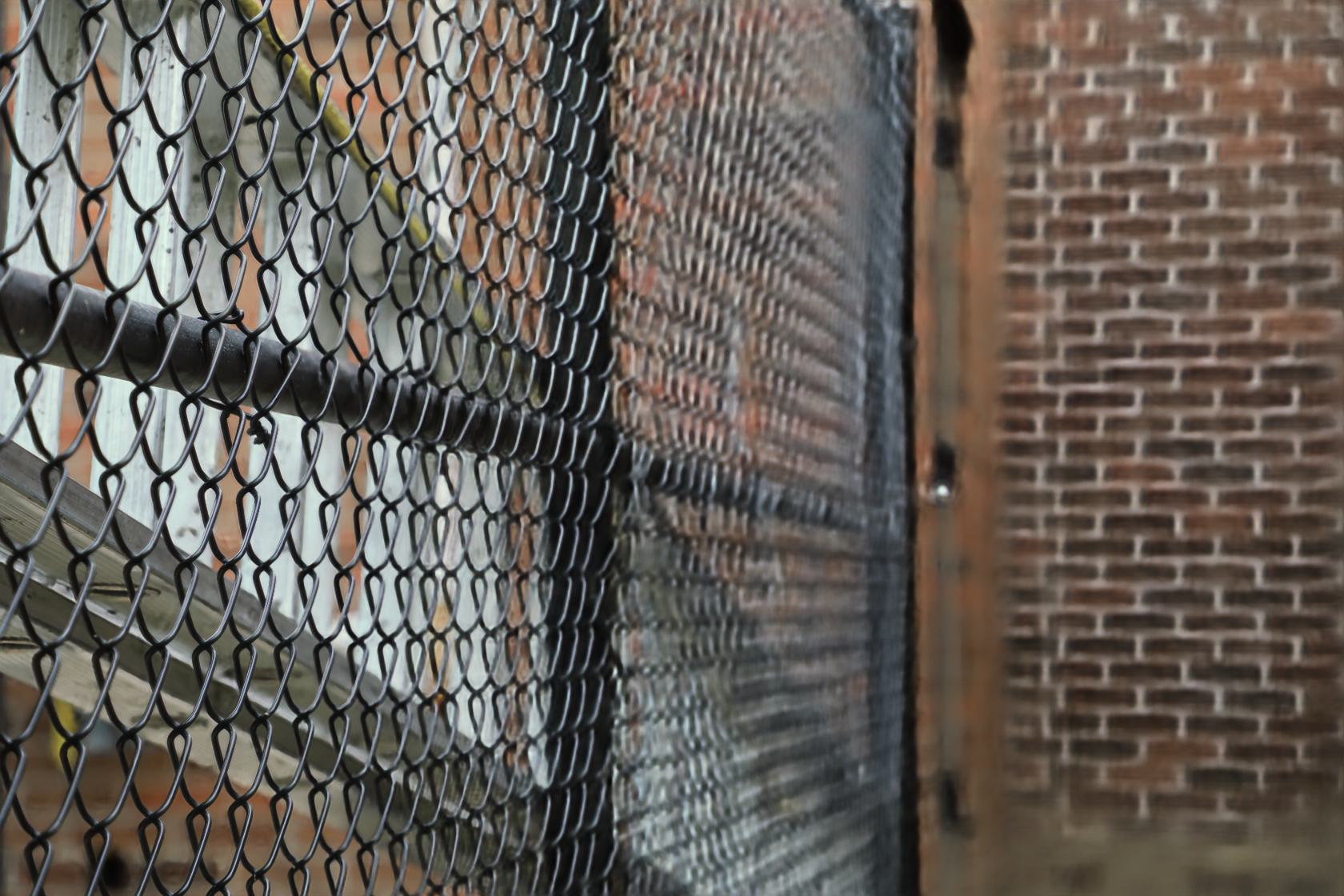}
	\includegraphics[width=0.16\textwidth]{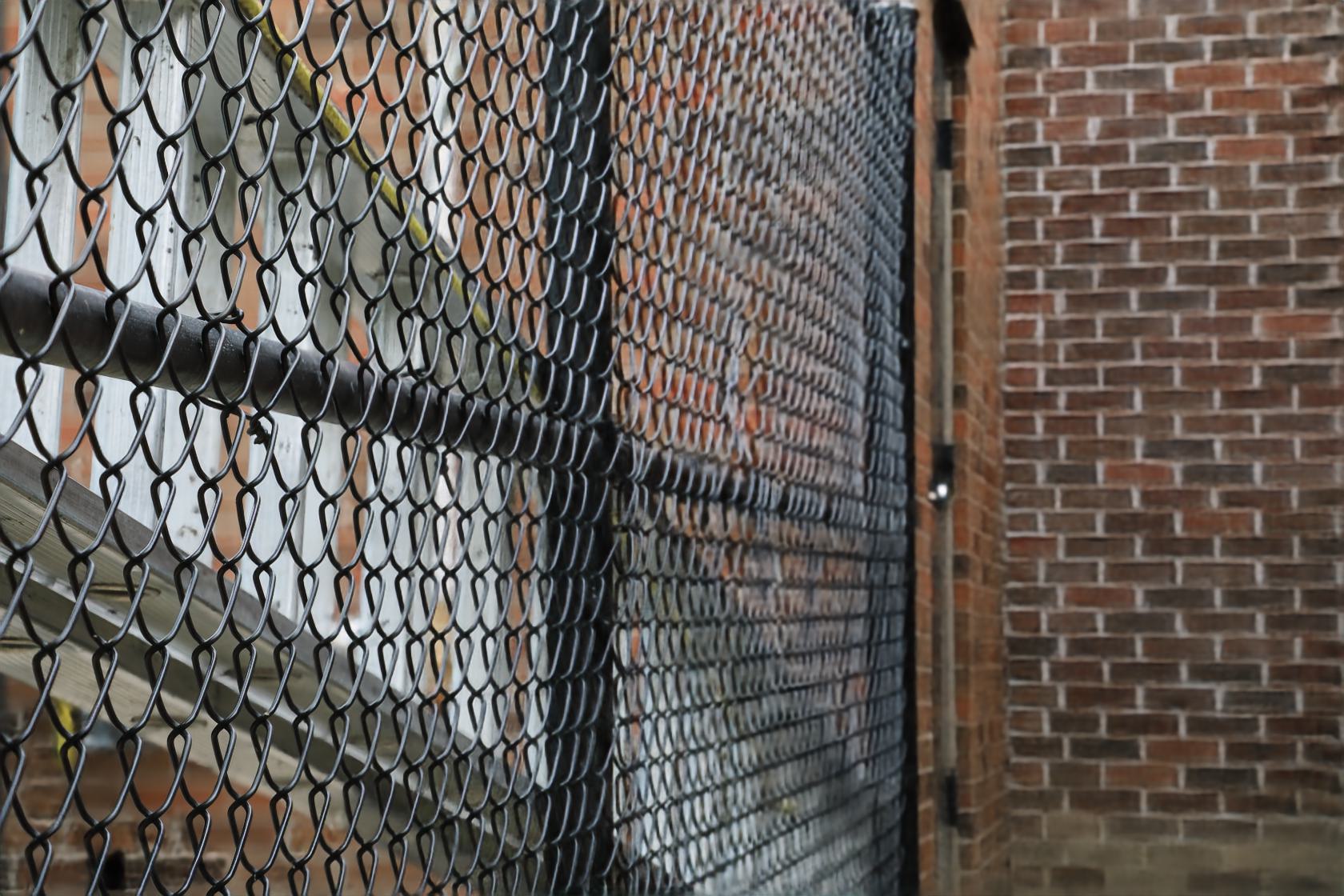}
	\includegraphics[width=0.16\textwidth]{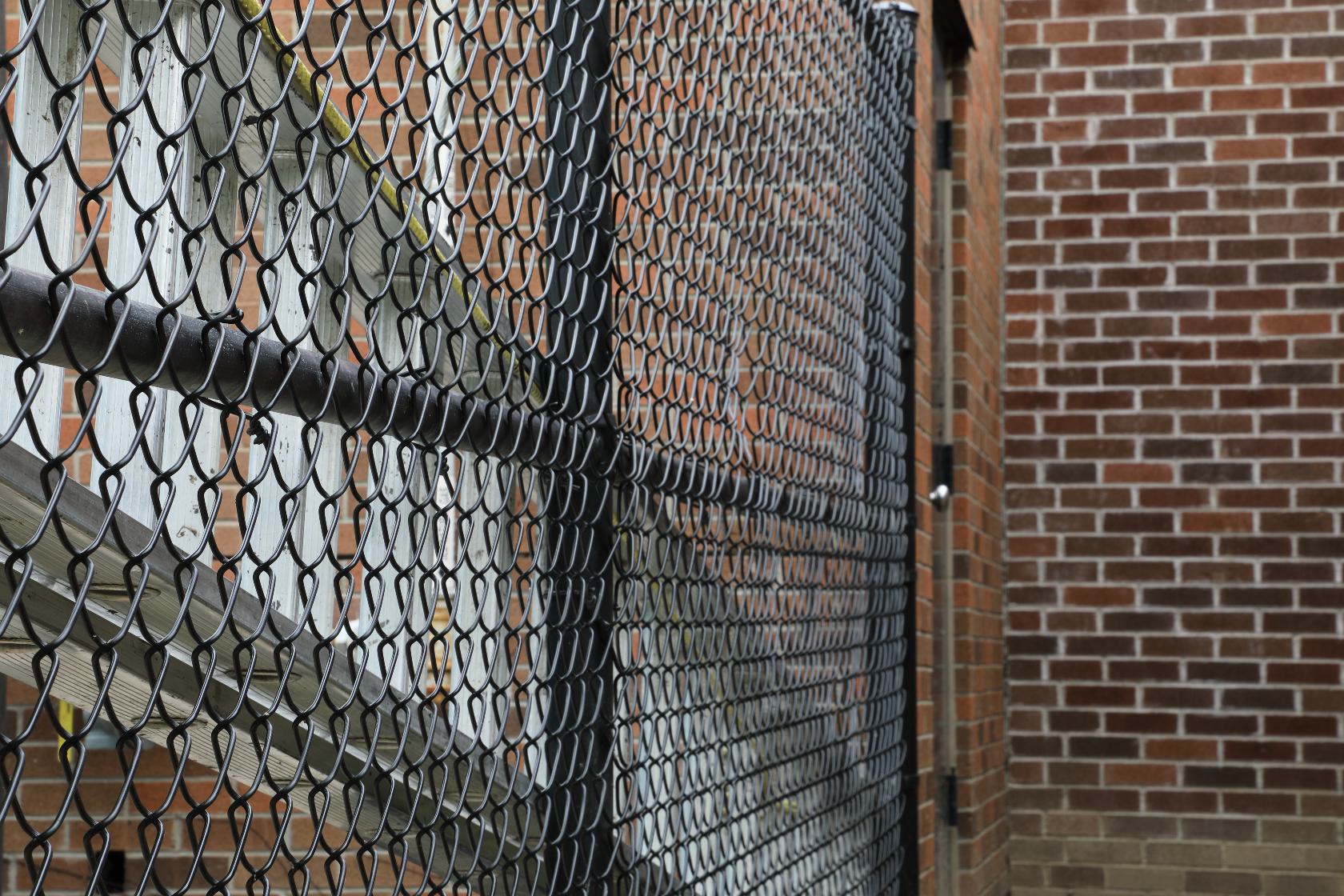}
	
	\frame{\includegraphics[width=0.16\textwidth]{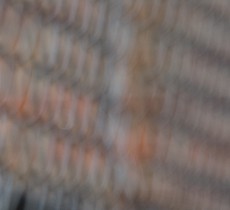}}
	\frame{\includegraphics[width=0.16\textwidth]{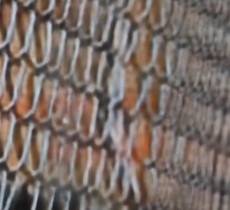}}
	\frame{\includegraphics[width=0.16\textwidth]{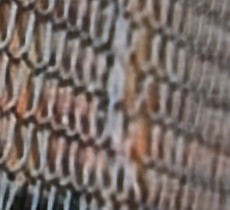}}
	\frame{\includegraphics[width=0.16\textwidth]{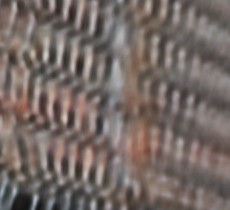}}
	\frame{\includegraphics[width=0.16\textwidth]{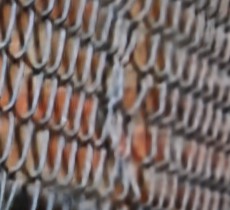}}
	\frame{\includegraphics[width=0.16\textwidth]{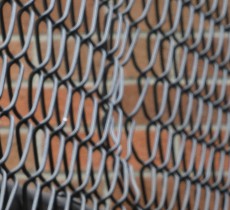}}
	
	\frame{\includegraphics[width=0.16\textwidth]{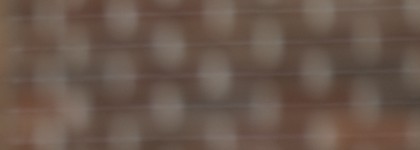}}
	\frame{\includegraphics[width=0.16\textwidth]{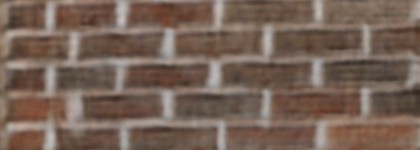}}
	\frame{\includegraphics[width=0.16\textwidth]{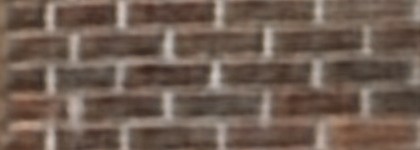}}
	\frame{\includegraphics[width=0.16\textwidth]{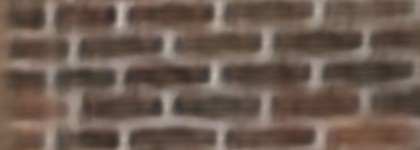}}
	\frame{\includegraphics[width=0.16\textwidth]{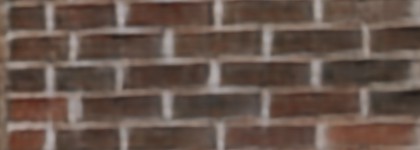}}
	\frame{\includegraphics[width=0.16\textwidth]{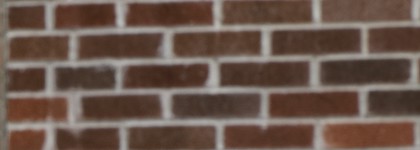}}
	
	\vspace{4pt}
	
	\includegraphics[width=0.16\textwidth]{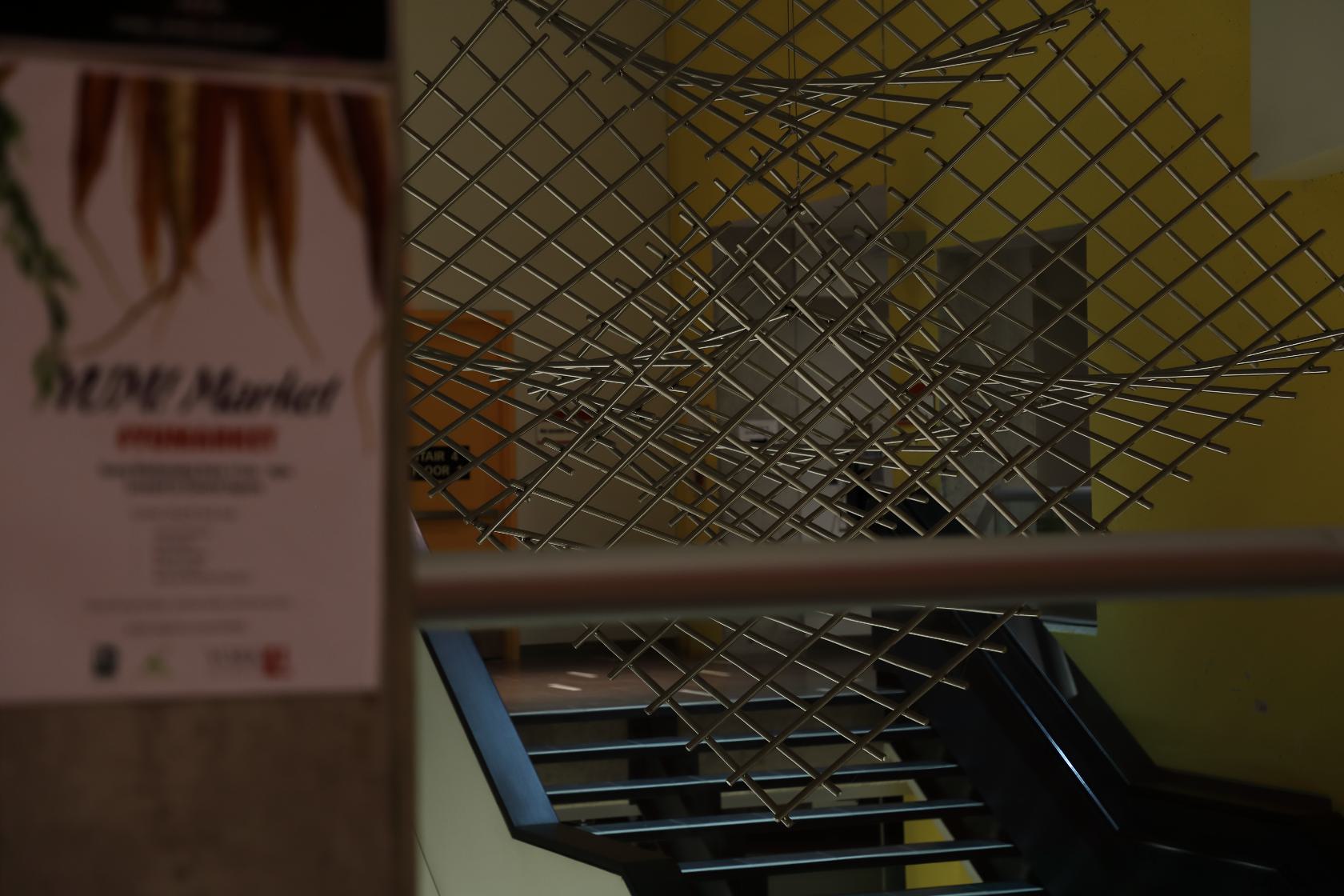}
	\includegraphics[width=0.16\textwidth]{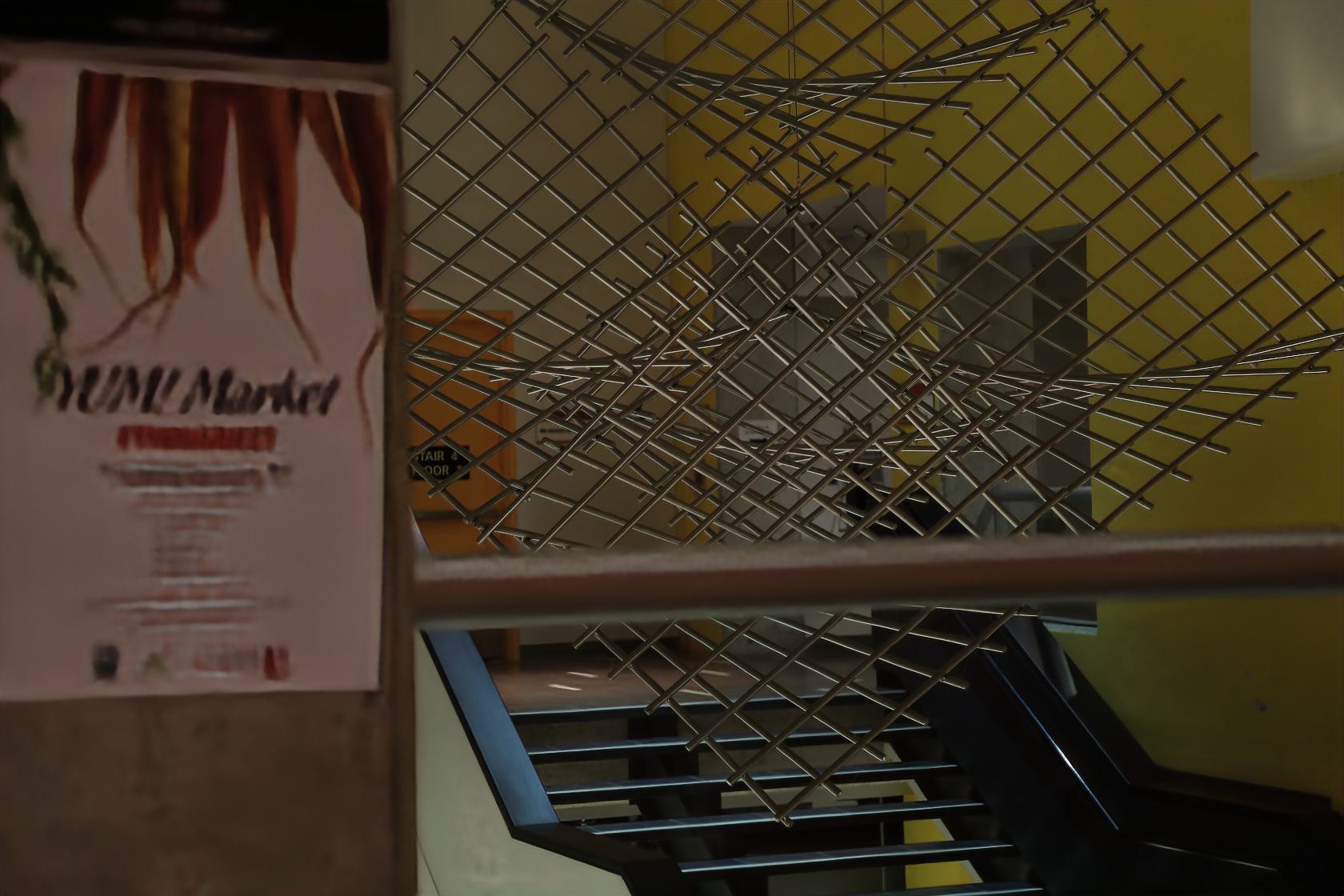}
	\includegraphics[width=0.16\textwidth]{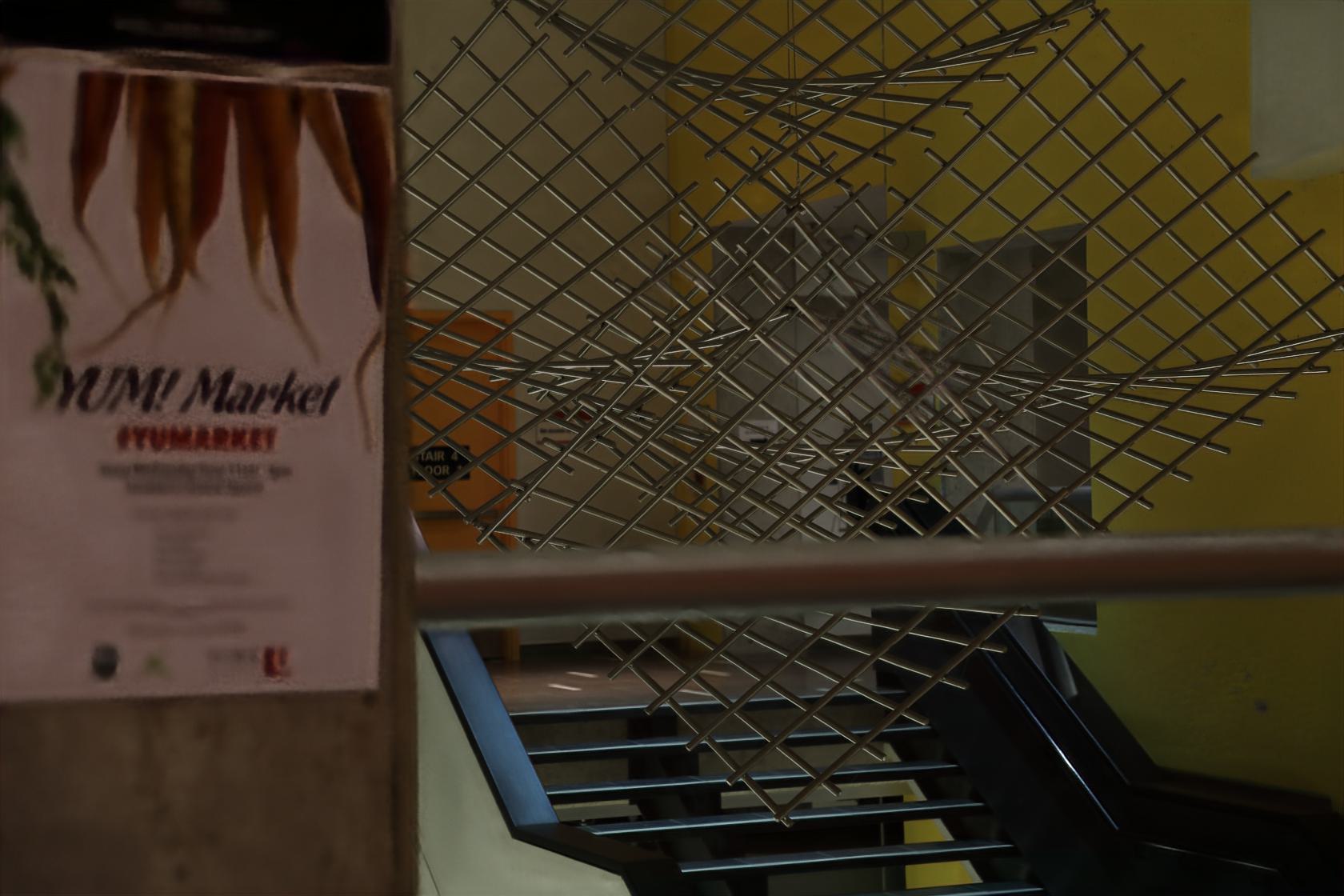}
	\includegraphics[width=0.16\textwidth]{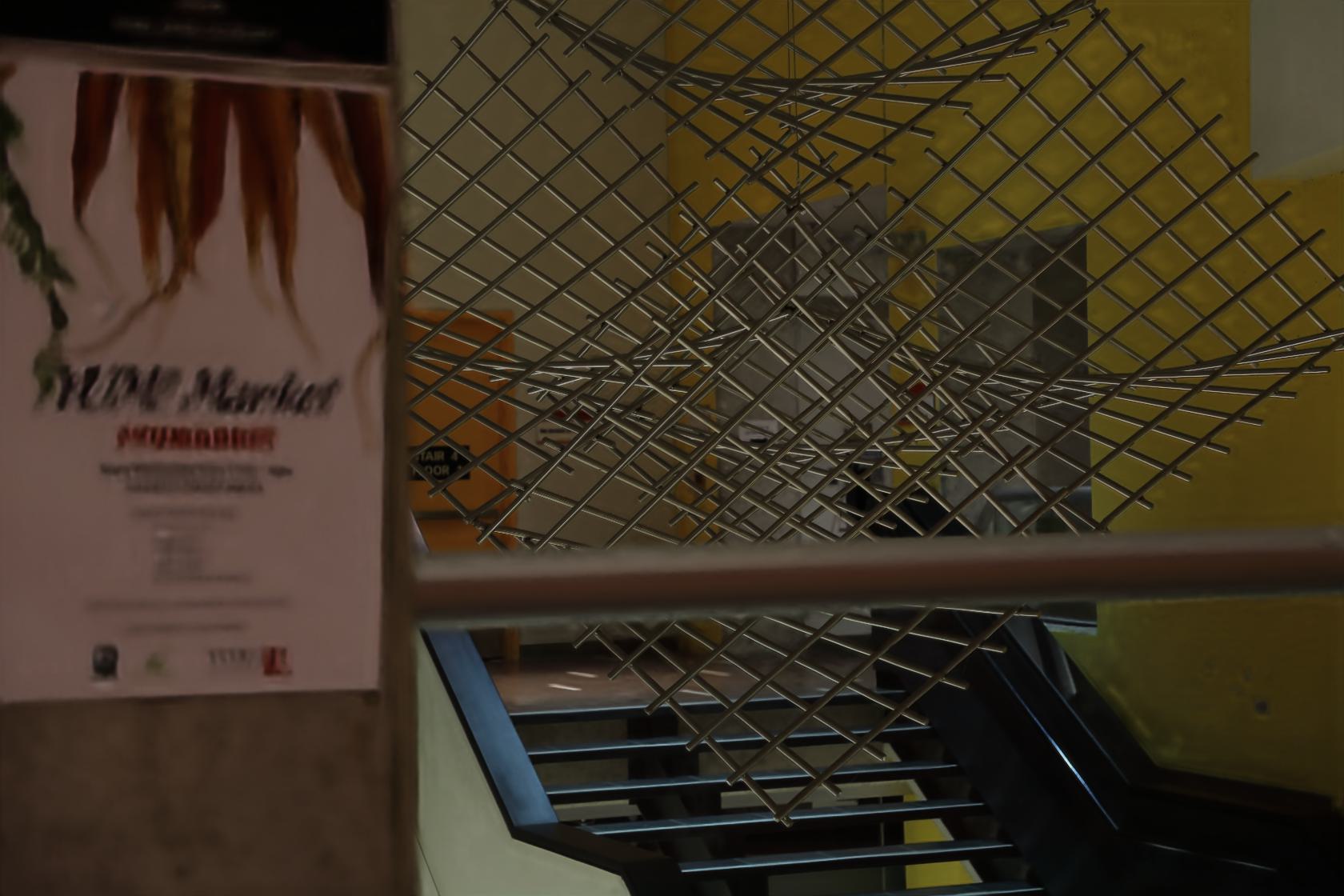}
	\includegraphics[width=0.16\textwidth]{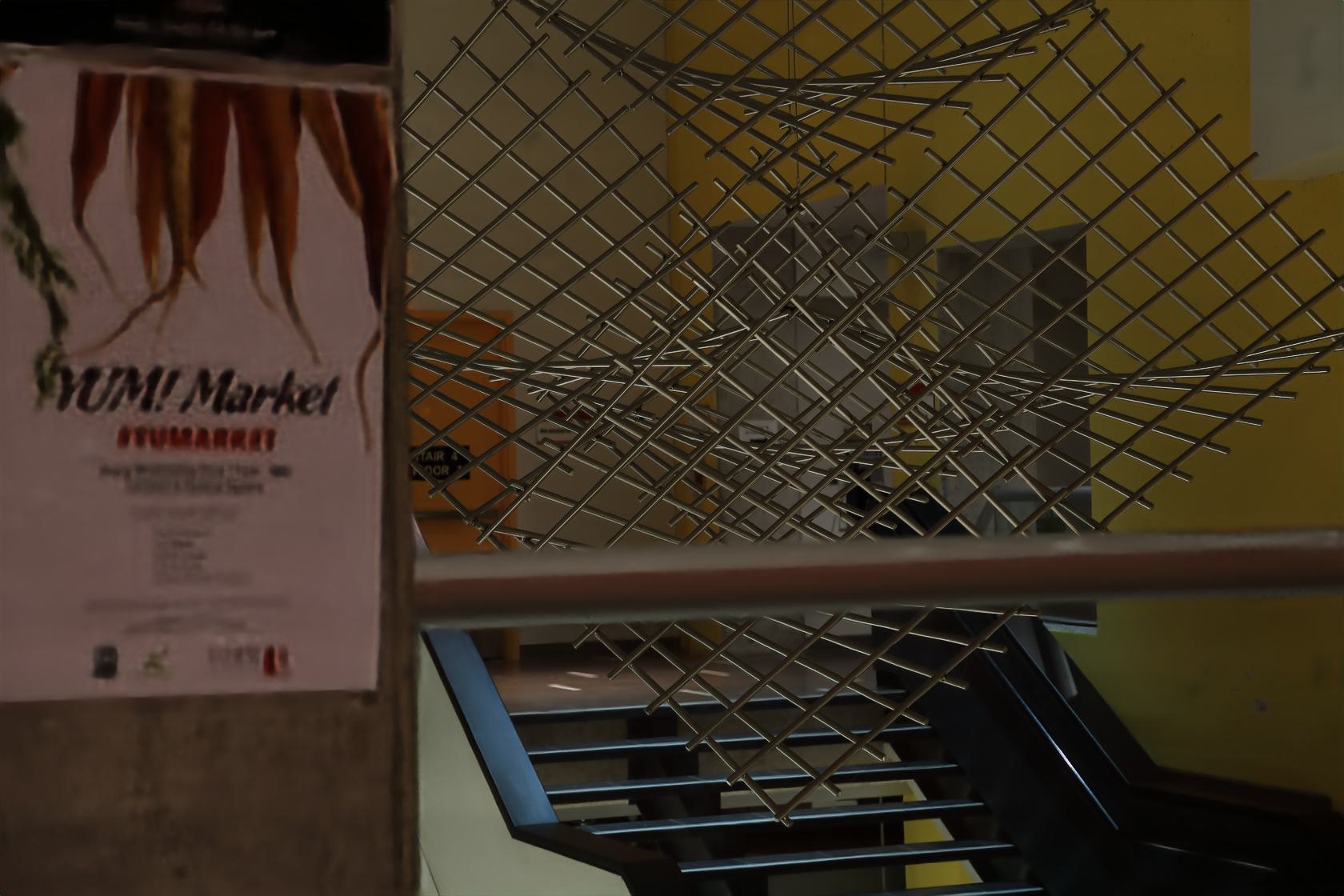}
	\includegraphics[width=0.16\textwidth]{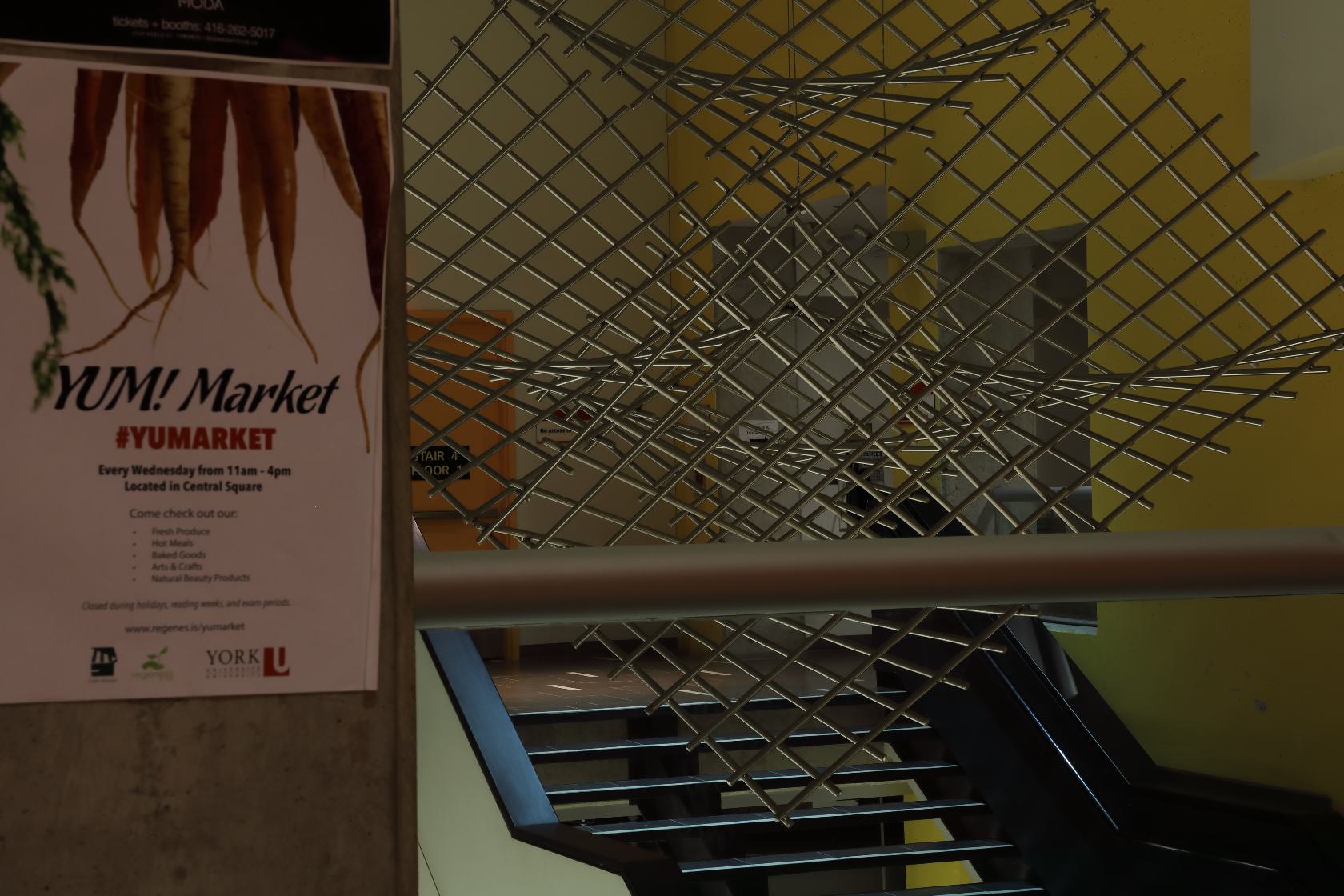}
	
	\frame{\includegraphics[width=0.16\textwidth]{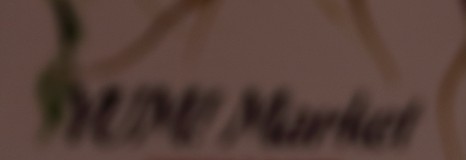}}
	\frame{\includegraphics[width=0.16\textwidth]{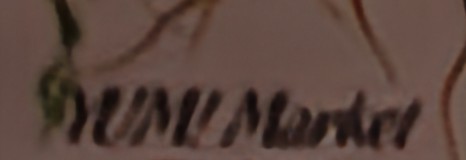}}
	\frame{\includegraphics[width=0.16\textwidth]{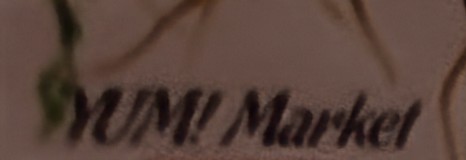}}
	\frame{\includegraphics[width=0.16\textwidth]{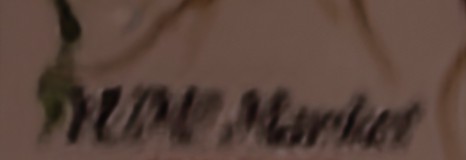}}
	\frame{\includegraphics[width=0.16\textwidth]{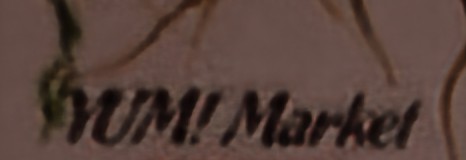}}
	\frame{\includegraphics[width=0.16\textwidth]{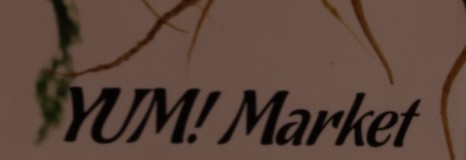}}
	
	\caption{Qualitative comparisons (part 2) of MCCNet against state-of-the-art methods. We include some zoomed and cropped regions for each test image to demonstrate the clear advantages of MCCNet. From left to right: Input, DPDNet \cite{dpdnet_eccv2020}, RDPD+\cite{rdpd_iccv2021}, MDP \cite{single_defocus_deblur_wacv2022}, MCCNet, Ground-truth.}
	\label{fig:qual_comp2}
\end{figure}

The qualitative results and comparison of MCCNet are shown in two parts in Fig.~\ref{fig:qual_comp1} and \ref{fig:qual_comp2} due to the space limitation. In Fig.~\ref{fig:qual_comp1}, the improvements of MCCNet are visible in the zoomed section of test images. In the first image, MCCNet can remove the defocus blur on words much better than the other methods, making it possible to read them. In the second image, it is visible that the other methods fail to recover the lines on the wall and floor fully. MCC recovers the text details in the third, fourth, and fifth images much better than other methods.

In Fig.~\ref{fig:qual_comp2}, it can be seen that MCC can recover the lamp post details much better than other methods in the first image. Similarly, in the second and third images, the details of the thin net and lines in the wall are recovered by MCCNet, while other methods fail to do so.

The qualitative results and comparison of MCCNet against the state-of-the-art methods reveal the superiority of MCCNet in removing the defocus blur.

\section{Ablation Study}
\label{sec:ablationstudies}

In this section, we present the ablation study results to understand the impact of various modules in MCCNet. The ablation study was designed to understand the impact following modules: Multi-scale Fusion module, Cross-correlation module and Multi-scale Feature Extraction module. We remove these modules individually and train MCCNet to observe the resulting PSNR and SSIM values. Tab.~\ref{tab:ablationstudy} shows the results of this ablation study.

It can be seen in Tab.~\ref{tab:ablationstudy} that removing the Multi-scale Fusion module decreases the PSNR and SSIM values slightly, whereas removing the Cross-correlation module decreases the PSNR and SSIM values considerably. The model does not use a siamese encoder in the \emph{no Cross-correlation module} setting, the DP left and right images are concatenated and fed to the network. Finally, when the Multi-scale Feature Extraction module is also removed, the performance drops even more. Interestingly, adding the Cross-correlation module (\emph{No Multi-scale Fusion} setting) leads to state-of-the-art PSNR and SSIM results.

\begin{table}[t]
	\renewcommand{\arraystretch}{1.4}
	\centering
	\setlength{\tabcolsep}{1em}
	\caption{Results of ablation study to understand the effect of different modules in MCCNet.}
	\label{tab:ablationstudy}
	\vspace{-8pt}
	\scalebox{0.8}{	
		\begin{tabular}{m{2.3cm}|c|c}
			\hline						
			\textbf{Setting}					& \textbf{PSNR}  & \textbf{SSIM}  \\ \hline \hline
			
			MCCNet 								& \textbf{25.85} & \textbf{0.802} \\  \hline
			
			No Multi-scale Fusion 				& 25.61 & 0.791 \\  \hline 
			
			No Cross-correlation 				& 25.12 & 0.784 \\  \hline 
			
			No Muti-scale Feature Extraction 	& 24.82 & 0.776 \\ \hline  
		\end{tabular}
	}	
	\vspace{-14pt}
\end{table}

\section{Conclusion}
\label{sec:conclusion}

In this work, we proposed a new defocused image deblurring method from dual-pixel images using an explicit cross-correlation between the dual-pixel image pair. More specifically, we adopted multi-scale cross-correlation to handle blur and disparities at different scales. Qualitative and quantitative evaluation of the proposed method established its superior performance compared to the state-of-the-art methods in the literature. The proposed method achieves state-of-the-art results with significantly less computational complexity than most defocus deblurring methods. Additionally, an ablation study was conducted to demonstrate the efficacy of various modules in the proposed network architecture.

%
%
%
\bibliographystyle{splncs04}
\bibliography{final_paper}
\end{document}